%% file: main.tex
\documentclass[letterpaper]{article} 
\usepackage{aaai25}
\usepackage{times}  
\usepackage{helvet}  
\usepackage{courier}  
\usepackage[hyphens]{url}  
\usepackage{graphicx} 
\usepackage{natbib}  
\usepackage{caption} 
\usepackage{algorithm}
\usepackage{algorithmic}

\usepackage{newfloat}
\usepackage{listings}
\DeclareCaptionStyle{ruled}{labelfont=normalfont,labelsep=colon,strut=off} 
\floatstyle{ruled}
\newfloat{listing}{tb}{lst}{}
\floatname{listing}{Listing}
\usepackage{colortbl}
\usepackage{booktabs}
\definecolor{mygray}{gray}{.9}
\usepackage{graphicx}
\newcommand{\SmallHeading}[1]{\noindent\textbf{#1.}\quad}

\usepackage{tikz}
\usepackage[edges]{forest}
\definecolor{hidden-draw}{RGB}{5, 4, 36}
\definecolor{hidden-blue}{RGB}{194,232,247}
\definecolor{hidden-orange}{RGB}{243,202,120}
\definecolor{hidden-yellow}{RGB}{242,244,193}
\definecolor{tree-level-1}{RGB}{245,20,85}
\definecolor{tree-level-2}{RGB}{246,86,118}
\definecolor{tree-level-3}{RGB}{248,177,193}
\definecolor{tree-leaf}{RGB}{176,230,198}
\definecolor{taxonomyColor}{RGB}{200,200,200} 

\definecolor{quantificationColor}{RGB}{243,202,120} 

\definecolor{futureColor}{RGB}{173,216,230} 

\usetikzlibrary{arrows.meta, positioning}

\usepackage{subcaption}
\usepackage{pgfplots}
\pgfplotsset{compat=1.17}
\usetikzlibrary{pgfplots.groupplots}

\usepackage[utf8]{inputenc}
\usepackage[T1]{fontenc}
\usepackage{newunicodechar}
\newunicodechar{❌}{\textcolor{red}{\texttimes}}

\usepackage{amsthm}
\usepackage{thmtools}
\declaretheoremstyle[
  spaceabove=\topsep,
  spacebelow=\topsep,
  headfont=\normalfont\bfseries,
  notefont=\normalfont\bfseries,
  bodyfont=\normalfont\itshape,
  headpunct=,
]{myplain}

\theoremstyle{myplain}

\newtheorem{definition}{Definition}

\usepackage{amsmath}
\usepackage{footnote}

\usepackage{amsfonts}

\usepackage[greek,english]{babel}

\usepackage{dsfont}

\usepackage{soul}

\usepackage{longtable}

\usepackage{xspace}
\colorlet{middleBarColor}{white}
\colorlet{leftBarColor}{blue}
\colorlet{rightBarColor}{red}

\newif\ifcomments
\commentstrue
\ifcomments
\definecolor{shaobopurple}{rgb}{0.8,0.0,0.8}
\newcommand\shaobo[1]{\textcolor{shaobopurple}{\textsf{\scriptsize[\textbf{shaobo\@:} #1]}}} 
\newcommand\shaoboi[1]{\textcolor{shaobopurple}{#1}} 
\newcommand\shaobom[1]{\marginpar{\raggedright\tiny\textcolor{shaobopurple}{\textsf{{\bfseries Shaobo\@:} #1}}}} 
\newcommand\shaobos{\bgroup\markoverwith{\textcolor{shaobopurple}{\rule[.4ex]{2pt}{0.8pt}}}\ULon} 
\else
\newcommand\shaobo[1]{}
\newcommand\shaoboi[1]{\ignorespaces}
\newcommand\shaobom[1]{}
\newcommand\shaobos[1]{#1}
\fi

\usepackage{mathtools}

\definecolor{codegreen}{rgb}{0,0.6,0}
\definecolor{codegray}{rgb}{0.5,0.5,0.5}
\definecolor{codepurple}{rgb}{0.58,0,0.82}
\definecolor{backcolour}{rgb}{0.95,0.95,0.92}

\lstdefinestyle{mystyle}{
    backgroundcolor=\color{backcolour},   
    commentstyle=\color{codegreen},
    keywordstyle=\color{magenta},
    numberstyle=\tiny\color{codegray},
    stringstyle=\color{codepurple},
    basicstyle=\ttfamily\footnotesize,
    breakatwhitespace=false,         
    breaklines=true,                 
    captionpos=b,                    
    keepspaces=true,                 
    numbers=left,                    
    numbersep=5pt,                  
    showspaces=false,                
    showstringspaces=false,
    showtabs=false,                  
    tabsize=2
}

\usepackage{xcolor} 

\definecolor{deepskyblue4}{RGB}{0, 104, 139} 
\definecolor{darkorange}{RGB}{255, 140, 0} 

\newcommand{\inlineLeftGradientBar}{
  \begin{tikzpicture}[baseline=-0.1ex]
    \shade[left color=deepskyblue4, right color=white] (0,0) rectangle (0.4,0.2);
  \end{tikzpicture}
}

\newcommand{\inlineRightGradientBar}{
  \begin{tikzpicture}[baseline=-0.1ex]
    \shade[right color=darkorange, left color=white] (0,0) rectangle (0.4,0.2);
  \end{tikzpicture}
}

\newcommand{\CS}[2]{\mathcal{CS}(#2 \vert #1)} 

\usepackage{amsmath}

\newcommand{\tauA}{$\tau$-$\mathcal{A}$\xspace}
\newcommand{\tauD}{$\tau$-$\mathcal{D}$\xspace}
\newcommand{\tauOverall}{$\tau$-all\xspace}
\newcommand{\CGP}{CGP\xspace}
\newcommand{\IGC}{IGC\xspace}

\newcommand{\CircleA}{\raisebox{-0.2ex}{\tikz\fill[orange, thick] (0,0.8) circle (.8ex - 0.2pt);}\xspace}
\definecolor{steelblue}{RGB}{70, 130, 180}
\newcommand{\SquareB}{\raisebox{-0.1ex}{\tikz\fill[steelblue, thick] (0,0) rectangle (1.4ex,1.4ex);}\xspace}

\usepackage{xinttools} 
\newcommand{\CircleANum}[2]{\raisebox{-0.2ex}{\tikz\draw[fill=orange!#1, thick] (0,0) circle (1.4ex) node[align=center] {\textcolor{black}{\small #2}};}\hspace{0.0em}\xspace}
\newcommand{\SquareDNum}[2]{\raisebox{-0.2ex}{\tikz\draw[fill=steelblue!#1, thick] (0,0) rectangle (2.5ex,2.5ex) node[midway]{#2};}\xspace}

\newcommand{\Figure}{Figure\xspace}
\newcommand{\Table}{Table\xspace}

\newcommand{\Section}{Section\xspace}
\newcommand{\Appendix}{Appendix\xspace}

\usepackage[framemethod=TikZ]{mdframed} 
\newenvironment{framedwithtag}[6]{%
  \mdfsetup{%
    frametitlealignment=\raggedright,
    frametitleaboveskip=-\ht\strutbox,
    frametitlebelowskip=0pt,
    frametitlefont=\normalfont,
    innertopmargin=10pt, 
    innerbottommargin=10pt,
    innerleftmargin=10pt,
    innerrightmargin=10pt,
    roundcorner=5pt,
    tikzsetting={line width=0pt}, 
    backgroundcolor=#3, 
    linecolor=gray!50,
    singleextra={%
      \node[anchor=north,rectangle,rounded corners,fill=#2,inner sep=3pt,outer sep=0pt,draw=none] at ([xshift=#4,yshift=#5]P) {#1};
    },
    linewidth=0.7pt, 
    skipabove=20pt, 
    skipbelow=10pt, 
  }
  \begin{mdframed}%
  #6
  \end{mdframed}%
}

\definecolor{light-gray}{gray}{0.95}
\newcommand{\code}[1]{\colorbox{light-gray}{\texttt{#1}}}
\newcommand{\variableP}[1]{\texttt{#1}}

\usepackage{array}
\newcolumntype{L}[1]{>{\raggedright\let\newline\\\arraybackslash\hspace{0pt}}m{#1}}
\newcolumntype{C}[1]{>{\centering\let\newline\\\arraybackslash\hspace{0pt}}m{#1}}
\newcolumntype{R}[1]{>{\raggedleft\let\newline\\\arraybackslash\hspace{0pt}}m{#1}}

\newcommand{\rankingColor}[0]{pink!30}
\newcommand{\crossColor}[0]{teal!30}
\newcommand{\clusteringColor}[0]{olive!30}

\newcommand{\midgrayline}[0]{\arrayrulecolor{gray!50}\midrule\arrayrulecolor{black}}

\usepackage{pifont}
\newcommand{\myCheckMark}[0]{\ding{51}}
\newcommand{\myCrossMark}[0]{\ding{55}}

\definecolor{darkblue}{RGB}{0, 0, 139}

\newcommand{\circlenum}[1]{\raisebox{-0.45
ex}{\tikz\draw[thick,orange] (0,0) circle (0.95ex) node[align=center] {\textcolor{black}{\textcolor{orange}{\tiny #1}}};}}
\newcommand{\squarenum}[1]{\raisebox{-0.45ex}{\tikz\draw[thick,darkblue] (0,0) rectangle (1.8ex,1.8ex) node[midway]{\textcolor{darkblue}{\tiny #1}};}}

\usepackage{adjustbox}

\title{Nuance Matters: Probing Epistemic Consistency in Causal Reasoning}
\author {
    Shaobo Cui\textsuperscript{\rm 1},
    Junyou Li\textsuperscript{\rm 2},
    Luca Mouchel\textsuperscript{\rm 1}, 
    Yiyang Feng\textsuperscript{\rm 1}, 
    Boi Faltings\textsuperscript{\rm 1}    
}
\affiliations {
    \textsuperscript{\rm 1}EPFL, Switzerland, \;
    \textsuperscript{\rm 2}University of Waterloo, Canada\\
    shaobo.cui@epfl.ch, j2626li@uwaterloo.ca, luca.mouchel@epfl.ch, yiyang.feng@epfl.ch, boi.faltings@epfl.ch
}

\usepackage{bibentry}

\begin{document}

\maketitle

\begin{abstract}
Previous research on causal reasoning often overlooks the subtleties crucial to understanding causal reasoning.  
To address this gap, our study introduces the concept of \textit{causal epistemic consistency}, which focuses on the self-consistency of Large Language Models~(LLMs) in differentiating intermediates with nuanced differences in causal reasoning. 
We propose a suite of novel metrics -- intensity ranking concordance, cross-group position agreement, and intra-group clustering -- to evaluate LLMs on this front. 
Through extensive empirical studies on 21 high-profile LLMs, including GPT-4, Claude3, and LLaMA3-70B, we have favoring evidence that current models struggle to maintain epistemic consistency in identifying the polarity and intensity of intermediates in causal reasoning. 
Additionally, we explore the potential of using internal token probabilities as an auxiliary tool to maintain causal epistemic consistency. 
In summary, our study bridges a critical gap in AI research by investigating the self-consistency over fine-grained intermediates involved in causal reasoning. 
\end{abstract}

%

\section{Introduction} \label{sec:intro}
\input{sections/introduction}

\section{Task Definition} \label{sec:task}
\input{sections/task}


\section{Metrics for Measuring Causal Epistemic Consistency} \label{sec:metrics}
\input{sections/metrics}

\section{Causal Epistemic Consistency of LLMs} \label{sec:empirical}
\input{sections/analysis}

\section{Beyond Prompting: Leveraging Internal Token Probability} \label{sec:correlation}
\input{sections/correlation}

\section{Related Work} \label{sec:related_work}
\input{sections/related_work}

\section{Conclusion} \label{sec:conclusion}
\input{sections/conclusion}

\bibliography{aaai25}

\appendix

\onecolumn
\clearpage
\section{Causal Conjunctions} \label{appendix:conjunction}
\input{appendices/conjunctions}

\section{Experimental Setup} \label{appendix:setup}
\input{appendices/setup}

\section{Further Discussion on Proposed Metrics} \label{appendix:metrics}
\input{appendices/metrics}

\section{More Results} \label{appendix:results}
\input{appendices/results}

\end{document}

%% file: sections/introduction.tex
Previous studies in causal reasoning have primarily focused on discovering or determining the existence of a causal relationship between two variables~\cite{roemmele-etal-2011-choice,cui-etal-2024-exploring}. 
However, these causal relationships are not always absolute. They can be heavily influenced by additional intermediate factors, which may vary in both polarity and intensity~\cite{fitzgerald1998towards,bauman2002toward}. The polarity of these intermediates indicates whether they support or defeat (oppose) the original causal relationship, while their intensity determines the strength of this supporting or defeating influence. 

Forming fine-grained differentiation is essential for precise causal modeling~\cite{iwasaki1994causality}; however, it is insufficient for LLMs to merely generate these intermediates. It is as equally important to ensure that these intermediates are reliable and credible~\cite{shi2023know}. 
One method to verify this is through assessing the consistency of LLMs' perception of the intermediates. We posit that if LLMs can correctly differentiate their generated intermediates based on varying polarities and intensities, these intermediates are self-consistent and thus, more reliable for making predictions and decisions. Drawing from this insight, our study proposes the concept of ``\textbf{causal epistemic consistency}'': 
\begin{definition}[Causal epistemic consistency]
Causal epistemic consistency refers to an LLM's ability to maintain \ul{self-consistency} in differentiating its generated intermediates in three aspects: (i) discerning \textbf{intensity}: accurately assessing the intensity nuance in their causal impact. (ii) differentiating \textbf{polarity}: effectively distinguishing between supporting and defeating intermediates, and (iii) forming cohesive \textbf{clusters}: creating well-separated clusters of intermediates based on their polarity and intensity. 
\end{definition}

\begin{figure}[!t]
    \centering
    \includegraphics[width=0.99999999\columnwidth]{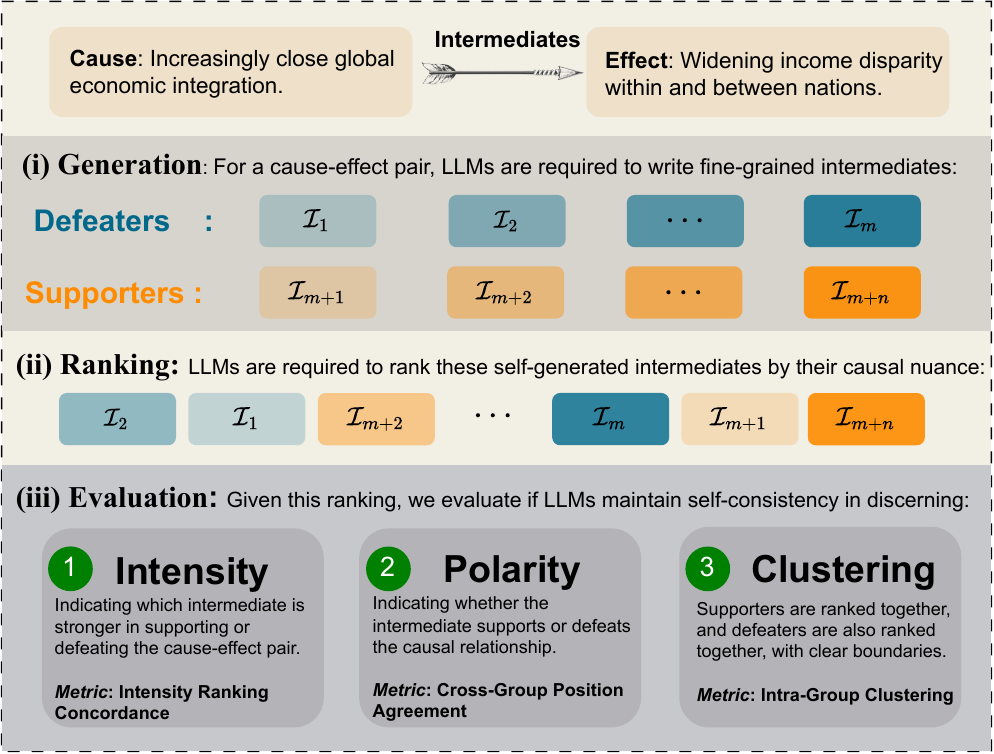}
    \caption{Overview of the evaluation framework for causal epistemic consistency. The first step involves instructing LLMs to generate fine-grained intermediates that influence a given causal relationship differently. The second step requires LLMs to rank their own generations based on their causal nuance. Finally, the proposed metrics are used to assess the self-consistency between ranking and generation, i.e., the LLMs' causal epistemic consistency.} 
    \label{fig:introduction:example}
\end{figure}

To quantify LLMs' ability to maintain causal epistemic consistency in the aforementioned aspects, we introduce a suite of novel metrics. These metrics include (i) Intensity ranking concordance, which measures the models' self-consistency in ranking self-generated intermediates with varying intensity; (ii) Cross-group position~(\CGP) agreement, which indicates the models' consistency in determining the polarity of intermediates, specifically whether they support or defeat the original causal relationship; and (iii) Intra-group clustering~(IGC), which assesses models' consistency to rank its generated intermediates of the same type closely together. We illustrate the evaluation framework of causal epistemic consistency in \Figure~\ref{fig:introduction:example}.

To unravel the causal epistemic consistency of current LLMs, our empirical study evaluates 21 high-profile LLMs, including the renowned closed-source GPT, Claude, and Gemini series, alongside various scales of cutting-edge open-source alternatives such as  Gemma~(2B and 7B) \cite{mesnard-etal-2024-gemma}, LLaMA2~(7B, 13B, and 70B)~\cite{touvron-etal-2023-llama}, Phi-3~(3.8B, 7B, and 14B)~\cite{abdin2024phi}, and LLaMA3~(8B and 70B)~\cite{meta2024introducing}. Contrary to initial expectations that LLMs would exhibit satisfactory performance, our findings reveal their striking incompetence in keeping causal epistemic consistency. Remarkably, even the advanced GPT-4 model performs unsatisfactorily.  This underscores the complexities and challenges these models face in maintaining causal consistency and capturing causal nuances.

Furthermore, we explore whether internal token probability can serve as a useful signal for LLMs to maintain causal epistemic consistency. 
Our comprehensive empirical study highlights the application scope of internal token probability for LLMs to maintain causal epistemic consistency.

To summarize, our contributions are fourfold: 
\begin{enumerate}
    \item \textbf{Introduction of Causal Epistemic Consistency:} We propose the novel concept of causal epistemic consistency over fine-grained intermediates in causal reasoning, emphasizing self-consistency in differentiating the nuances hidden in fine-grained intermediates. 
    \item \textbf{Development of Evaluation Metrics:} We introduce a comprehensive suite of metrics designed to assess LLMs' causal epistemic consistency, covering aspects of intensity ranking concordance, cross-group position agreement, and intra-group clustering. 
    \item \textbf{Extensive Empirical Evaluation:} We assess the performance of 21 LLMs on their causal epistemic consistency, highlighting their deficiencies in maintaining causal epistemic consistency.   
    \item \textbf{Internal Token Probability Exploration:} We investigate the potential of using internal token probabilities as an auxiliary tool to help LLMs maintain causal epistemic consistency and highlight its application scope. 
\end{enumerate}

%% file: sections/task.tex
 \subsection{Problem Formulations} \label{sec:task:problem}

Causal epistemic consistency measures an LLM's self-consistency between generating fine-grained intermediates and subsequently ranking those fine-grained intermediates. 

Specifically, in the generation phase, for a defeasible cause-effect pair $(C, E)$, an LLM is tasked with generating an ordered sequence $\mathcal{I}$ of fine-grained intermediates, consisting of a subsequence $\mathcal{D}=(\mathcal{I}_1, \mathcal{I}_2, \cdots, \mathcal{I}_m)$ as the defeater group and a subsequence $\mathcal{A}=(\mathcal{I}_{m+1}, \mathcal{I}_{m+2}, \cdots, \mathcal{I}_{m+n})$ as the supporter group.
Each individual intermediate changes the causal strength of $(C, E)$ differently. Specifically, the causal influence of these intermediates is expected in the following order: 
\newcommand{\gradientbox}[3]{
    \tikz[baseline=(X.base)]
        \node[rectangle, inner sep=2mm, outer sep=0pt, left color=#1, right color=#2] (X) {#3};
}
\begin{equation} \label{eq:intermediate_causal_strength}
\begin{aligned}
    \gradientbox{deepskyblue4!80}{deepskyblue4!50}{$\CS{C\oplus \mathcal{I}_1}{E}$}&\gradientbox{deepskyblue4!50}{deepskyblue4!5}{$\leq \cdots \leq \CS{C\oplus \mathcal{I}_m}{E}$} \\    
    &\leq \CS{C}{E} \\
    \gradientbox{darkorange!5}{darkorange!35}{$\CS{C\oplus \mathcal{I}_{m+1}}{E}$}&\gradientbox{darkorange!35}{darkorange!80}{$\leq \cdots \leq \CS{C\oplus \mathcal{I}_{m+n}}{E}$} \\    
\end{aligned}
\end{equation}
where $\CS{C}{E}$ measures the causal strength~\cite{luo-etal-2016-commonsense,zhang-etal-2022-rock}, quantifying the likelihood that the cause event $C$ would lead to the occurrence of the effect event $E$. 
\footnote{In this context, we assume that only one fine-grained intermediate is active for a cause-effect pair at a time. This design choice reflects the reality that a single argument is more often responsible for influencing the causal relationship than multiple arguments acting simultaneously.}
The $\oplus$ means the combination of two events.
The gradient bar \protect\inlineLeftGradientBar{} illustrates the varying degrees of intensity of the defeating intermediates, while the gradient bar \protect\inlineRightGradientBar{} represents the supporting intermediates. The color gradient darkens as the intensity increases, with a darker shades indicating a stronger influence, whether supporting or defeating. 

Subsequently, in the ranking phase, the same LLM is asked again to rank its own generated intermediates $\mathcal{I}$, obtaining $\mathcal{I}'$, a permutation of $\mathcal{I}$. Ideally, an LLM with perfect causal epistemic consistency should have $\mathcal{I}=\mathcal{I}'$, satisfying the requirements of intensity, polarity, and clustering perfectly.

\subsection{Key Research Questions} \label{sec:task:rqs}

The study addresses three primary research questions:

\begin{itemize}
    \item \textbf{RQ I}: How can we comprehensively measure the ability of LLMs to maintain the epistemic consistency over fine-grained intermediates in causal reasoning? 
    \item \textbf{RQ II}: How well do current LLMs, with varying architectures and scales, maintain their causal epistemic consistency? 
    \item \textbf{RQ III}: Are there any alternatives to prompting for LLMs to maintain causal epistemic consistency? 
\end{itemize}

To answer \textbf{RQ I}, we propose novel metrics introduced in \Section~\ref{sec:metrics}, which not only serve our specific study but also have broader applications across various tasks. In \Section~\ref{sec:empirical}, we dive into the performance of twenty-one leading LLMs, exploring their ability to maintain epistemic consistency, thereby addressing \textbf{RQ II}. Lastly, in \Section~\ref{sec:correlation}
, we assess whether internal token probability offers a more effective—or perhaps less effective—alternative to prompting for preserving causal epistemic consistency in LLMs, answering \textbf{RQ III}.

%% file: sections/metrics.tex
\begin{figure}[htp!]
    \centering
    \resizebox{0.99\linewidth}{!}{
        \includegraphics{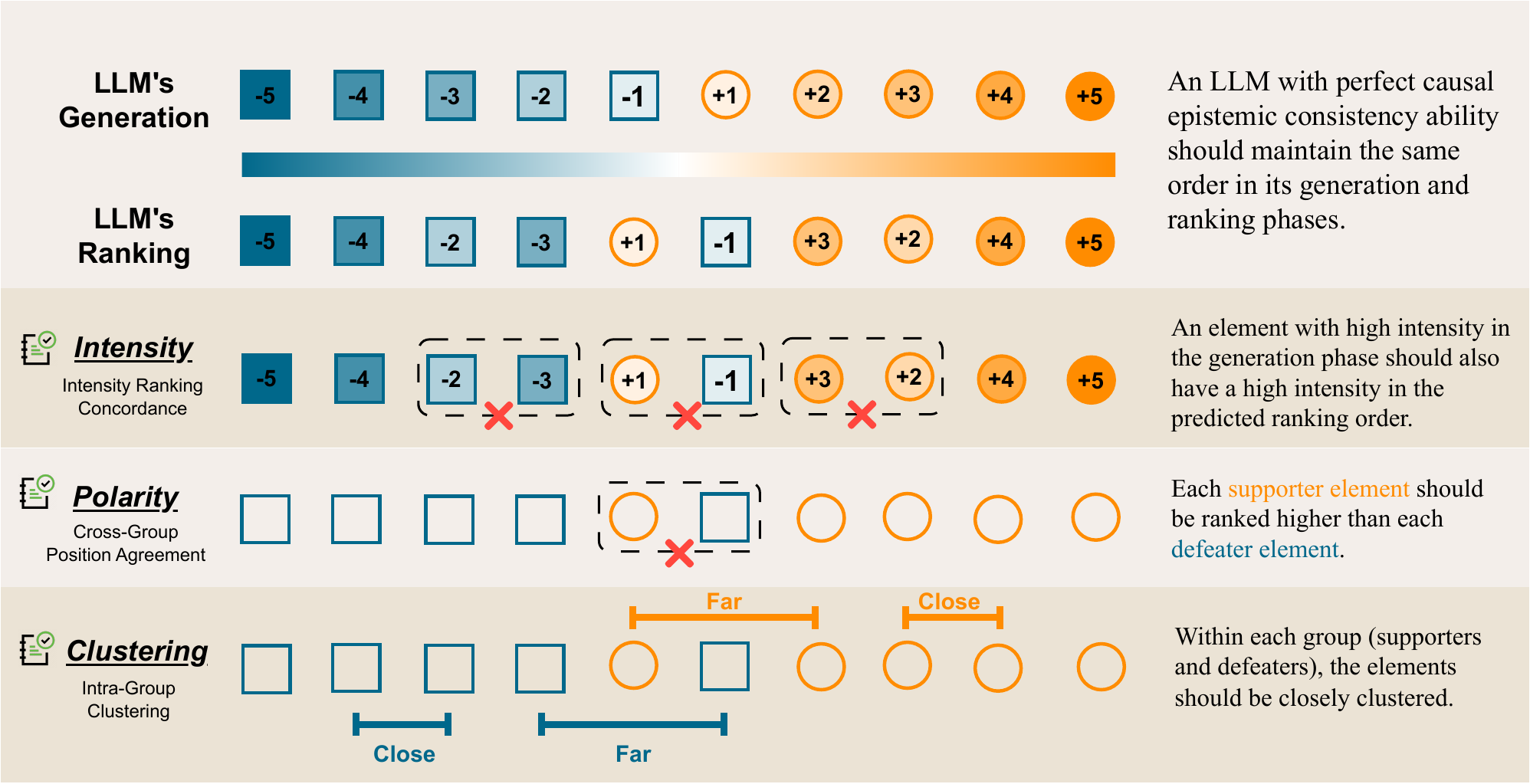}
    }
    \caption{Illustration of the proposed metrics from three aspects: intensity~(\Section~\ref{sec:metrics:ranking}), polarity~(\Section~\ref{sec:metrics:cgp}), and clustering~(\Section~\ref{sec:metrics:clustering}). These metrics measure the self-consistency of LLMs in generating and ranking \textcolor{orange}{supporting}~(\circlenum{}) and \textcolor{steelblue}{defeating}~(\squarenum{}) intermediates with varying intensities. Numbers \squarenum{-5}, \squarenum{-4}, ..., \circlenum{+4}, \circlenum{+5} indicate the intensity of the generated intermediates, with the lowest value~(\squarenum{-5}) being the strongest generated defeater and the highest value~(\circlenum{+5}) the strongest generated supporter.
    }
    \label{fig:metrics:illustration}
\end{figure}

To evaluate the causal epistemic consistency of LLMs from the aspects of intensity, polarity, and clustering,  we propose three types of automatic metrics: intensity ranking concordance, cross-group position agreement, and intra-group clustering. A graphical illustration of these metrics is shown in \Figure~\ref{fig:metrics:illustration}.
The mathematical notations below are consistent with \Section~\ref{sec:task:problem}.

\subsection{Intensity: Intensity Ranking Concordance} \label{sec:metrics:ranking}
To assess the concordance between the order from the generation phase and the order from the ranking phase of these fine-grained intermediates, we leverage the Kendall Tau distance~\cite{kendall-1938-new}. This metric quantifies the similarity between two orders by counting the number of pairwise agreements and disagreements. 
For a sequence $\mathcal{I}$ of LLM-generated intermediates and its permutation $\mathcal{I}'$ ranked by the same LLM, a pair of elements from $\mathcal{I}$ is called \textit{concordant} if they appear in the same order in both $\mathcal{I}$ and $\mathcal{I}'$. Conversely, the pair is called \textit{discordant} if their order is reversed in $\mathcal{I}'$ compared to $\mathcal{I}$.
The Kendall Tau $\tau$ is calculated as: 
\begin{equation}
    \tau = \frac{(\text{\# concordant pairs}) - (\text{\# discordant pairs})}{k(k-1)/2} 
\end{equation}
where $k$ is the number of elements in the list, and $k(k-1)/2$ is the total number of pairs. The metric ranges from -1 to 1, where 1 indicates that these two lists are identical; -1 indicates completely reversed rankings; and values close to 0 indicate no association between the two lists. 
For our task, we have three intensity ranking concordance metrics: \tauA, \tauD, and \tauOverall, which evaluate the intensity ranking concordance within the supporter group, the defeater group, and the entire sequence of intermediates, respectively.

\subsection{Polarity: Cross-Group Position~(\CGP)} \label{sec:metrics:cgp} 
To assess the relative positioning of elements between these two polarities--the defeater group $\mathcal{D}$ and the supporter group $\mathcal{A}$--we propose the \ul{C}ross-\ul{G}roup \ul{P}osition~(\CGP) metric. This metric penalizes instances where elements from $\mathcal{A}$ are ranked lower than those from $\mathcal{D}$~\footnote{We define the index of the strongest defeater to be the lowest and the strongest supporter to be the highest, consistent with \Section~\ref {sec:task:problem}.}. 
Specifically, \CGP is defined as:
\begin{equation}
   \resizebox{0.99\hsize}{!}{%
   $\text{\CGP}(\mathcal{I}', \mathcal{A}, \mathcal{D}) = 1 - \frac{\sum_{a\in\mathcal{A}} 
   \sum_{d\in\mathcal{D}} \mathds{1}[\text{index}(a) < \text{index}(d)]}{\vert \mathcal{A} \vert \times \vert \mathcal{D} \vert}$
   }
\end{equation}
where $\text{index}(x)$ denotes the index of element $x$ in the ranked sequence $\mathcal{I'}$. $\mathds{1}[\cdot]$ denotes the indicator function that is set to 1 if the condition is true and 0 otherwise. \CGP measures how often elements from $\mathcal{A}$ precede the elements of $\mathcal{D}$ in the ranked sequence $\mathcal{I'}$. It is normalized to the range [0, 1] by dividing with the maximum possible violations, i.e., $\vert \mathcal{A} \vert \times \vert \mathcal{D} \vert$.  Higher values indicate better differentiation between groups $\mathcal{A}$ and $\mathcal{D}$. 

\begin{table*}[h]
\centering
\resizebox{0.9\linewidth}{!}{
\input{figures/table_minipages/full_model_comparison}
}
\caption{Empirical study of LLMs on the proposed metrics for causal epistemic consistency. }
  \label{tab:full_results}
\end{table*}

\subsection{Clustering: Intra-Group Clustering~(\IGC)} \label{sec:metrics:clustering}
In this subsection, we introduce \ul{I}ntra-\ul{G}roup \ul{C}lustering (\IGC), a metric for LLMs' causal epistemic consistency by assessing the clustering degree of supporting and defeating intermediates. The intuition behind \IGC is that all defeaters and all supporters should form cohesive clusters, with a minimal number of polarity changes~(from supporting to defeating, or vice versa) when iterating the sequence.

\SmallHeading{Clustering Distance Based on Polarity Change} Given the LLM-ranked intermediates $\mathcal{I}'$, we define $L_i$ to indicate which polarity~(supporter $\mathcal{A}$ or defeater $\mathcal{D}$) each intermediate $\mathcal{I}'_i$ belongs to. $L_i$ is represented as a binary polarity that either $L_i=\mathcal{A}$ or $L_i=\mathcal{D}$.
$d(i,j)$ is the sequence clustering distance between $\mathcal{I}'_i$ and $\mathcal{I}'_j$, calculated as follows: 
\begin{equation}
    d(i,j) = \displaystyle\sum_{k=i}^{j-1} \mathds{1}[L_k\neq L_{k+1} \land L_{k+1}\neq L_i]
\end{equation}
where $i<j$. The distance is based on the number of polarity changes, excluding reversions to the initial polarity. 

\SmallHeading{\IGC: A Measure of Clustering Quality in Sequence}
With the distance based on polarity change, we use the silhouette score~\cite{rousseeuw-1987-silhouettes,shahapure-nicholas-2020-cluster} to measure how similar an element is to its own cluster compared to other clusters in sequence: 
\begin{equation}\label{eq:silhouette_sample_score}
    s(i) = \frac{d_{nc}(i) - d_{ic}(i)}{\max(d_{ic}(i), d_{nc}(i))}
\end{equation}
where $d_{ic}(i)$ and $d_{nc}(i)$ are the intra-cluster distance and nearest cluster distance for each intermediate $\mathcal{I}'_i$.
\begin{enumerate}
    \item The intra-cluster distance $d_{ic}(i)$ captures the mean distance between $\mathcal{I}'_i$ and all other intermediates belonging to the same group, reflecting \textit{internal cohesion}. It is calculated as:
        \begin{equation}
            d_{ic}(i) = \frac{1}{|L_i|-1} \sum_{L_j = L_i, \mathcal{I}'_j \neq \mathcal{I}'_i} d(i, j).
        \end{equation}
    \item The nearest cluster distance $d_{nc}(i)$ captures the mean distance between $\mathcal{I}'_i$ and all other points belonging to a different group, demonstrating the level of \textit{separation} from other clusters. It is calculated as:
        \begin{equation}
            d_{nc}(i) = \frac{1}{|\mathcal{I}'|-|L_i|} \sum_{L_j \neq L_i} d(i, j).
        \end{equation}
\end{enumerate}

The final Intra-Group Clustering (\IGC) metric is computed as the average clustering of all elements:
\begin{equation}
    \text{\IGC} = \frac{1}{\vert \mathcal{I}' \vert} \sum_{i=1}^{\vert \mathcal{I}' \vert} s(i). 
\end{equation}

\SmallHeading{Range and Implications of \IGC }
The range of $s(i)$ is $[-1,1]$: (i) Close to 1: The element is near its own group and far from the neighboring groups; 
(ii) Close to 0: The element is on the border between its cluster and a neighboring cluster.
(iii) Close to -1: The element is in the wrong cluster.
\IGC quantifies the quality of cluster assignments, with a high score indicating well-clustered sequences. It is a general metric applicable to various contexts related to sequence clustering. Further details are in \Appendix~\ref{appendix:metrics:clustering}.

%% file: figures/table_minipages/full_model_comparison.tex
\newcommand{\stdvalue}[1]{$\pm$ \scriptsize #1}
\setlength{\columnsep}{2.3cm}

\begin{tabular}{l|p{\columnsep}p{\columnsep}p{\columnsep}|C{\columnsep}|C{\columnsep}}
\toprule 
Aspect & \multicolumn{3}{>{\columncolor{\rankingColor}}m{7.8cm}}{\centering \textit{Intensity Ranking Concordance}} 
& \multicolumn{1}{>{\columncolor{\crossColor}}m{\columnsep}}{\textit{Cross-Group Position}} 
& \multicolumn{1}{>{\columncolor{\clusteringColor}}m{\columnsep}}{\textit{Intra-Group Clustering}} 
\\
    \midrule
    & \tauA$\uparrow$ & \tauD$\uparrow$ & \tauOverall$\uparrow$ & \centering \CGP $\uparrow$ &  \multicolumn{1}{>{\columncolor{white}}c}{\IGC $\uparrow$} \\
    \midrule
    \multicolumn{6}{>{\columncolor{mygray}}c}{\textit{Closed-source LLMs}}\\ \midgrayline
    GPT-3.5 Turbo & 0.074 \stdvalue{0.429} & 0.045 \stdvalue{0.407} & 0.304 \stdvalue{0.409} & 0.750 \stdvalue{0.329} & 0.762 \stdvalue{0.244} \\
    GPT-4 & 0.384 \stdvalue{0.413} & 0.203 \stdvalue{0.440} & 0.587 \stdvalue{0.347} & 0.911 \stdvalue{0.235} & 0.916 \stdvalue{0.176} \\
    GPT-4 Turbo & 0.397 \stdvalue{0.541} & 0.226 \stdvalue{0.459} & 0.526 \stdvalue{0.510} & 0.849 \stdvalue{0.330} & 0.942 \stdvalue{0.151} \\
    GPT-4o mini & 0.142 \stdvalue{0.444} & 0.154 \stdvalue{0.418} & 0.472 \stdvalue{0.375} & 0.865 \stdvalue{0.281} & 0.889 \stdvalue{0.196} \\
    GPT-4o & 0.317 \stdvalue{0.466} & 0.229 \stdvalue{0.426} & 0.637 \stdvalue{0.266} & \textbf{0.964 \stdvalue{0.164}} & \textbf{0.978 \stdvalue{0.099}} \\

    \midgrayline

    Claude 3 Haiku & 0.120 \stdvalue{0.429} & 0.069 \stdvalue{0.388} & 0.406 \stdvalue{0.344} & 0.828 \stdvalue{0.270} & 0.809 \stdvalue{0.234} \\
    Claude 3 Sonnet & 0.272 \stdvalue{0.429} & 0.046 \stdvalue{0.423} & 0.533 \stdvalue{0.290} & 0.916 \stdvalue{0.204} & 0.893 \stdvalue{0.195} \\
    Claude 3 Opus & 0.509 \stdvalue{0.457} & 0.381 \stdvalue{0.451} & \textbf{0.688 \stdvalue{0.342}} & 0.941 \stdvalue{0.204} & 0.957 \stdvalue{0.131} \\
    Claude 3.5 Sonnet & \textbf{0.610 \stdvalue{0.507}} & \textbf{0.440 \stdvalue{0.501}} & 0.662 \stdvalue{0.492} & 0.885 \stdvalue{0.286} & 0.932 \stdvalue{0.159} \\
    
    \midgrayline

    Gemini 1.5 Flash & 0.108 \stdvalue{0.451} & 0.115 \stdvalue{0.412} & 0.429 \stdvalue{0.362} & 0.842 \stdvalue{0.274} & 0.838 \stdvalue{0.225} \\

    Gemini 1.5 Pro & 0.475 \stdvalue{0.435} & 0.165 \stdvalue{0.463} & 0.587 \stdvalue{0.326} & 0.900 \stdvalue{0.212} & 0.875 \stdvalue{0.205} \\

    \multicolumn{6}{>{\columncolor{mygray}}c}{\textit{Open-source LLMs}}\\ \midgrayline
    Gemma-2B & -0.021 \stdvalue{0.412} & 0.001 \stdvalue{0.410} & -0.002 \stdvalue{0.245} & 0.502 \stdvalue{0.190} & 0.468 \stdvalue{0.083} \\ 
    Gemma-7B& -0.006 \stdvalue{0.392} & 0.016 \stdvalue{0.389} & 0.085 \stdvalue{0.256} & 0.575 \stdvalue{0.203} & 0.484 \stdvalue{0.122} \\
    \midgrayline
    LLaMA2-7B & -0.018 \stdvalue{0.406} & 0.001 \stdvalue{0.412} & -0.029 \stdvalue{0.261} & 0.477 \stdvalue{0.200} & 0.475 \stdvalue{0.092} \\

    LLaMA2-13B & -0.000 \stdvalue{0.411} & 0.026 \stdvalue{0.417} & 0.072 \stdvalue{0.256} & 0.560 \stdvalue{0.197} & 0.480 \stdvalue{0.109} \\
    LLaMA2-70B &  0.012 \stdvalue{0.409} & 0.010 \stdvalue{0.434} & 0.234 \stdvalue{0.349} & 0.707 \stdvalue{0.271} & 0.629 \stdvalue{0.215} \\
 
    \midgrayline
    Phi-3 Mini (3.8B)& 0.135 \stdvalue{0.431} & 0.012 \stdvalue{0.393} & 0.300 \stdvalue{0.336} & 0.740 \stdvalue{0.275} & 0.659 \stdvalue{0.222} \\

    Phi-3-Small (7.4B) & 0.092 \stdvalue{0.443} & 0.204 \stdvalue{0.422} & 0.347 \stdvalue{0.348} & 0.753 \stdvalue{0.254} & 0.672 \stdvalue{0.220} \\
    Phi-3 Medium (14B) & -0.056 \stdvalue{0.441} & 0.154 \stdvalue{0.406} & 0.356 \stdvalue{0.367} & 0.801 \stdvalue{0.286} & 0.801 \stdvalue{0.230}  \\
    \midgrayline
    LLaMA3-8B &  0.030 \stdvalue{0.444} & 0.139 \stdvalue{0.436} & 0.273 \stdvalue{0.387} & 0.712 \stdvalue{0.285} & 0.639 \stdvalue{0.217} \\
    LLaMA3-70B &  \textbf{0.357 \stdvalue{0.469}} & \textbf{0.343 \stdvalue{0.419}} & \textbf{0.586 \stdvalue{0.415}} & \textbf{0.887 \stdvalue{0.274}} & \textbf{0.923 \stdvalue{0.177}}  \\
    \multicolumn{6}{>{\columncolor{mygray}}c}{\textit{Random}}\\ \midgrayline
    Random & -0.003 \stdvalue{0.409} & 0.005 \stdvalue{0.406} & -0.008 \stdvalue{0.249} & 0.496 \stdvalue{0.192} & 0.467 \stdvalue{0.077} \\ 
  \bottomrule
\end{tabular}

%% file: sections/analysis.tex
\newlength{\radarwidth}
\setlength{\radarwidth}{0.23\textwidth} 
\newcommand{\trimLeft}{0.2cm}
\newcommand{\trimBottom}{0.2cm}
\newcommand{\trimRight}{0.2cm}
\newcommand{\trimTop}{0.2cm}

\begin{figure*}[htp]
    \centering
        \includegraphics[width=\radarwidth]{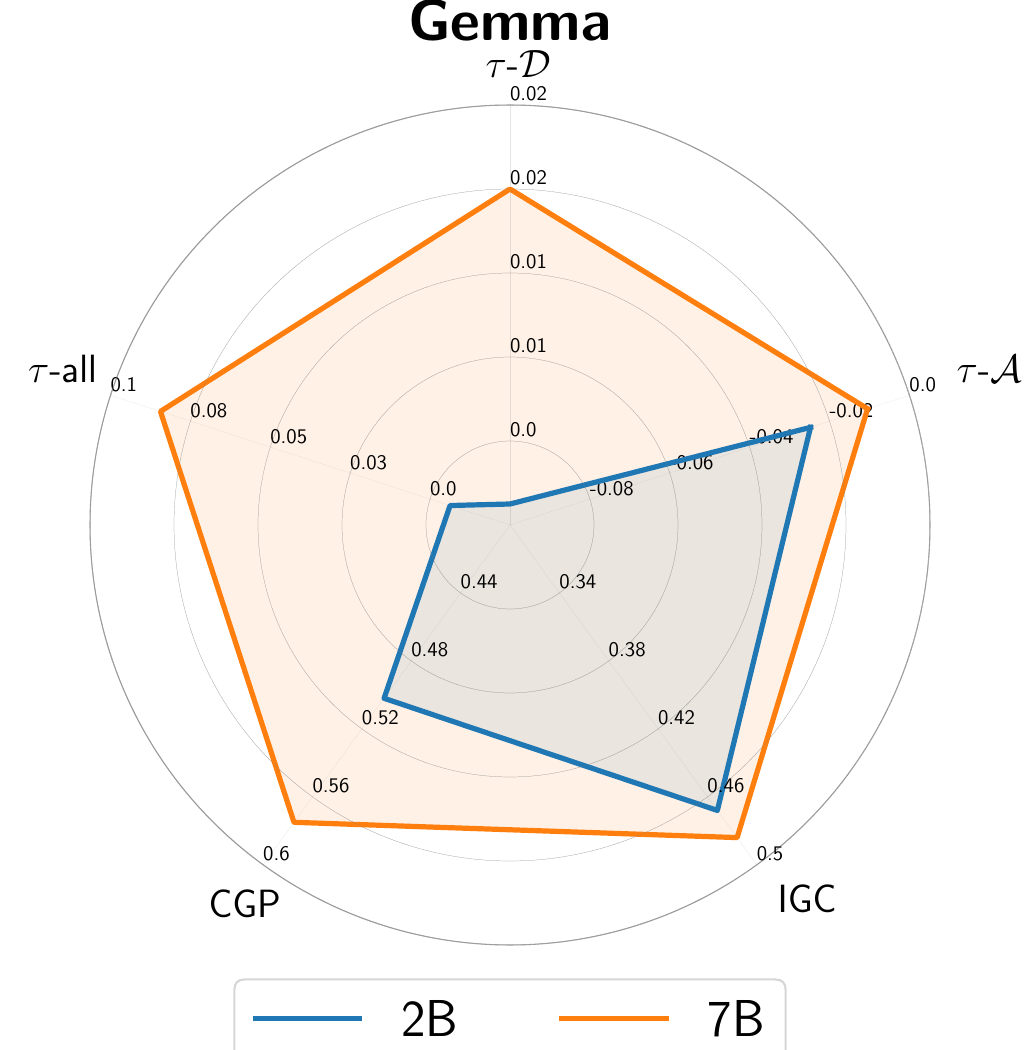}
    \hfill
        \includegraphics[width=\radarwidth]{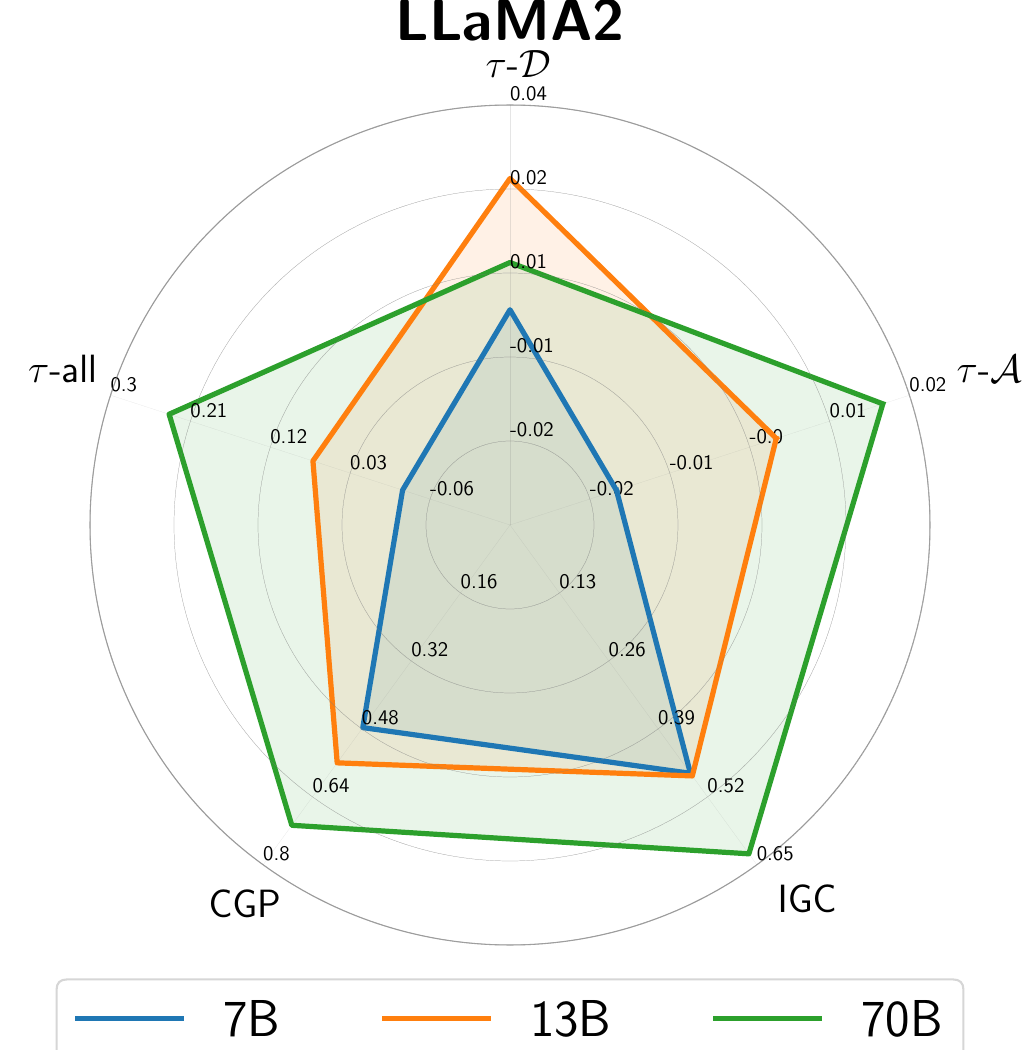}
    \hfill
        \includegraphics[width=\radarwidth]{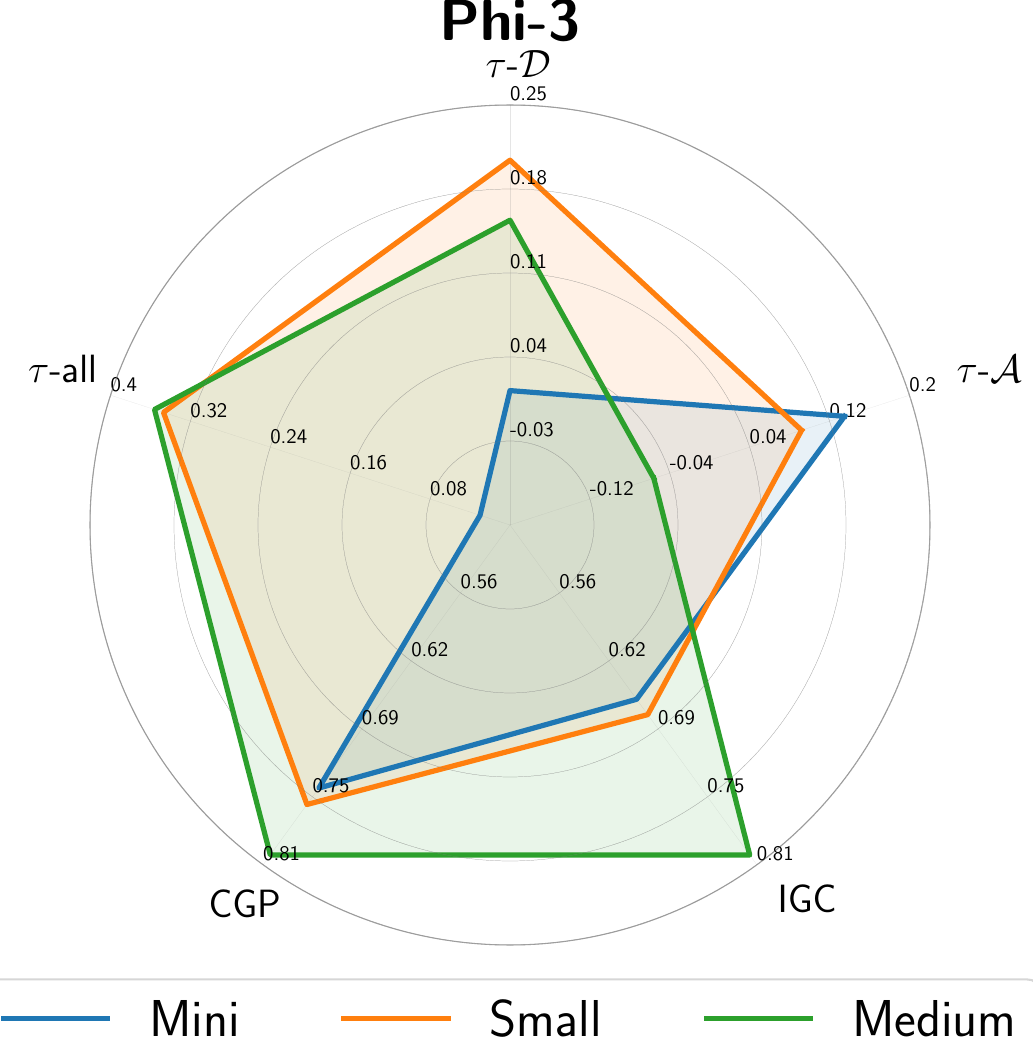}
    \hfill
        \includegraphics[width=\radarwidth]{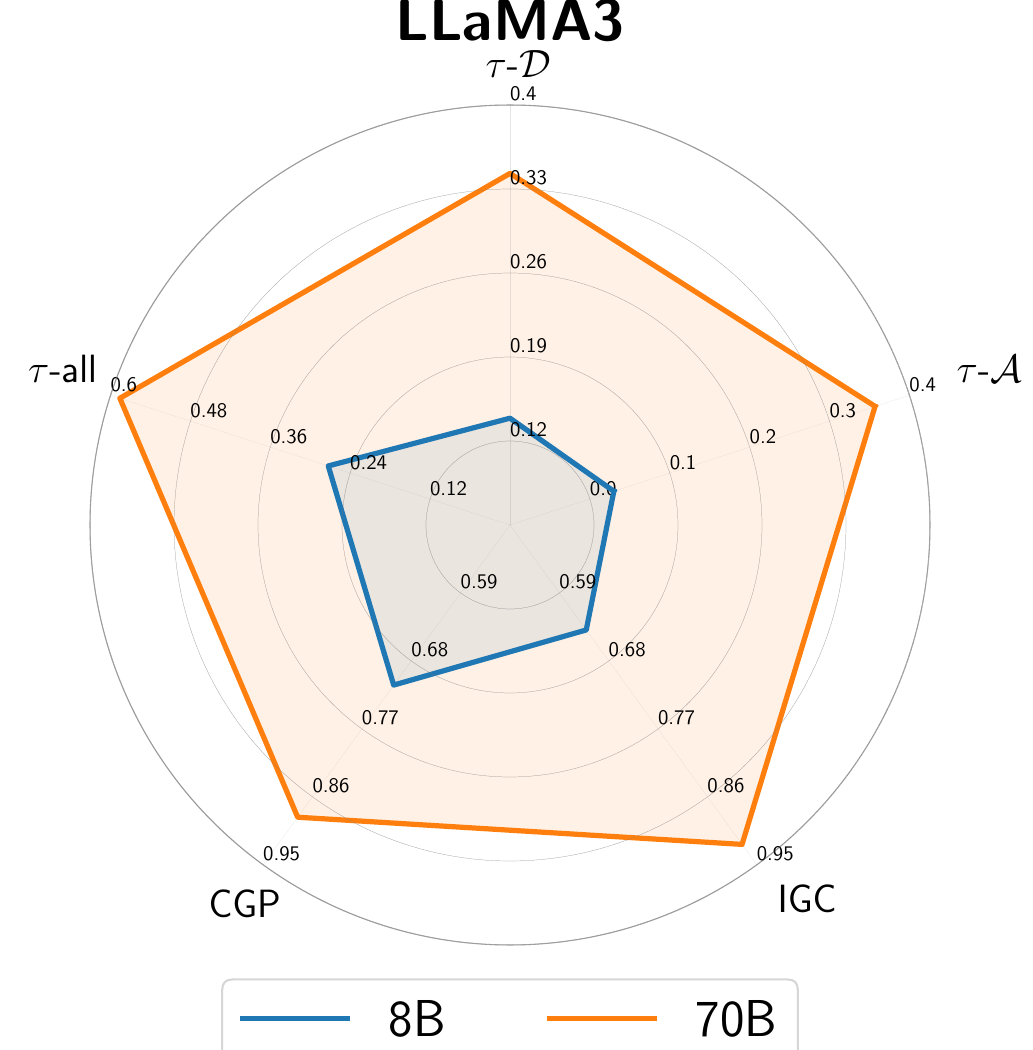}
        \caption{Radar charts comparing the performance of various LLM architectures and sizes~(Gemma, LLaMA2,  Phi-3, and LLaMA3) in maintaining causal epistemic consistency. Each color of the radar plot lines represents a different model size. }
    \label{fig:model_scale}
\end{figure*}

\subsection{Experimental Setup} \label{sec:empirical:setup}
\SmallHeading{Foundational Dataset} To ensure the defeasibility of causal pairs, allowing models to generate intermediates with varying polarity and intensity, we utilize the test dataset of $\delta$-CAUSAL~\cite{cui-etal-2024-exploring} as our foundational dataset, which comprises 1,970 defeasible cause-effect pairs.

\SmallHeading{Three-Phase Assessment for LLMs' Causal Epistemic Consistency} 
There are three main phases in our experiments: 
(i) \ul{Intermediate generation}: We provide LLMs with a single cause-effect pair and two preliminary intermediates: one supporting and one defeating. For each supporter and defeater, we instruct the LLMs to generate two weaker and two stronger intermediates. As a result, we compile a total of 10 intermediates as sequence $\mathcal{I}$, divided into two subsequences: subsequence $\mathcal{D}$ comprised of $m=5$ intermediates that challenge the cause-effect relationship with differing intensities; and subsequence $\mathcal{A}$ consisting of $n=5$ supporting intermediates that reinforce the cause-effect pair, also with varying intensities. The prompt for generating these fine-grained intermediates is presented in \Figure~\ref{fig:pair_prompt_intermediate_generation}; 
(ii) \ul{Intermediate ranking}: From these generated intermediates, we use the same LLM to rank the intermediates to identify their polarities~(supporting or defeating) and intensity. The prompt for ranking these fine-grained intermediates is presented in \Figure~\ref{fig:prompt_intermediate_ranking}; and
(iii) \ul{Evaluation}: Based on the actual order of generated intermediates in the first phase and the predicted ranking order in the second phase, we evaluate the causal epistemic consistency from the perspectives of Intensity Ranking Concordance~(\tauA, \tauD, \tauOverall), Cross-Group Position~(CGP) agreement, and Intra-Group Clustering~(IGC). 

\SmallHeading{Backbone Models} We assess a comprehensive suite of LLMs for causal epistemic consistency. Our evaluation includes: (i) 11 Closed-source models: GPT-3.5 Turbo, GPT-4, GPT-4 Turbo, GPT-4o, GPT-4o mini, Claude 3~(Haiku, Sonnet, and Opus), Claude 3.5~(Sonnet)~\cite{anthropic2024claude3}, Gemini 1.5~(Flash and Pro)~\cite{geminiteam2024gemini15unlockingmultimodal}; (ii) 10 Open-source models: Gemma~(2B and 7B)~\cite{mesnard-etal-2024-gemma}, LLaMA2~(7B, 13B, and 70B)~\cite{touvron-etal-2023-llama}, Phi-3~(mini, small, and medium)~\cite{abdin2024phi}, and LLaMA3~(8B and 70B)~\cite{meta2024introducing}.

\subsection{Experimental Results} \label{sec:empirical:results}

Table~\ref{tab:full_results} presents a quantitative comparison of different models on causal epistemic consistency. 
\begin{itemize}
    \item \textbf{Closed-source models generally outperform open-source models}: For instance, GPT-4o achieves a \tauOverall score of 0.632, a \CGP score of 0.962, and an IGC score of 0.973, whereas LLaMA3-70B, the best-performing open-source model, only achieves a \tauOverall score of 0.586, a \CGP score of 0.887, and an IGC score of 0.923. 
    \item \textbf{Maintaining consistency in intensity is more challenging than achieving consistency in polarity and clustering}: The patterns across different metrics are consistent among different models, suggesting that while LLMs can effectively maintain consistency over differentiating between supporting and defeating intermediates and clustering intermediates of the same polarity together, they find it more challenging to maintain consistent intensity rankings. Namely, achieving consistency over the nuances of causal intensity remains difficult. 
\end{itemize}

\subsection{Does a Larger Model Scale Mean Better Causal Epistemic Consistency?}  \label{sec:empirical:scale}
Previous works~\cite{kaplan-etal-2020-scaling,hoffmann-etal-2022-training} have shown that with the increase in model scale, the improvement in performance follows a power-law relationship. 
However, the effectiveness of `just scaling' for general causal understanding, especially in the context of causality, has become a subject of intense debate~\cite{zecevic-etal-2023-causal}.  

Inspired by this question, we investigate whether increasing the model scale improves the causal epistemic consistency of LLMs. Since this model scale study is only possible for models available in multiple sizes, we conduct experiments with: (i) Gemma at sizes of 2B and 7B; (ii) LLaMA2 at sizes of 7B, 13B, and 70B; (iii) Phi-3 at sizes of 3.8B, 7B, and 14B; and (iv) LLaMA3 at sizes of 8B and 70B.
The experimental results are presented in \Figure~\ref{fig:model_scale}. 
From these results, we clearly observe that \textbf{an increase in model size generally enhances causal epistemic consistency}. For instance, LLaMA2 and LLaMA3 demonstrate significant improvements at larger scales, particularly at 70B, where the causal epistemic consistency scores are notably higher compared to their smaller-scale counterparts. 

\subsection{Visualization of Causal Epistemic Consistency} 
\begin{figure}[htp!]
\centering
\resizebox{0.99999\linewidth}{!}{
\includegraphics[width=0.5\linewidth]{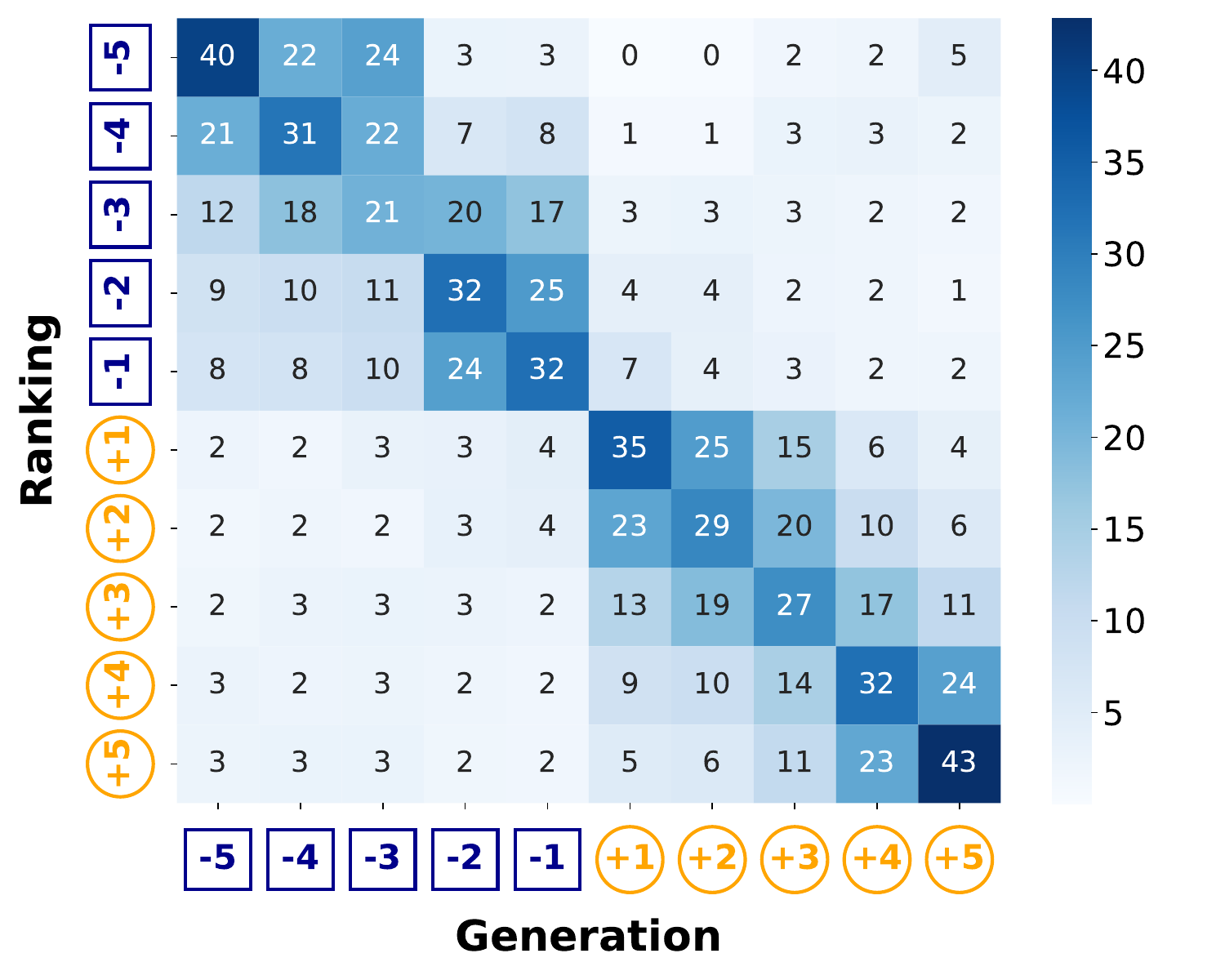} 
\includegraphics[width=0.5\linewidth]{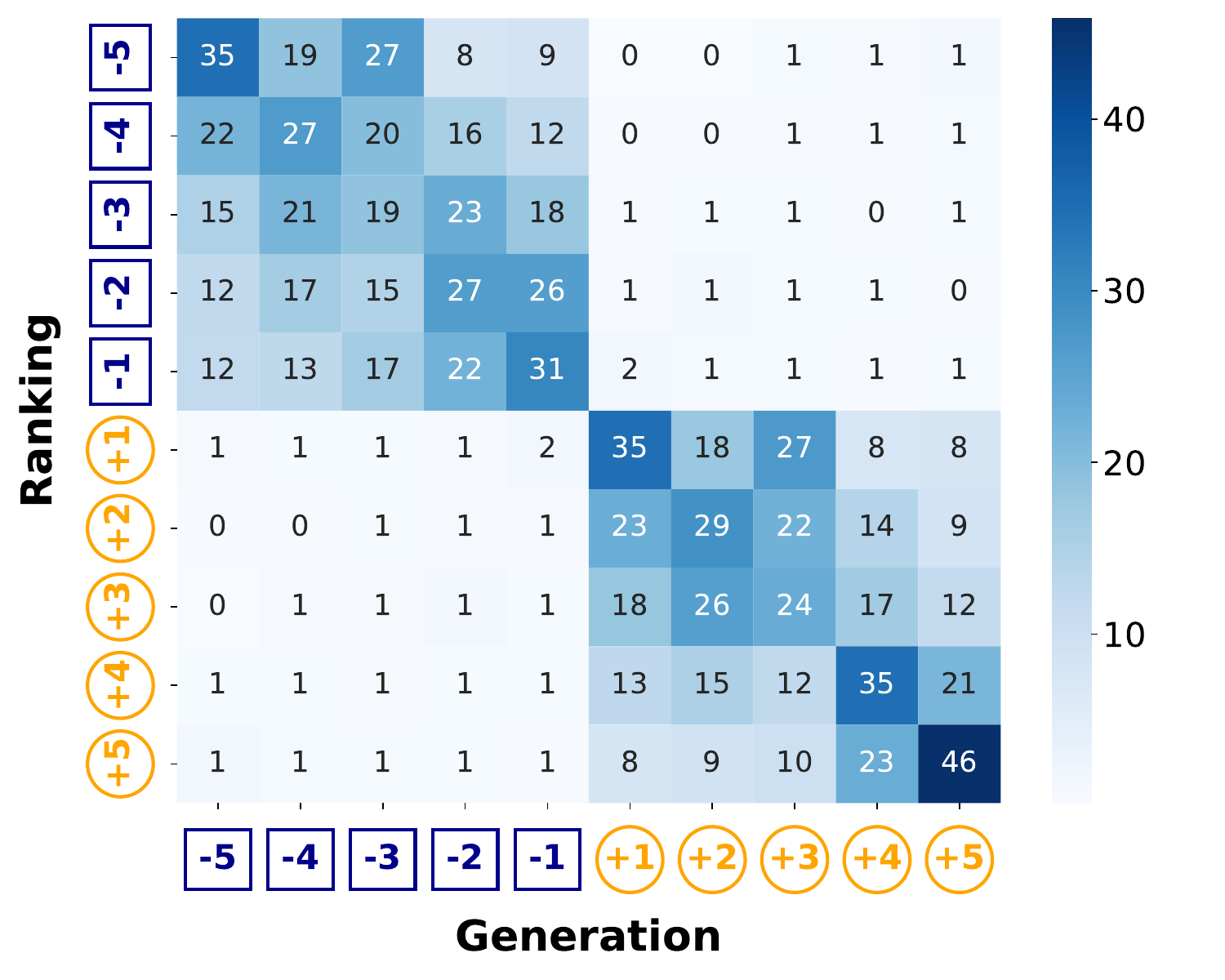}

}
\caption{
Visualization of LLaMA3-70B's~(left) and GPT-4o's~(right) alignment of intermediates' predicted ranking versus their generation phase ranking, indicating the models' self-consistency in intensity, polarity, and clustering. 
Each matrix element $(i,j)$ indicates the percentage of instances where an intermediate ranked at position $i$ during the generation phase was ranked at position $j$ during the ranking phase. For example, (\squarenum{-3}, \circlenum{+4}) indicates the percentage of instances with a label of defeater with an intensity of 3 in the generation phase that was ranked as the supporter with an intensity of 4 during the ranking phase. 
}
\label{fig:confusion_matrix}
\end{figure}

We plot the causal epistemic consistency matrices of LLaMA3-70B and GPT-4o
in \Figure~\ref{fig:confusion_matrix}. 
In these matrices, the x-axis from left to right and the y-axis from top to bottom correspond to \squarenum{-5}\squarenum{-4}\squarenum{-3}\squarenum{-2}\squarenum{-1}\circlenum{+1}\circlenum{+2}\circlenum{+3}\circlenum{+4}\circlenum{+5}, where the square symbol \squarenum{-*} represents defeaters while the circle symbol \circlenum{+*} represents supporters. The numbers inside the symbols indicate the supporting or defeating intensity, with larger absolute values signifying stronger intensity (i.e., \squarenum{-5} is the strongest defeater and \circlenum{+5} is the strongest supporter). 
These matrices visualize how well the models maintain causal epistemic consistency by comparing the labels of intermediates of the generation phase with the predicted labels in the ranking phase.

The confusion matrices of other models are presented in \Appendix~\ref{appendix:results}. 
From the results of the best closed-source and open-source models, we have the following observations: 
\begin{itemize}
    \item \textbf{Diagonal Dominance}: Higher values along the diagonal indicate better causal epistemic consistency. This dominance shows that the model often maintains consistency in both polarity and intensity by correctly matching the labels of intermediates from the generation phase to the ranking phase.
    \item \textbf{Off-Diagonal Elements}: These off-diagonal elements represent the number of instances where the predicted labels during the ranking phase diverge from the labels during the generation phase. Higher values in these cells suggest cases where the model struggles to maintain consistency. For instance, a higher value far from the diagonal indicates a more significant discrepancy between the ranking and generation phases, reflecting lower causal epistemic consistency due to overestimation or underestimation of the intensity of the generated intermediates.
    \item \textbf{Cluster Separation}: These matrices also indicate that these two models cluster supporting and defeating intermediates well, as shown by lower values in the lower left and upper right corners.
\end{itemize}

%% file: sections/correlation.tex
\begin{figure*}[htp]
    \centering
    \resizebox{0.93\linewidth}{!}{
        \includegraphics{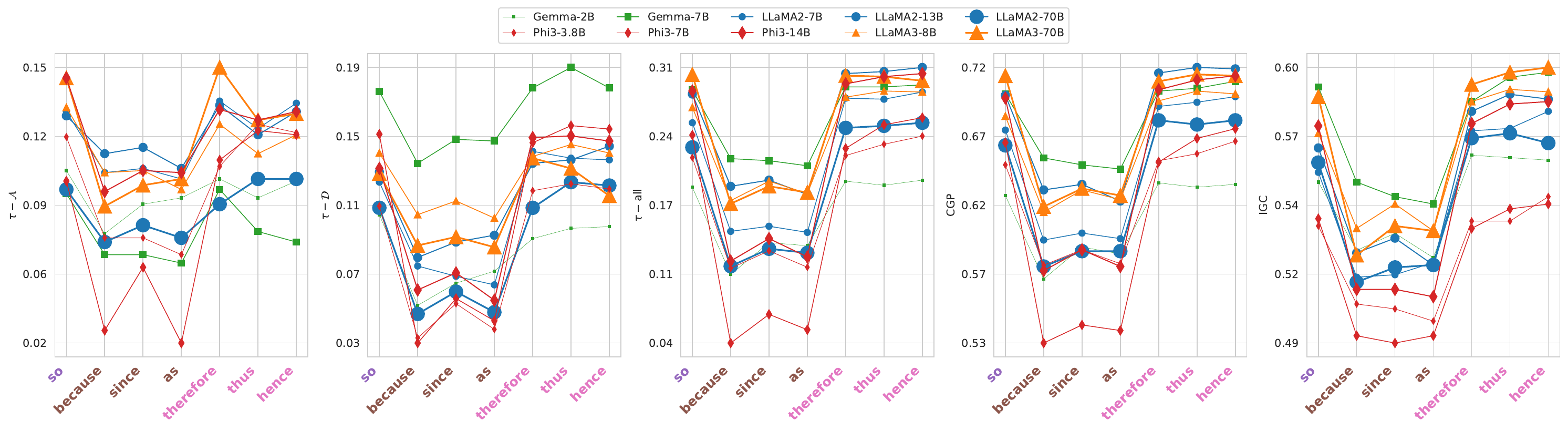}
    }
    \caption{Impact of various conjunction words on the causal epistemic consistency across different LLMs. 
    The x-axes categorize conjunction words into \textcolor[RGB]{141, 105, 184}{coordinating conjunctions}, \textcolor[RGB]{132, 88, 78}{subordinate conjunctions}, and \textcolor[RGB]{213, 125, 190}{conjunctive adverbs}. The y-axes display values for causal epistemic consistency metrics.
    The analysis encompasses diverse model types (distinguished by marker color and shape) at different scales (represented by line thickness and marker size).
    }
    \label{fig:conjunctions}
\end{figure*}
This section explores using internal token probability as an alternative to the prompting method in \Section~\ref{sec:empirical} for maintaining causal epistemic consistency. 
\subsection{Internal Token Probability} \label{sec:correlation:estimation} 
Internal token probability has proven to be a reliable indicator for sequence correlation estimation~\cite{malinin-gales-2021-uncertainty,farquhar-etal-2024-detecting,cui-etal-2024-exploring}. 
For each cause-effect pair $(C, E)$ and any supporting or defeating intermediate $\mathcal{I}_j$, we utilize the token probabilities $p$ to estimate the causal strength $\CS{C\oplus \mathcal{I}_j}{E}$ in \Section~\ref{sec:task:problem}:
\begin{equation} \label{eq:conditional_probability}
    \CS{C\oplus \mathcal{I}_j}{E} =  \prod_{i} p(E_i \vert C\oplus \mathcal{I}_j, w, E_{<i})
\end{equation}
where $E_i$ is the $i_{\text{th}}$ token of $E$ and $E_{<i}$ is the first $i-1$ tokens of $E$.
$p(E_i \vert C\oplus \mathcal{I}_j, w, E_{<i})$ is the internal (conditional) token probability. 
The conjunction word $w$ connects the combination of the cause and the intermediate to the effect, and \ul{explicitly indicates} the causation such as ``because'' and ``therefore''.

\subsection{Experimental Setup} \label{sec:correlation:setup} 
\SmallHeading{Models and Datasets} As closed-source models often do not provide a \code{logprob} API usage~\footnote{Even though \code{logprob} is provided, users cannot compute the probability of an arbitrary token given an input. The potential reason might be to avoid model distillation.}, our investigation resorts to open-source LLMs including  Gemma~(2B and 7B)~\cite{mesnard-etal-2024-gemma}, LLaMA2~(7B, 13B, and 70B)~\cite{touvron-etal-2023-llama}, Phi-3~(3.8B, 7B, and 14B), and LLaMA3~(8B and 70B).  We use the same foundation dataset described in \Section~\ref{sec:empirical:setup}.

\SmallHeading{Three-Phase Assessment} The experiment in this section involves three phases: 
(i) \ul{Intermediate generation}: This phase involves generating a sequence of intermediates, $\mathcal{I}$, following the same procedure described in \Section~\ref{sec:empirical:setup}; 
(ii) \ul{Intermediate ranking based on conditional token probability}: In this phase, we calculate the causal strength based on the conditional token probability using $\left\{ \CS{C\oplus \mathcal{I}_j}{E} \vert \mathcal{I}_j \in \mathcal{I}\right\}$.
(iii) \ul{Evaluation}: We assess the models' causal epistemic consistency using rankings from the generation phase and conditional probability values, based on the proposed metrics in \Section~\ref{sec:metrics}.

\SmallHeading{Conjunction Word Choices} We study multiple conjunction words, including (i) coordinating conjunctions~\cite{grammarlyFANBOYSCoordinating}: ``so''; (ii) subordinate conjunctions~\cite{grammarlyWhatSubordinating}:  ``because'', ``since'', and ``as''; and (iii) conjunctive adverbs~\cite{grammarlyConjunctiveAdverbs}: ``therefore'', ``thus'', and ``hence''.

\subsection{Results and Discussion} \label{sec:correlation:results}

We analyze the results from two aspects: (i) the impact of conjunction words on models' causal epistemic consistency; and (ii) the efficacy of internal token probability against the prompting strategy.

\SmallHeading{Comparison of Different Conjunction Words}
We present the impact of different conjunction words on models' causal epistemic consistency, with distinctions highlighted by varying colors on the x-axis labels in \Figure~\ref{fig:conjunctions}. A consistent trend is observed across different models and causal epistemic consistency metrics. Specifically, coordinating conjunctions (``so'') and conjunctive adverbs (``therefore'', ``thus'', ``hence'') yield better results, while subordinate conjunctions (``because'', ``since'', ``as'') underperform. We posit that placing subordinate conjunctions at the beginning of sentences aligns poorly with the natural language patterns seen by LLMs, potentially degrading performance.

\begin{figure}[htp!]
\centering
\resizebox{0.89\linewidth}{!}{
\includegraphics{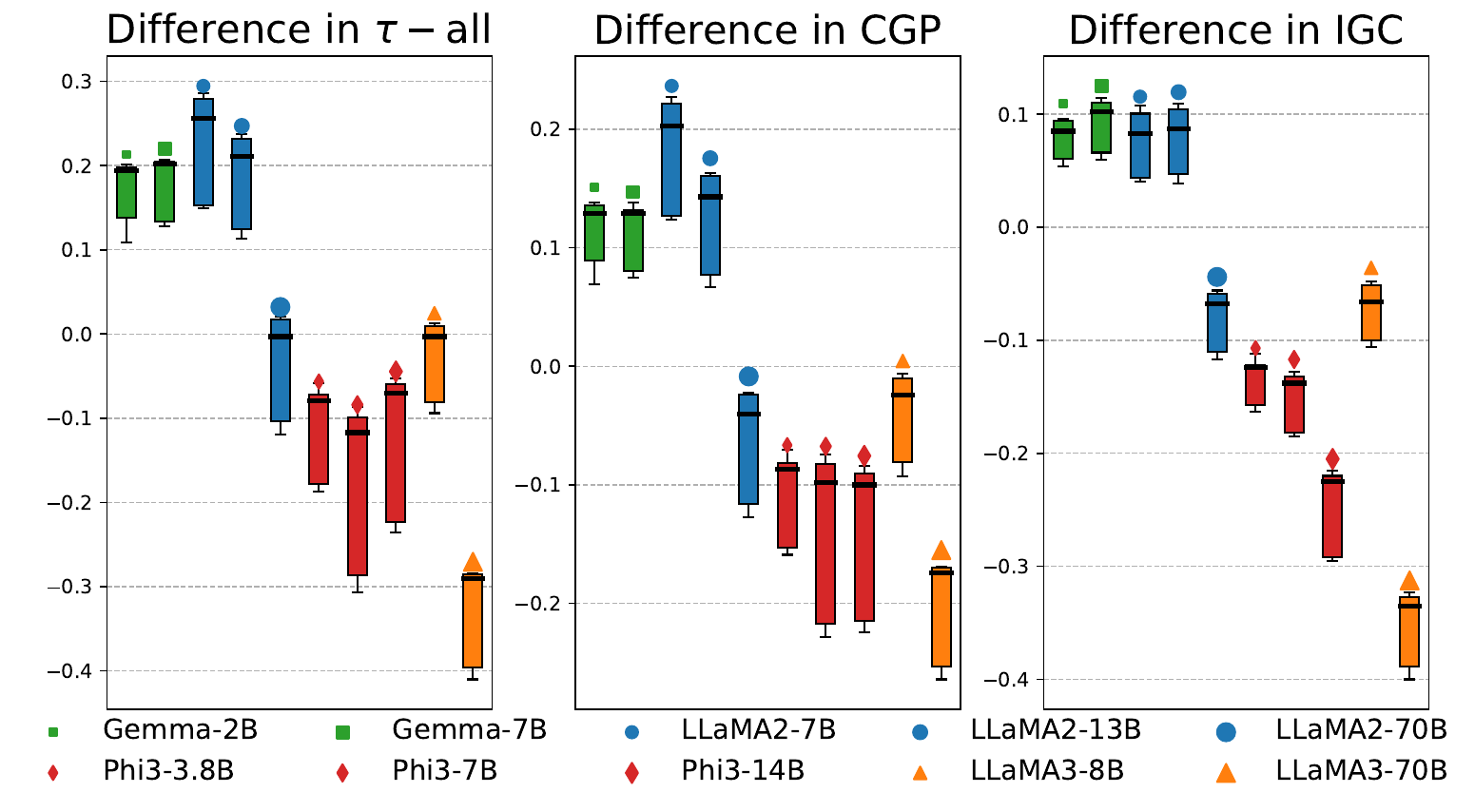}
}
\caption{  
Relative differences in three metrics (\tauOverall, \CGP, and \IGC) for various models using internal token probabilities versus prompting methods. Each box plot represents the distribution of differences for a model across different conjunction words. Markers above the box plots indicate models, aligned with the legend at the bottom.
The marker shape and color indicate the model type, and the marker size represents the model size.
}
\label{fig:prompting_vs_conjunction} 
\end{figure}

\SmallHeading{Comparison with Prompting} 
In Figure~\ref{fig:prompting_vs_conjunction}, we compare the efficacy of internal conditional token probability for evaluating causal epistemic consistency with that of prompting-based strategies. 
We present the relative difference in the three most representative metrics~(\tauOverall, \CGP, and \IGC) for various models~(Gemma, LLaMA2, Phi-3, and LLaMA3) when compared against the prompting aspect. Each subplot corresponds to one of the metrics, showing the differences for each model. Each model is represented by a box plot, calculated from differences given various conjunctions~(``so'', ``because'', ``since'', ``as'', ``therefore'', ``thus'', and ``hence''). Notably, the Gemma model and medium-sized LLaMA2 models (7B, 13B) exhibit enhanced performance under the internal token probability method compared with prompting methods, indicating the effectiveness of internal token probability strategy on some models.

%% file: sections/related_work.tex
\SmallHeading{LLMs and Causality} The investigation of LLMs in understanding and generating causal relations has garnered increasing attention. Previous studies often criticize LLMs for their propensity to inaccurately identify and comprehend the complex causal patterns among these facts~\cite{jin-etal-2024-can,li2024lookwithinllmshallucinate,zecevic-etal-2023-causal,cui-etal-2024-odyssey}. 
Our study further contributes to this discourse by evaluating LLMs' self-consistency in reasoning about fine-grained intermediates in causality and by providing metrics and empirical evidence for LLMs' causal epistemic consistency.

\SmallHeading{Defeasibility in Causal Reasoning} 
Our study of fine-grained intermediates in causality extends the research initiated by $\delta$-CAUSAL~\cite{cui-etal-2024-exploring}, which introduced the concepts of defeaters and supporters in causal analysis. While $\delta$-CAUSAL provided a foundational framework for understanding causal defeasibility, it did not delve into the granularity necessary for nuanced causal reasoning. 
Our research advances this field by moving beyond the binary classification of intermediates as simply supporting or opposing. We refine the categorization of intermediates by considering both their polarity stance~(supporting or opposing) and the intensity of their influence. This nuanced approach enhances the precision of causal analysis, enabling more reliable predictions in complex AI systems.

\SmallHeading{Hallucination of LLMs}  LLMs suffer from generating nonsensical, fallacious, and undesirable content, known as hallucinations~\cite{huang-etal-2023-survey,mouchel-etal-2024-logical,cui-etal-2024-unveiling}. 
The most pertinent hallucination to causal epistemic consistency is the self-contradictory hallucination~\cite{mundler-etal-2024-selfcontradictory}, which means that LLMs generate two contradictory sentences given the same context. 
Specifically, our study on causal epistemic consistency investigates whether the causal intermediates generated by an LLM at various intensities contradict the ones ranked by the same LLM, similar to self-contradictory hallucinations.
However, our study is distinctive in that we focus on the discrepancies between the causal intermediate generation and differentiating behaviors of LLMs, rather than the inconsistencies within the generated text.
Additionally, our task focuses on self-consistency from a causal perspective, including the polarity~(either supporting or defeating) and the intensity of these nuanced intermediates. 

%% file: sections/conclusion.tex
In conclusion, this study introduces causal epistemic consistency as a crucial framework for assessing the self-consistency of LLMs in distinguishing fine-grained causal intermediates. Supported by a novel suite of evaluation metrics, our comprehensive empirical analysis of 21 LLMs reveals significant limitations in their ability to maintain this consistency. This research addresses a critical gap in the understanding of complex causal reasoning and lays the foundation for the development of more self-consistent models capable of handling intricate causal relationships.

%% file: appendices/conjunctions.tex
In the context of causal epistemic consistency, the choice of conjunction words can significantly influence the interpretation of cause-effect relationships. Conjunctions serve as linguistic bridges that connect causes and effects, helping to clarify the nature and strength of these relationships. This section details the types of conjunctions used in our experiments and their implications for causal reasoning. 
Conjunctions that indicate causation can be broadly categorized into three types: 
\begin{enumerate}
    \item Coordinating conjunctions: These conjunctions are words that connect two or more clauses of the same grammatical types~\cite{grammarlyFANBOYSCoordinating}.  ``For'' and ``so'' are noteworthy because they usually indicate a causal relationship between two clauses. 
    \item Subordinating conjunctions: This type of conjunction links a dependent clause to an independent clause~\cite{grammarlyWhatSubordinating}. ``Because'', ``since'', and ``as'' signify a causal relationship that the dependent clause is the cause of the independent clause. 
    \item Conjunctive adverbs: These adverbs or adverb phrases connect two independent clauses by indicating their relationship~\cite{grammarlyConjunctiveAdverbs}. ``Therefore'', ``thus'', and ``hence'' are common adverbs that indicate a causal relationship.  
\end{enumerate}
The typical usages of these conjunctions are presented in \Table~\ref{tab:causal_connectives}. 
\begin{table}[htb!]
    \centering
    \resizebox{0.5\textwidth}{!}{
    \begin{tabular}{p{2cm}p{4.4cm}C{2.2cm}} 
        \toprule 
        \textbf{Conjunction} & \textbf{Usage} & Applicable to autoregressive LLMs\\
        \midrule
        \rowcolor{mygray} \multicolumn{3}{c}{\textit{Coordinating conjunctions}}\\
        For &  \{effect\}, for \{cause\} & \myCrossMark \\
        So &  \{cause\}, so \{effect\} & \myCheckMark\\ 
        \rowcolor{mygray} \multicolumn{3}{c}{\textit{Subordinating conjunctions}}\\
        Because & Because \{cause\}, \{effect\} & {\centering \myCheckMark} \\
        Since & Since \{cause\}, \{effect\} & \myCheckMark \\
        As & As \{cause\}, \{effect\} & \myCheckMark \\
        \rowcolor{mygray} \multicolumn{3}{c}{\textit{Conjunctive Adverbs}}\\
        Therefore & \{cause\}; therefore, \{effect\} & \myCheckMark \\
        Thus & \{cause\}; thus, \{effect\} & \myCheckMark \\
        Hence & \{cause\}; hence, \{effect\} & \myCheckMark \\
        \bottomrule 
    \end{tabular}
    }
    \caption{Categorization of causal conjunctions used in the study, detailing their application in conditional probability calculations. }
    \label{tab:causal_connectives}
\end{table}

Though multiple conjunctions signify causality, the autoregressive nature of LLMs restricts our options to the conjunctions where the ``cause'' precedes the ``effect'' in the sentence.

%% file: appendices/setup.tex
\subsection{Configurations for Computing Infrastructure}
The computing infrastructure of our experiments is as follows: the CPU model is an AMD EPYC 7543 32-Core processor. The GPU model is NVIDIA A100-SXM4-80GB. The total memory size is 503GB. The operating system is Ubuntu 20.04.6 LTS (Focal Fossa). The relevant libraries can be found in the \texttt{requirements.txt} file of our attached code supplementary file. 
We list the most essential packages in Table~\ref{tab:tools}. 

\begin{table} 
    \centering
    \resizebox{\textwidth}{!}{
    \begin{tabular} 
    {lllp{6.5cm}} 
       \toprule 
        \textbf{Artifacts} & \textbf{Citation} & \textbf{Link} & \textbf{License}\\ 
        PyTorch & \cite{paszke-etal-2019-pytorch} & \url{https://pytorch.org/} & BSD-3 License\\
        transformers & \cite{wolf-etal-2020-transformers} & \url{https://huggingface.co/docs/transformers/index} & Apache License 2.0\\
        Accelerate & \cite{accelerate} & \url{https://huggingface.co/docs/accelerate/index} & Apache License 2.0\\
        nltk & \cite{bird-loper-2004-nltk} & \url{https://www.nltk.org/} & Apache License 2.0 \\
        numpy & \cite{harris-etal-2020-array} & \url{https://numpy.org/} & BSD License \\
        matplotlib & \cite{hunter-2009-matplotlib} & \url{https://matplotlib.org/} & BSD compatible License\\
        OpenAI API & N/A & \url{https://platform.openai.com/docs/api-reference} & MIT License\\
      \bottomrule 
    \end{tabular}
    }
    \caption{Details of the artifacts we use. }
    \label{tab:tools}
\end{table}

\subsection{Prompt Design of LLMs} \label{appendix:setup:prompt}

\SmallHeading{Generation}  Directly prompting models to generate 10 arguments—five defeaters followed by five supporters—has proven challenging and frequently results in unsatisfactory outputs, requiring multiple attempts for the same cause-effect pairs. 
To address this, we generate supporters and defeaters in a pairwise manner. This involves using the original defeater and supporter from the data and prompting the model four times for each cause-effect pair. The prompts are structured as follows:

\begin{itemize}
\item Generate two weaker defeaters.
\item Generate two stronger defeaters.
\item Generate two weaker supporters.
\item Generate two stronger supporters.
\end{itemize}

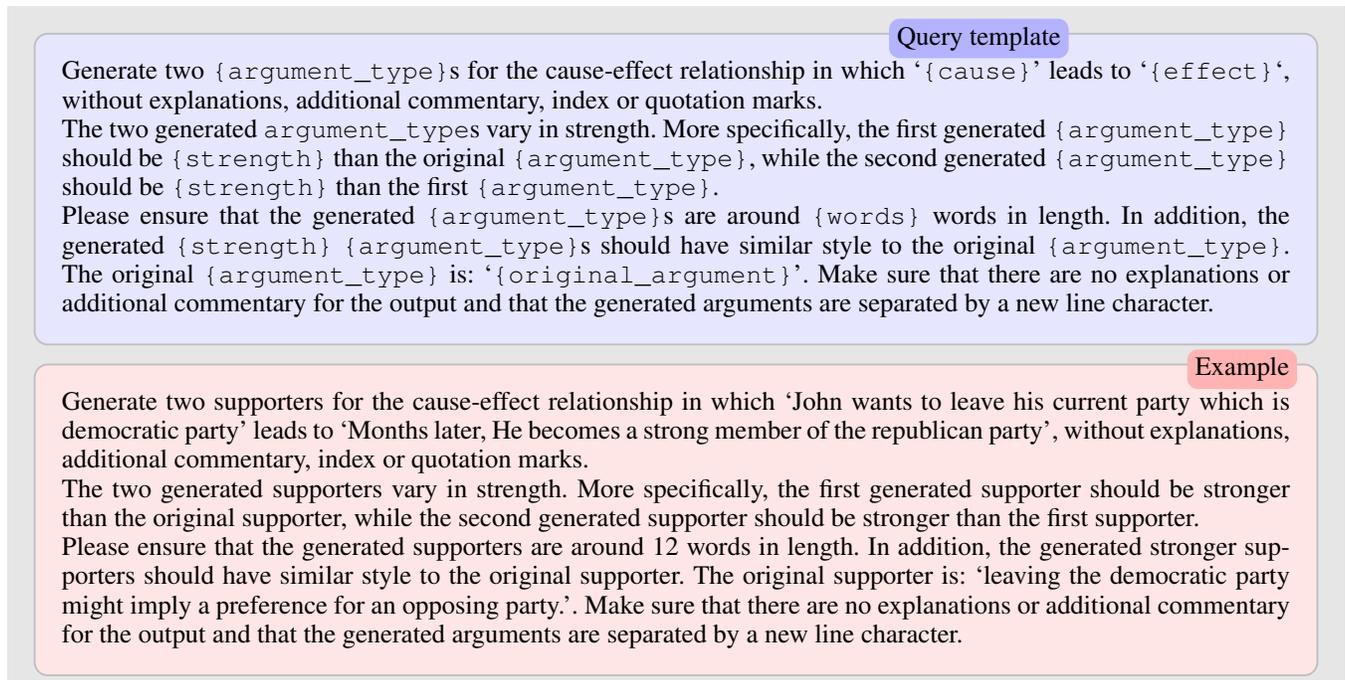
\begin{figure*}[htb]
    \centering\input{figures/prompts/pair_intermediate_generation}
    \caption{Example prompts for generating fine-grained intermediates with LLMs. The \variableP{\{argument\_type\}} is selected from \{defeater, supporters\}, while the \variableP{\{strength\}} is selected from \{stronger, weaker\}. }
    \label{fig:pair_prompt_intermediate_generation}
\end{figure*}

By prompting the model to generate only two intermediates at a time, the text becomes easier to parse. Furthermore, each model exhibits unique output characteristics. For instance, LLaMA often begins with "Sure, here is [...]" before listing arguments. Due to these unique output formats across different models, we provide tailored scripts for each model to ensure consistent and accurate text generation.
The prompt we design to generate these fine-grained intermediates is illustrated in Figure~\ref{fig:pair_prompt_intermediate_generation}. For the hyperparameter, we use the default hyperparameter.   After the prompting, we have a set of intermediates consisting of a subset of supporters, denoted as $\mathcal{A}$, and a subset of defeaters, denoted as $\mathcal{D}$.

\SmallHeading{Ranking} The prompt we use to rank these fine-grained intermediates is present in Figure~\ref{fig:prompt_intermediate_ranking}.  
\begin{figure}[htb]
    \centering
    \input{figures/prompts/intermediate_classification}
    \caption{Example prompts for ranking fine-grained intermediates, guiding LLMs to order arguments based on their influence on the cause-effect relationship.}
    \label{fig:prompt_intermediate_ranking}
\end{figure}

\subsection{Conditional Probability Estimation} \label{appendix:setup:probability} 

\begin{lstlisting}[language=Python] 
from abc import ABC, abstractmethod
import torch

class ConditionalProb(ABC):
    """
    An abstract base class for computing conditional probabilities in different models including
    LLaMA, Mistral, and Gemma.
    """

    def __init__(self, args):
        self.args = args
        self.device = torch.device('cuda' if args.cuda else 'cpu')

        self.base_model = None
        self.tokenizer = None
        self.ignored_id_in_label = None
        self.initialize_model_and_tokenizers()
        self.set_ignored_id_in_label()

        self.base_model.eval()
        self.base_model.to(self.device)

    @abstractmethod
    def initialize_model_and_tokenizers(self):
        """
        Abstract method that must be implemented by all subclasses to set up the model and tokenizers.
        """
        pass

    @abstractmethod
    def set_ignored_id_in_label(self):
        pass

    @torch.no_grad()
    def calculate_conditional_probability(self, input_text, output_text):
        # Before everything start, double-check.
        assert (self.base_model is not None and self.tokenizer is not None and self.ignored_id_in_label is not None), \
            "Make sure the base_model, tokenizers, and ignored_id_in_label well set. "

        print("input_text: {} \noutput_text: {}".format(input_text, output_text))
        
        # Process the input and output text
        input_text = input_text.strip()
        output_text = output_text.strip()
        input_text = input_text + " "  # Add a space at the end
        combined_text = input_text + output_text

        # Combine the input and output text
        combined_inputs = self.tokenizer(combined_text, return_tensors="pt").to(self.device)

        # Create labels, mask the input text part
        labels = combined_inputs['input_ids'].clone()
        labels[labels == self.tokenizer.pad_token_id] = self.ignored_id_in_label  # -100 is ignored in loss computation
        labels[:, :len(self.tokenizer.encode(input_text.strip()))] = self.ignored_id_in_label  # Remove the last space by strip()

        # Compute the loss
        loss = self.base_model(**combined_inputs, labels=labels).loss

        # Compute the probability of the output text
        output_length = len(self.tokenizer.encode(output_text)) - 1 # -1 to remove the first special token
        probability = torch.exp(-loss * output_length)
        ret_dict = {"input_txt": input_text, "output_text": output_text, "conditional_prob": probability}
        return ret_dict
\end{lstlisting}

Please note that the \code{base\_model} should be \texttt{*ForCausalLM}~(\texttt{GemmaForCausalLM}, \texttt{AutoModelForCausalLM}, and \texttt{LlamaForCausalLM}), which is for autoregressive language modelling. This series of models predicts the next token in the sequence given all previous tokens. In other words, the model attends only to the leftward context. 

\SmallHeading{Other formulae} Apart from the conditional probability discussed in Section~\ref{sec:correlation}, alternative approaches exist for estimating the correlation degree between two events. 

The average conditional probability is defined as
\begin{equation}
    \text{P}_{\text{avg}}(x, y) = \frac{\sum^{\vert y \vert}_{i=1} p(y_i \vert x, y_{< i})}{\vert y \vert} 
\end{equation}

\citet{holtzman-etal-2021-surface} introduce domain conditional pointwise mutual information~(PMI) to measure the correlation between $x$ and $y$. 
\begin{equation}
    \text{PMI}_{\text{DC}}(x, y, \text{domain}) = \frac{p(y \vert x, \text{domain})}{p(y \vert \text{domain})} 
\end{equation}

However, both formulations involve scaling--either by the sentence length of $y$ or the sequential probability of $y$. Consequently, these approaches do not alter the conclusion regarding the ranking order of intermediates discussed in \Section~\ref{sec:correlation:estimation}.

%% file: figures/prompts/pair_intermediate_generation.tex
\begin{mdframed}[backgroundcolor=gray!20,linewidth=0pt] 
    \begin{framedwithtag}{Query template}{blue!30}{blue!10}{-45mm}{2mm}{%
    Generate two \variableP{\{argument\_type\}}s for the cause-effect relationship in which ‘\variableP{\{cause\}}’ leads to ‘\variableP{\{effect\}}‘, without explanations, additional commentary, index or quotation marks.

    The two generated \variableP{{argument\_type}}s vary in strength. More specifically, the first generated \variableP{\{argument\_type\}} should be \variableP{\{strength\}} than the original \variableP{\{argument\_type\}}, while the second generated \variableP{\{argument\_type\}} should be \variableP{\{strength\}} than the first \variableP{\{argument\_type\}}.
    
    Please ensure that the generated \variableP{\{argument\_type\}}s are around \variableP{\{words\}} words in length. In addition, the generated \variableP{\{strength\}} \variableP{\{argument\_type\}}s should have similar style to the original \variableP{\{argument\_type\}}. The original \variableP{\{argument\_type\}} is: ‘\variableP{\{original\_argument\}}’. Make sure that there are no explanations or additional commentary for the output and that the generated arguments are separated by a new line character.
    }
    \end{framedwithtag}
    \vspace{-10mm}

    \begin{framedwithtag}{Example}{red!30}{red!10}{-10mm}{2mm}{%
        Generate two supporters for the cause-effect relationship in which `John wants to leave his current party which is democratic party' leads to `Months later, He becomes a strong member of the republican party', without explanations, additional commentary, index or quotation marks. 
        
        The two generated supporters vary in strength. More specifically, the first generated supporter should be stronger than the original supporter, while the second generated supporter should be stronger than the first supporter. 
        
        Please ensure that the generated supporters are around 12 words in length. In addition, the generated stronger supporters should have similar style to the original supporter. The original supporter is: `leaving the democratic party might imply a preference for an opposing party.'. Make sure that there are no explanations or additional commentary for the output and that the generated arguments are separated by a new line character.
    }
    \end{framedwithtag}
\end{mdframed} 

%% file: figures/prompts/intermediate_classification.tex
\begin{mdframed}[backgroundcolor=gray!20,linewidth=0pt] 
    \begin{framedwithtag}{Prompt template}{blue!30}{blue!10}{-45mm}{2mm}{%
    Given a defeasible cause-effect pair and ten arguments with varying strength, please give a ranking of the arguments based on whether they strengthen or weaken the argumentative strength of the cause-effect pair. Note that the ten arguments consist of five supporting arguments and five defeating arguments. The ranking should be in the order from the argument that weakens the argumentative strength of the pair the most to the argument that strengthens the argumentative strength the most. 
    
    In addition, please ensure that the result only contains indices referring to each argument, separated by a single space and without any additional explanation or comments.
    
    The cause is `\{cause\}' and the effect is `\{effect\}'.
    
    The ten arguments are:
    
    1. \{argument\_1\} \newline
    2. \{argument\_2\} \newline
    3. \{argument\_3\} \newline
    4. \{argument\_4\} \newline
    5. \{argument\_5\} \newline
    6. \{argument\_6\} \newline
    7. \{argument\_7\} \newline
    8. \{argument\_8\} \newline
    9. \{argument\_9\} \newline
    10. \{argument\_10\}
    }
    \end{framedwithtag}
    \vspace{-10mm}
    \begin{framedwithtag}{Example}{red!30}{red!10}{-10mm}{2mm}{%
        Given a defeasible cause-effect pair and ten arguments with varying strength, please give a ranking of the arguments based on whether they strengthen or weaken the force of reasons of the cause-effect pair. Note that the ten arguments consist of five supporting arguments and five defeating arguments. The ranking should be in the order from the argument that weakens the argumentative strength of the pair the most to the argument that strengthens the argumentative strength the most. 
        
        In addition, please ensure that the result only contains indices referring to each argument, separated by a single space and without any addition explanation or comments.
        
        The cause is `John wants to leave his current party which is democratic party' and the effect is `Months later, He becomes a strong member of the republican party'.
        
         The ten arguments are: \\
         1. John is appointed to a nonpartisan governmental position. \\
         2. John decides to become an independent politician.  \\
         3. John changes his mind and runs on the Democratic ticket. \\
         4. leaving the democratic party might imply a preference for an opposing party. \\
         5. abandoning the democratic party strongly indicates a shift towards republican ideals and membership. \\
         6. leaving one party could indicate a desire to join another.  \\
         7. John changes his position to remain an independent voter. \\
         8. John changes his mind and votes for Democratic candidates.  \\
         9. departing the democratic party suggests a likelihood of aligning with the republican opposition.  \\
         10. deciding to leave might show interest in an alternative political group.
    }
    \end{framedwithtag}
\end{mdframed} 

%% file: appendices/metrics.tex
In this section, we first present more discussion for the novel intra-clustering metrics in \Appendix~\ref{appendix:metrics:clustering}, which covers the implication of the polarity changes and more case studies. Additionally, to better understand the difference between these proposed metrics, we explain these metrics with examples in \Appendix~\ref{appendix:metrics:case_study}. 

\subsection{Intra-Group Clustering} \label{appendix:metrics:clustering}

\SmallHeading{Implication of Polarity Changes} \label{appendix:metrics:clustering:implications}
Polarity changes in a sequence often indicate transitions between different states, representing cluster changes. By quantifying these polarity changes as distances, \IGC accurately captures these cluster changes. 
Namely, in the context of sequence clustering, counting polarity changes shifts the focus to transitions rather than mere index differences. 
For example, in a sequence of customer interactions, a transition from browsing items to adding to the shopping cart has a greater impact on cluster formulation. 

Besides, polarity change provides an intuitive measure for evaluating the quality of sequence clustering. A sequence with fewer internal polarity changes is more cohesive, as there are no interruptions within different snippets of the sequences. Conversely, frequent polarity changes suggest that the sequences are more intertwined, indicating that the clusters are not distinctly separated but rather mixed together. It reflects overlapping or intertwined behavioral patterns of these snippets with the sequence. 

In summary, with polarity changes, we can better understand the clustering quality, leading to more meaningful insights from the sequence data.

\subsubsection{Case study examples} \label{appendix:metrics:clustering:case}
We use the following example {\SquareB \SquareB \CircleA \SquareB \SquareB \CircleA \CircleA \CircleA \SquareB \CircleA} to detail the calculation process of \IGC. Based on the clustering distance definition, the distance metrics is 
\begin{equation*}
\begin{vmatrix}
0 & 0 & 1 & 1 & 1 & 2 & 2 & 2 & 2 & 3 \\
0 & 0 & 1 & 1 & 1 & 2 & 2 & 2 & 2 & 3 \\
1 & 1 & 0 & 1 & 1 & 1 & 1 & 1 & 2 & 2 \\
1 & 1 & 1 & 0 & 0 & 1 & 1 & 1 & 1 & 2 \\
1 & 1 & 1 & 0 & 0 & 1 & 1 & 1 & 1 & 2 \\
2 & 2 & 1 & 1 & 1 & 0 & 0 & 0 & 1 & 1 \\
2 & 2 & 1 & 1 & 1 & 0 & 0 & 0 & 1 & 1 \\
2 & 2 & 1 & 1 & 1 & 0 & 0 & 0 & 1 & 1 \\
2 & 2 & 2 & 1 & 1 & 1 & 1 & 1 & 0 & 1 \\
3 & 3 & 2 & 2 & 2 & 1 & 1 & 1 & 1 & 0 \\
\end{vmatrix}
\end{equation*}
The silhouette score for each element in the sequence individually is {\small [$\frac{(1 + 2 + 2 + 2 + 3)/5 - (0 + 1 + 1 + 2)/4}{\max(((1 + 2 + 2 + 2 + 3)/5, (0 + 1 + 1 + 2)/4)}$, $\frac{(1 + 2 + 2 + 2 + 3)/5 - (0 + 1 + 1 + 2)/4}{\max(((1 + 2 + 2 + 2 + 3)/5, (0 + 1 + 1 + 2)/4)}$, $\frac{(1 + 1 + 1 + 1 + 2)/5 - (1 + 1 + 1 + 2)/4}{\max((1 + 1 + 1 + 1 + 2)/5, (1 + 1 + 1 + 2)/4})$, $\frac{(1 + 1 + 1 + 1 + 2)/5 - (1 + 1 + 1 + 2)/4}{\max((1 + 1 + 1 + 1 + 2)/5, (1 + 1 + 0 + 1)/4})$, $\frac{(1 + 1 + 1 + 1 + 2)/5 - (1 + 1 + 1 + 2)/4}{\max((1 + 1 + 1 + 1 + 2)/5, (1 + 1 + 0 + 1)/4})$, $\frac{(2 + 2 + 1 + 1 + 1)/5 - (1 + 0 + 0 + 1)/4}{\max((1 + 1 + 1 + 1 + 2)/5, (1 + 1 + 0 + 1)/4})$, $\frac{(2 + 2 + 1 + 1 + 1)/5 - (1 + 0 + 0 + 1)/4}{\max((1 + 1 + 1 + 1 + 2)/5, (1 + 1 + 0 + 1)/4})$, $\frac{(2 + 2 + 1 + 1 + 1)/5 - (1 + 0 + 0 + 1)/4}{\max((1 + 1 + 1 + 1 + 2)/5, (1 + 1 + 0 + 1)/4})$, $\frac{(2 + 1 + 1 + 1 + 1)/5 - (2 + 2 + 1 + 1)/4}{\max((2 + 1 + 1 + 1 + 1)/5, (2 + 2 + 1 + 1)/4)}$, $\frac{(3 + 3 + 2 + 2 + 1)/5 - (2 + 1 + 1 + 1)/4}{\max((3 + 3 + 2 + 2 + 1)/5, (2 + 1 + 1 + 1)/4)}$]} = [0.5,  0.5, -0.04, 0.375, 0.375, 0.643, 0.643, 0.643, -0.2, -.432], and the final \IGC value is 0.387.

\subsubsection{Optimal Cases} \label{appendix:metrics:clustering:optimal}
An optimal case for \IGC is that {\SquareB\SquareB\SquareB\SquareB\SquareB\CircleA\CircleA\CircleA\CircleA\CircleA}. Following the calculation rule, the distance matrix is: 
\begin{equation*}
\begin{vmatrix}
0 & 0 & 0 & 0 & 0 & 1 & 1 & 1 & 1 & 1 \\
0 & 0 & 0 & 0 & 0 & 1 & 1 & 1 & 1 & 1 \\
0 & 0 & 0 & 0 & 0 & 1 & 1 & 1 & 1 & 1 \\
0 & 0 & 0 & 0 & 0 & 1 & 1 & 1 & 1 & 1 \\
0 & 0 & 0 & 0 & 0 & 1 & 1 & 1 & 1 & 1 \\
1 & 1 & 1 & 1 & 1 & 0 & 0 & 0 & 0 & 0 \\
1 & 1 & 1 & 1 & 1 & 0 & 0 & 0 & 0 & 0 \\
1 & 1 & 1 & 1 & 1 & 0 & 0 & 0 & 0 & 0 \\
1 & 1 & 1 & 1 & 1 & 0 & 0 & 0 & 0 & 0 \\
1 & 1 & 1 & 1 & 1 & 0 & 0 & 0 & 0 & 0 \\
\end{vmatrix}
\end{equation*}
The silhouette score for each element in the sequence individually is [1.0, 1.0, 1.0, 1.0, 1.0, 1.0, 1.0, 1.0, 1.0, 1.0]. And the final \IGC value is 1.0, too.

\SmallHeading{Edge cases} An edge case for the IGC is {\SquareB\SquareB\SquareB\SquareB\SquareB\SquareB\SquareB\SquareB\SquareB\CircleA}. In this case, the distance matrix is: 
\begin{equation*}
\begin{vmatrix}
0 & 0 & 0 & 0 & 0 & 0 & 0 & 0 & 0 & 1 \\
0 & 0 & 0 & 0 & 0 & 0 & 0 & 0 & 0 & 1 \\
0 & 0 & 0 & 0 & 0 & 0 & 0 & 0 & 0 & 1 \\
0 & 0 & 0 & 0 & 0 & 0 & 0 & 0 & 0 & 1 \\
0 & 0 & 0 & 0 & 0 & 0 & 0 & 0 & 0 & 1 \\
0 & 0 & 0 & 0 & 0 & 0 & 0 & 0 & 0 & 1 \\
0 & 0 & 0 & 0 & 0 & 0 & 0 & 0 & 0 & 1 \\
0 & 0 & 0 & 0 & 0 & 0 & 0 & 0 & 0 & 1 \\
0 & 0 & 0 & 0 & 0 & 0 & 0 & 0 & 0 & 1 \\
1 & 1 & 1 & 1 & 1 & 1 & 1 & 1 & 1 & 0 \\
\end{vmatrix}
\end{equation*}

In this case, if we calculate the silhouette score for all elements using $s(i) = \frac{b(i) - a(i)}{\max(a(i), b(i))}$, the silhouette scores for elements in the sequence individually are [1.0, 1.0, 1.0, 1.0, 1.0, 1.0, 1.0, 1.0, 1.0, \ul{0.0}]. For the last elements, the silhouette score is 0, which means this element is on the border of two clusters. However, please note that this element \SquareB is also the only element of its type in the sequence. Motivated by this case, we update the silhouette score calculation as: 
\begin{equation}\label{eq:silhouette_sample_score_appendix}
    s(i) = 
            \begin{cases}
                1 & \text{if $i$ is alone its cluster}, \\
                \frac{b(i) - a(i)}{\max(a(i), b(i))}  & \text{otherwise.}
           \end{cases}
\end{equation}
which considers the edge case when there is only one element inside certain group. With this updated formula, the silhouette scores for all elements are [1.0, 1.0, 1.0, 1.0, 1.0, 1.0, 1.0, 1.0, 1.0, 1.0], which follows our intuition that are the elements are perfectly clustered. Similarly, in the dual case {\CircleA\SquareB\SquareB\SquareB\SquareB\SquareB\SquareB\SquareB\SquareB\SquareB}, the distance matrix is
\begin{equation*}
\begin{vmatrix}
0 & 1 & 1 & 1 & 1 & 1 & 1 & 1 & 1 & 1 \\
1 & 0 & 0 & 0 & 0 & 0 & 0 & 0 & 0 & 0 \\
1 & 0 & 0 & 0 & 0 & 0 & 0 & 0 & 0 & 0 \\
1 & 0 & 0 & 0 & 0 & 0 & 0 & 0 & 0 & 0 \\
1 & 0 & 0 & 0 & 0 & 0 & 0 & 0 & 0 & 0 \\
1 & 0 & 0 & 0 & 0 & 0 & 0 & 0 & 0 & 0 \\
1 & 0 & 0 & 0 & 0 & 0 & 0 & 0 & 0 & 0 \\
1 & 0 & 0 & 0 & 0 & 0 & 0 & 0 & 0 & 0 \\
1 & 0 & 0 & 0 & 0 & 0 & 0 & 0 & 0 & 0 \\
1 & 0 & 0 & 0 & 0 & 0 & 0 & 0 & 0 & 0 \\
\end{vmatrix}
\end{equation*}
The silhouette scores for all elements are [1.0, 1.0, 1.0, 1.0, 1.0, 1.0, 1.0, 1.0, 1.0, 1.0].

However, when the sole member of a group appears in the middle inside another group, such as {\SquareB\SquareB\SquareB\SquareB\CircleA\SquareB\SquareB\SquareB\SquareB\SquareB}, now the distance matrix becomes
\begin{equation*}
\begin{vmatrix}
0 & 0 & 0 & 0 & 1 & 1 & 1 & 1 & 1 & 1 \\
0 & 0 & 0 & 0 & 1 & 1 & 1 & 1 & 1 & 1 \\
0 & 0 & 0 & 0 & 1 & 1 & 1 & 1 & 1 & 1 \\
0 & 0 & 0 & 0 & 1 & 1 & 1 & 1 & 1 & 1 \\
1 & 1 & 1 & 1 & 0 & 1 & 1 & 1 & 1 & 1 \\
1 & 1 & 1 & 1 & 1 & 0 & 0 & 0 & 0 & 0 \\
1 & 1 & 1 & 1 & 1 & 0 & 0 & 0 & 0 & 0 \\
1 & 1 & 1 & 1 & 1 & 0 & 0 & 0 & 0 & 0 \\
1 & 1 & 1 & 1 & 1 & 0 & 0 & 0 & 0 & 0 \\
1 & 1 & 1 & 1 & 1 & 0 & 0 & 0 & 0 & 0 \\
\end{vmatrix}. 
\end{equation*}
The silhouette scores for elements in the sequence individually are [0.38, 0.38, 0.38, 0.38, \ul{1.0}, 0.5, 0.5, 0.5, 0.5, 0.5].

\subsection{Illustration of Proposed Metrics with Examples} \label{appendix:metrics:case_study}
\begin{table}[t]
\resizebox{0.99\linewidth}{!}{
\input{figures/table_minipages/metrics_comparison_via_example}
}
\caption{Case studies illustrating different aspects captured by the proposed metrics: intensity ranking concordance, cross-group position agreement, and intra-group clustering. }
\label{tab:metrics_comparison_via_cases}
\end{table}

We illustrate the proposed metrics with the following cases: 
\begin{itemize}
    \item \textit{Optimal case}: The optimal case is \resizebox{!}{0.8em}{\SquareDNum{100}{-5}\SquareDNum{80}{-4}\SquareDNum{60}{-3}\SquareDNum{40}{-2}\SquareDNum{20}{-1}\CircleANum{20}{+1}\CircleANum{40}{+2}\CircleANum{60}{+3}\CircleANum{80}{+4}\CircleANum{100}{+5}}, where the ranking order matches the order in the generation phase perfectly. In this case, all the values of the metrics are 1.0, which is the desired property for proper evaluation metrics. 
    \item \textit{Cases to show intensity ranking concordance for intensity discerning}: In this sequence: \resizebox{!}{0.8em}{ \SquareDNum{100}{-5}\SquareDNum{80}{-4}\SquareDNum{60}{-3}\SquareDNum{40}{-2}\CircleANum{20}{+1}\SquareDNum{20}{-1}\CircleANum{40}{+2}\CircleANum{60}{+3}\CircleANum{80}{+4}\CircleANum{100}{+5}}, similar as the optimal case, but a slight difference between \resizebox{!}{0.8em}{\SquareDNum{20}{-1}} and \resizebox{!}{0.8em}{\CircleANum{20}{+1}}. This difference doesn't change the intensity ranking concordance within the supporter and the defeater group. Namely, \tauA and \tauD keep the same. However, this difference changes the intensity ranking concordance for the entire intermediate set, i.e., \tauOverall. 
    \item \textit{Cases to show the changes captured by cross-group position agreement for polarity differentiation}: 
    In the first sequence: \resizebox{!}{0.8em}{\SquareDNum{100}{-5}\SquareDNum{80}{-4}\SquareDNum{40}{-2}\CircleANum{20}{+1}\CircleANum{40}{+2}\SquareDNum{20}{-1} \SquareDNum{60}{-3}\CircleANum{60}{+3}\CircleANum{80}{+4}\CircleANum{100}{+5}}, two supporters (\resizebox{!}{0.8em}{\CircleANum{20}{+1}} and \resizebox{!}{0.8em}{\CircleANum{40}{+2}}) are ranked before two defeaters (\resizebox{!}{0.8em}{\SquareDNum{20}{-1}} and \resizebox{!}{0.8em}{\SquareDNum{60}{-3}}). In this case, the number of cross-group position disagreements is 2 $\times$ 2 = 4. In the second sequence: \resizebox{!}{0.8em}{\SquareDNum{100}{-5}\SquareDNum{80}{-4}\SquareDNum{60}{-3}\SquareDNum{40}{-2}\CircleANum{20}{+1}\CircleANum{40}{+2}\CircleANum{60}{+3}\CircleANum{80}{+4}\SquareDNum{20}{-1}\CircleANum{100}{+5}}, four supporters (\resizebox{!}{0.8em}{\CircleANum{20}{+1}}, \resizebox{!}{0.8em}{\CircleANum{40}{+2}}, \resizebox{!}{0.8em}{\CircleANum{60}{+3}}, and \resizebox{!}{0.8em}{\CircleANum{80}{+4}}) are ranked before one defeater (\resizebox{!}{0.8em}{\SquareDNum{20}{-1}}), resulting in 4 $\times$ 1 = 4 cross-group position disagreements. We observe that both sequences achieve the same \CGP value of 0.840, verifying the efficacy of our proposed \CGP metric.
    \item \textit{Cases to show the changes captured by intra-group clustering for cluster formulation}: In the first sequence \resizebox{!}{0.8em}{\SquareDNum{100}{-5}\SquareDNum{80}{-4}\CircleANum{20}{+1}\SquareDNum{40}{-2}\SquareDNum{20}{-1}\SquareDNum{60}{-3}\CircleANum{40}{+2}\CircleANum{60}{+3}\CircleANum{80}{+4}\CircleANum{100}{+5}} and the second sequence \resizebox{!}{0.8em}{\CircleANum{80}{+4}\CircleANum{60}{+3}\SquareDNum{20}{-1}\CircleANum{40}{+2}\CircleANum{20}{+1}\CircleANum{100}{+5}\SquareDNum{40}{-2}\SquareDNum{60}{-3}\SquareDNum{80}{-4}\SquareDNum{100}{-5}}. Although these two sequences differ significantly, they follow the same sequence clustering pattern. Specifically, the first sequence follows the pattern \squarenum{} \squarenum{} \circlenum{} \squarenum{} \squarenum{} \squarenum{} \circlenum{} \circlenum{} \circlenum{} \circlenum{}, while the second sequence follows the dual clustering pattern \circlenum{} \circlenum{} \squarenum{} \circlenum{} \circlenum{} \circlenum{} \squarenum{} \squarenum{} \squarenum{} \squarenum{}. This dualarit is verified by the identical \IGC metric values of 0.543 for both sequences. 
\end{itemize}

%% file: figures/table_minipages/metrics_comparison_via_example.tex
\newcommand{\stdvalue}[1]{$\pm$ \tiny #1}
\setlength{\columnsep}{2.0cm}

\begin{tabular}{l|C{\columnsep}C{\columnsep}C{\columnsep}|C{\columnsep}|C{\columnsep}}
\toprule 
Sequence & \multicolumn{3}{>{\columncolor{\rankingColor}}m{6.8cm}}{\centering \textit{Intensity ranking concordance}} 
& \multicolumn{1}{>{\columncolor{\crossColor}}m{2.0cm}}{\textit{Cross-group position}} 
& \multicolumn{1}{>{\columncolor{\clusteringColor}}m{2.0cm}}{\textit{Intra-group clustering}} 
\\
    \midrule
    & \tauA$\uparrow$ & \tauD$\uparrow$ & \tauOverall$\uparrow$ & \centering \CGP $\uparrow$ &  \multicolumn{1}{>{\columncolor{white}}c}{\IGC $\uparrow$} \\
    \midrule
    \multicolumn{6}{>{\columncolor{mygray}}c}{\textit{Optimal case}}\\ \midgrayline
    \SquareDNum{100}{-5}\SquareDNum{80}{-4}\SquareDNum{60}{-3}\SquareDNum{40}{-2}\SquareDNum{20}{-1}\CircleANum{20}{+1}\CircleANum{40}{+2}\CircleANum{60}{+3}\CircleANum{80}{+4}\CircleANum{100}{+5} & 1.0 & 1.0 & 1.0 & 1.0 & 1.0   \\ 
    \midgrayline
    \multicolumn{6}{>{\columncolor{mygray}}c}{\textit{Cases to show changes captured by intensity ranking concordance. }}\\ 
    \midgrayline
    \SquareDNum{100}{-5}\SquareDNum{80}{-4}\SquareDNum{60}{-3}\SquareDNum{40}{-2}\SquareDNum{20}{-1} \CircleANum{20}{+1}\CircleANum{40}{+2}\CircleANum{60}{+3}\CircleANum{80}{+4}\CircleANum{100}{+5} & 1.000 & 1.000 & 1.000 & 1.000 & 1.000   \\ 
    \SquareDNum{100}{-5}\SquareDNum{80}{-4}\SquareDNum{60}{-3}\SquareDNum{40}{-2}\CircleANum{20}{+1}\SquareDNum{20}{-1}\CircleANum{40}{+2}\CircleANum{60}{+3}\CircleANum{80}{+4}\CircleANum{100}{+5} & 1.000  & 1.000  & 0.867  & 0.960  & 0.694     \\ \midgrayline    
    \multicolumn{6}{>{\columncolor{mygray}}c}{\textit{Cases to show the changes captured by cross-group position agreement. }}\\ \midgrayline
    \SquareDNum{100}{-5}\SquareDNum{80}{-4}\SquareDNum{40}{-2}\CircleANum{20}{+1}\CircleANum{40}{+2}\SquareDNum{20}{-1} \SquareDNum{60}{-3}\CircleANum{60}{+3}\CircleANum{80}{+4}\CircleANum{100}{+5} & 1.000  & -0.200  & 0.289  & \ul{0.840}  & 0.510  \\
    \SquareDNum{100}{-5}\SquareDNum{80}{-4}\SquareDNum{60}{-3}\SquareDNum{40}{-2}\CircleANum{20}{+1}\CircleANum{40}{+2}\CircleANum{60}{+3}\CircleANum{80}{+4}\SquareDNum{20}{-1}\CircleANum{100}{+5} & 1.000  & 1.000  & 0.467 & \ul{0.840}  & 0.502   \\   
    \multicolumn{6}{>{\columncolor{mygray}}c}{\textit{Cases to show the changes captured by intra-group clustering. }}\\ \midgrayline
    \SquareDNum{100}{-5}\SquareDNum{80}{-4}\CircleANum{20}{+1}\SquareDNum{40}{-2}\SquareDNum{20}{-1} \SquareDNum{60}{-3}\CircleANum{40}{+2}\CircleANum{60}{+3}\CircleANum{80}{+4}\CircleANum{100}{+5} &  1.000  & -0.200  & 0.244  & 0.880  & \ul{0.543}    \\   
    \CircleANum{80}{+4}\CircleANum{60}{+3}\SquareDNum{20}{-1}\CircleANum{40}{+2}\CircleANum{20}{+1} \CircleANum{100}{+5}\SquareDNum{40}{-2}\SquareDNum{60}{-3}\SquareDNum{80}{-4}\SquareDNum{100}{-5} &-0.600  & 0.200  & 0.378  & 0.120  & \ul{0.543}  \\     
  \bottomrule
\end{tabular}

%% file: appendices/results.tex
\newlength{\heatmapwidth}
\setlength{\heatmapwidth}{0.3\textwidth}

\subsection{Visualizations of Causal Epistemic Consistency of All Models} 
In this subsection, we present the visualizations of causal epistemic consistency of all the studied LLMs in \Figure~\ref{fig:closed_source_llm_heatmap}. Each subfigure within the figure represents a specific model, showing how each LLMs performs regarding causal epistemic consistency. The result indicates that larger models tend to exhibit more stable and consistent differentiation of fine-grained intermediates.

\begin{figure*}
    \centering
    \begin{tabular}{ccc}
        \begin{subfigure}[b]{\heatmapwidth}
            \centering
            \includegraphics[width=\textwidth]{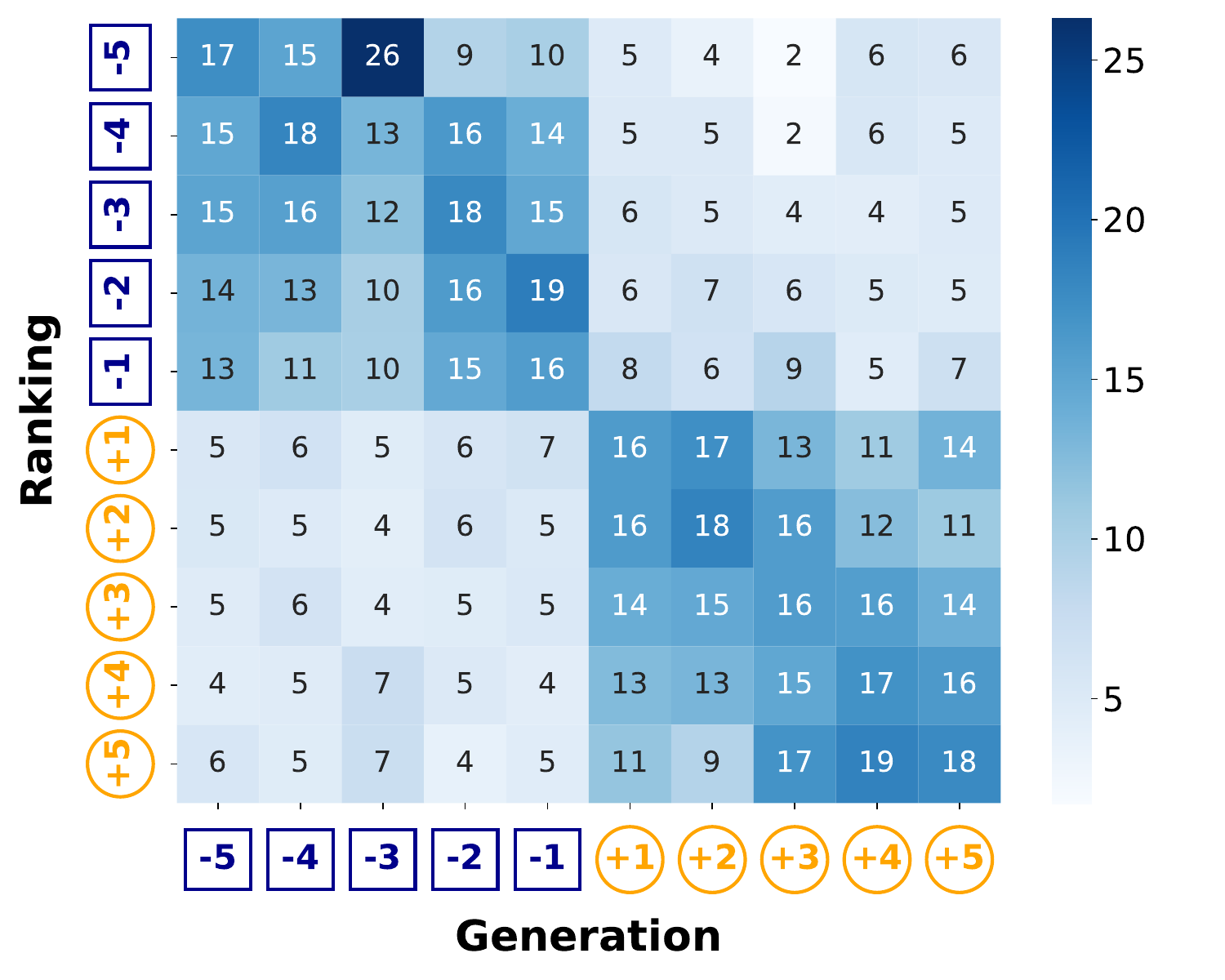}
            \caption{GPT-3.5 Turbo.}
        \end{subfigure} &
        \begin{subfigure}[b]{\heatmapwidth}
            \centering
            \includegraphics[width=\textwidth]{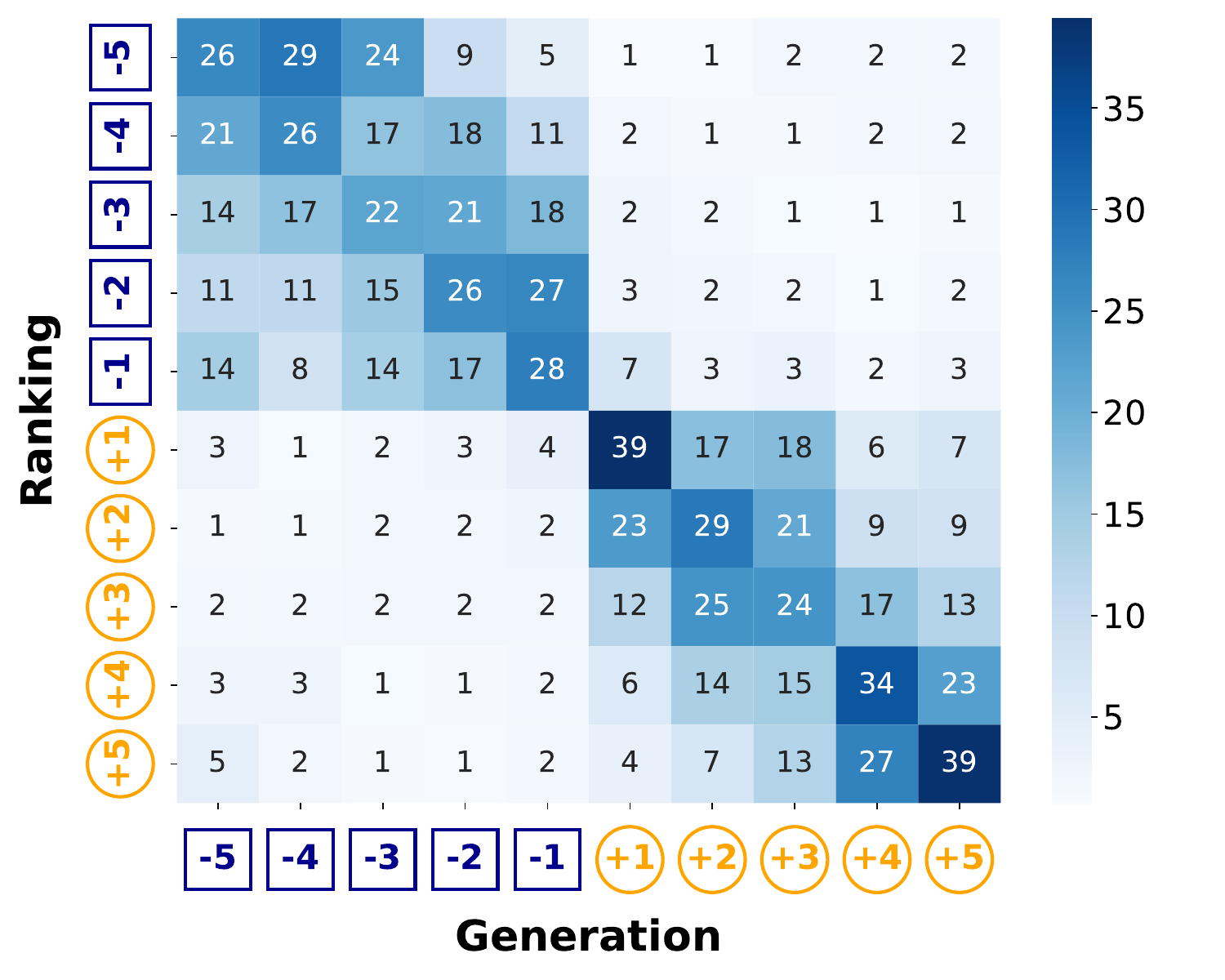}
            \caption{GPT-4.}
        \end{subfigure} &
        \begin{subfigure}[b]{\heatmapwidth}
            \centering
            \includegraphics[width=\textwidth]{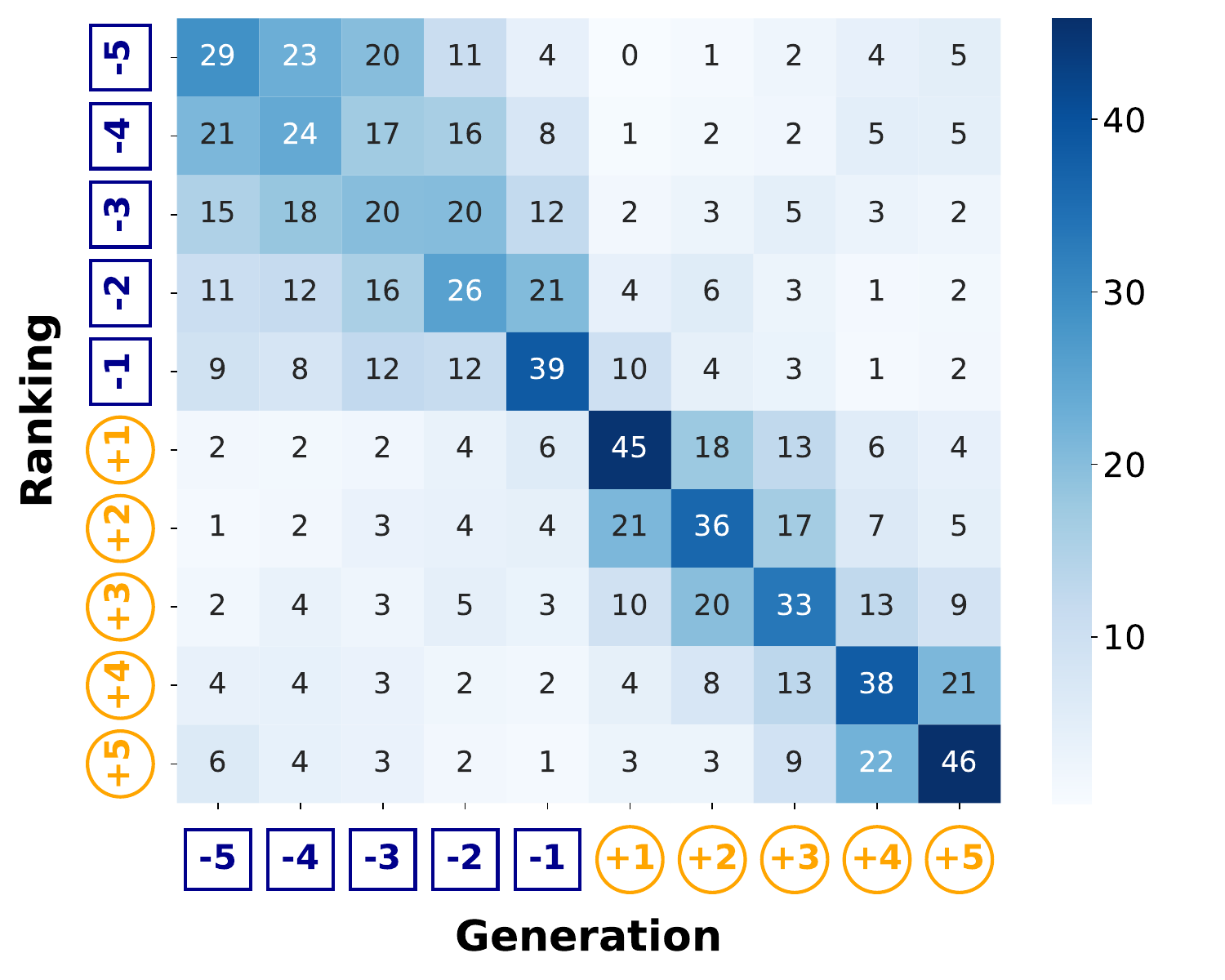}
            \caption{GPT-4 Turbo.}
        \end{subfigure} \\
        
        \begin{subfigure}[b]{\heatmapwidth}
            \centering
            \includegraphics[width=\textwidth]{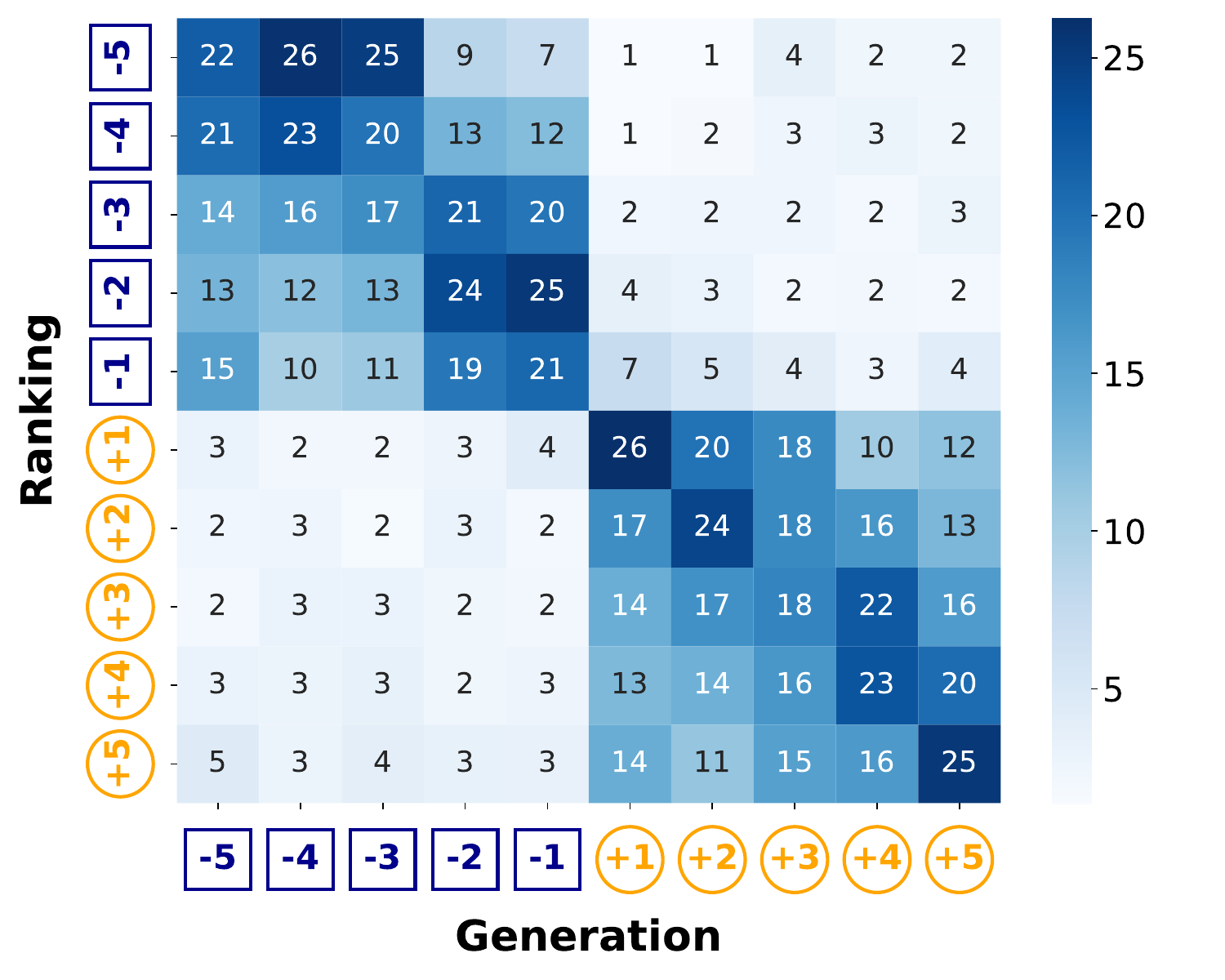}
            \caption{GPT-4o mini}
        \end{subfigure} &
        \begin{subfigure}[b]{\heatmapwidth}
            \centering
            \includegraphics[width=\textwidth]{figures/results/confusion_matrices/confusion_matrix_ranking_gpt-4o.pdf}
            \caption{GPT-4o. }
        \end{subfigure} &
        \begin{subfigure}[b]{\heatmapwidth}
            \centering
            \includegraphics[width=\textwidth]{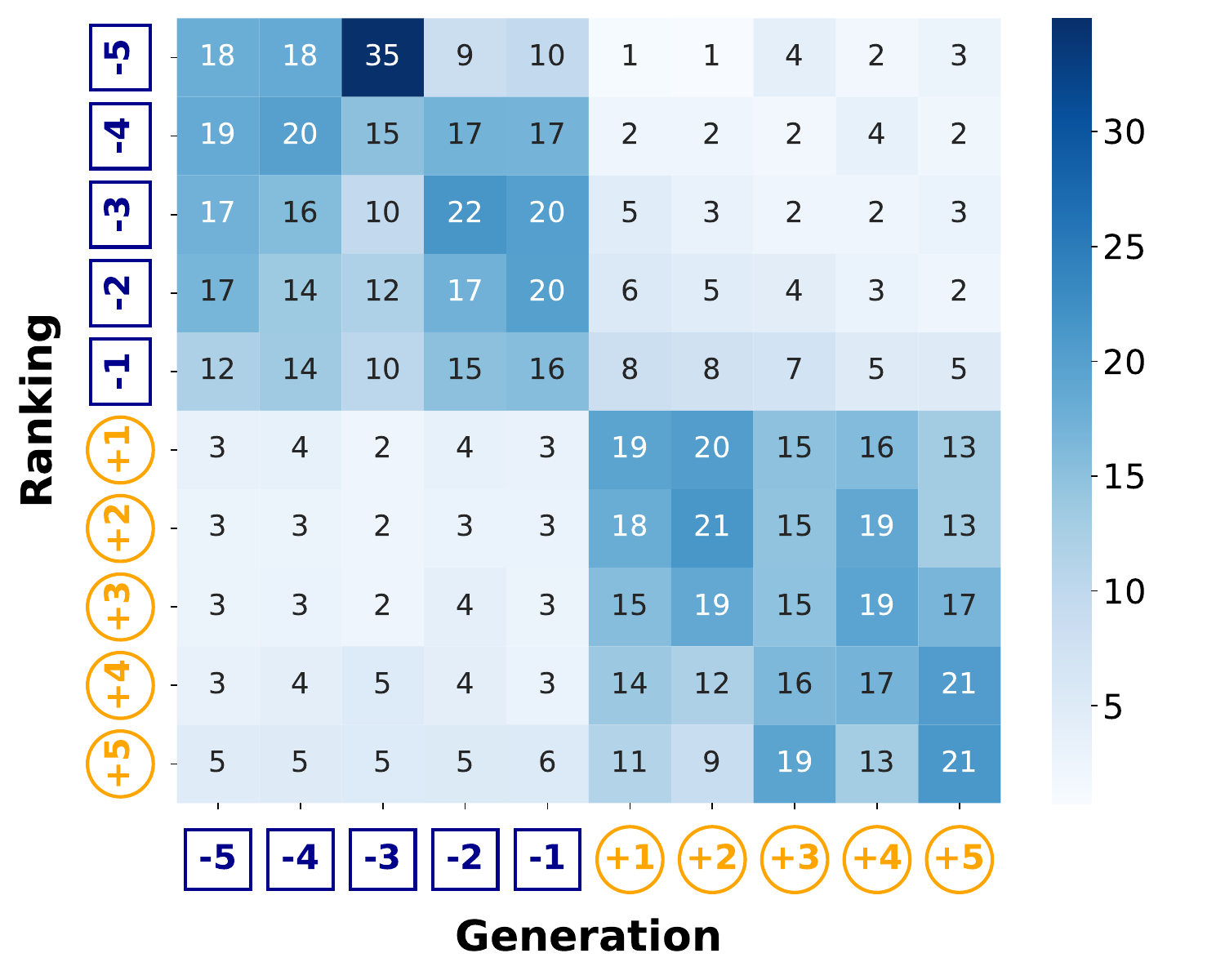}
            \caption{Claude 3 Haiku}
        \end{subfigure} \\
        \begin{subfigure}[b]{\heatmapwidth}
            \centering
            \includegraphics[width=\textwidth]{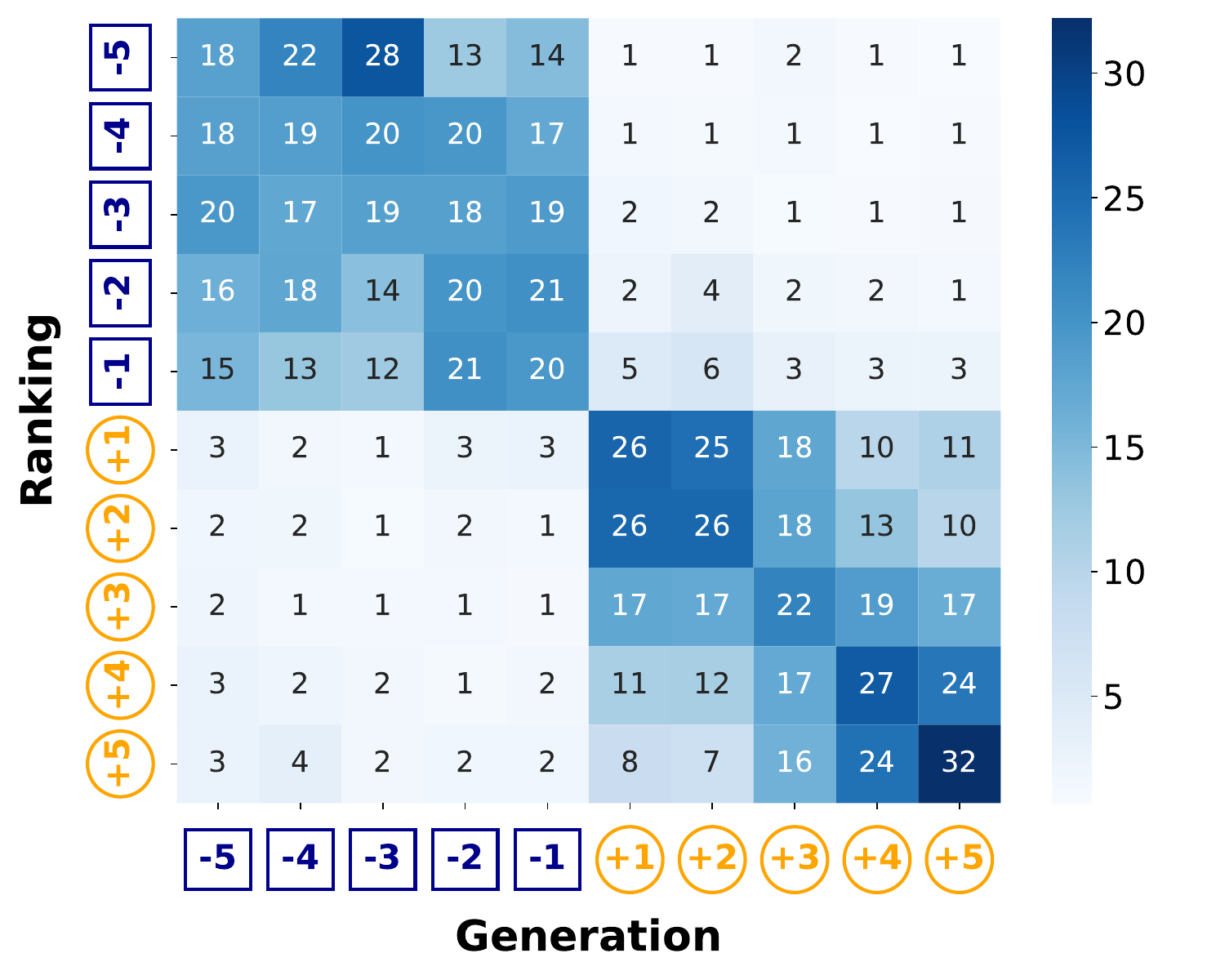}
            \caption{Claude 3 Sonnet. }
        \end{subfigure} & 
        \begin{subfigure}[b]{\heatmapwidth}
            \centering
            \includegraphics[width=\textwidth]{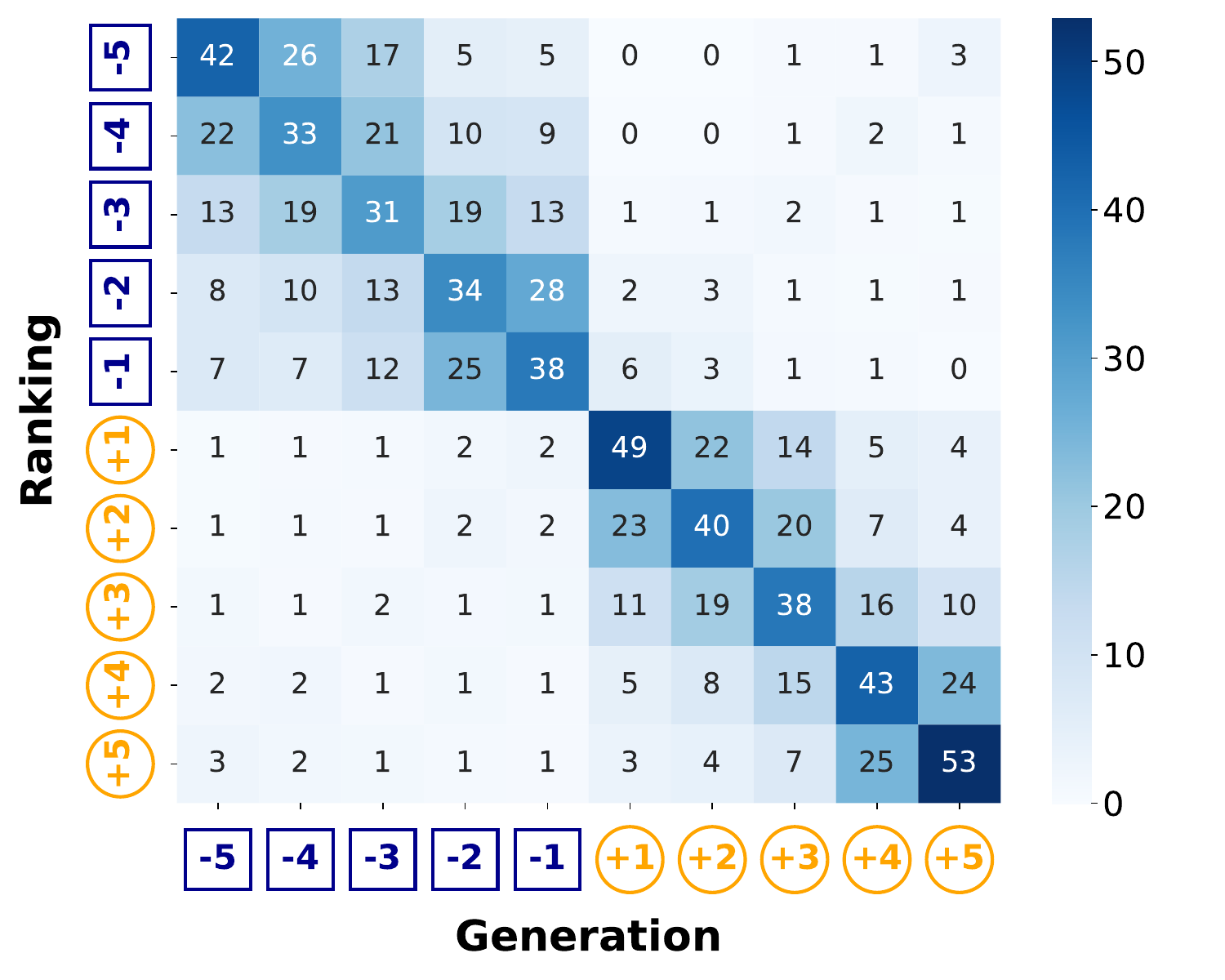}
            \caption{Claude 3 Opus. }
        \end{subfigure} & 
        \begin{subfigure}[b]{\heatmapwidth}
            \centering
            \includegraphics[width=\textwidth]{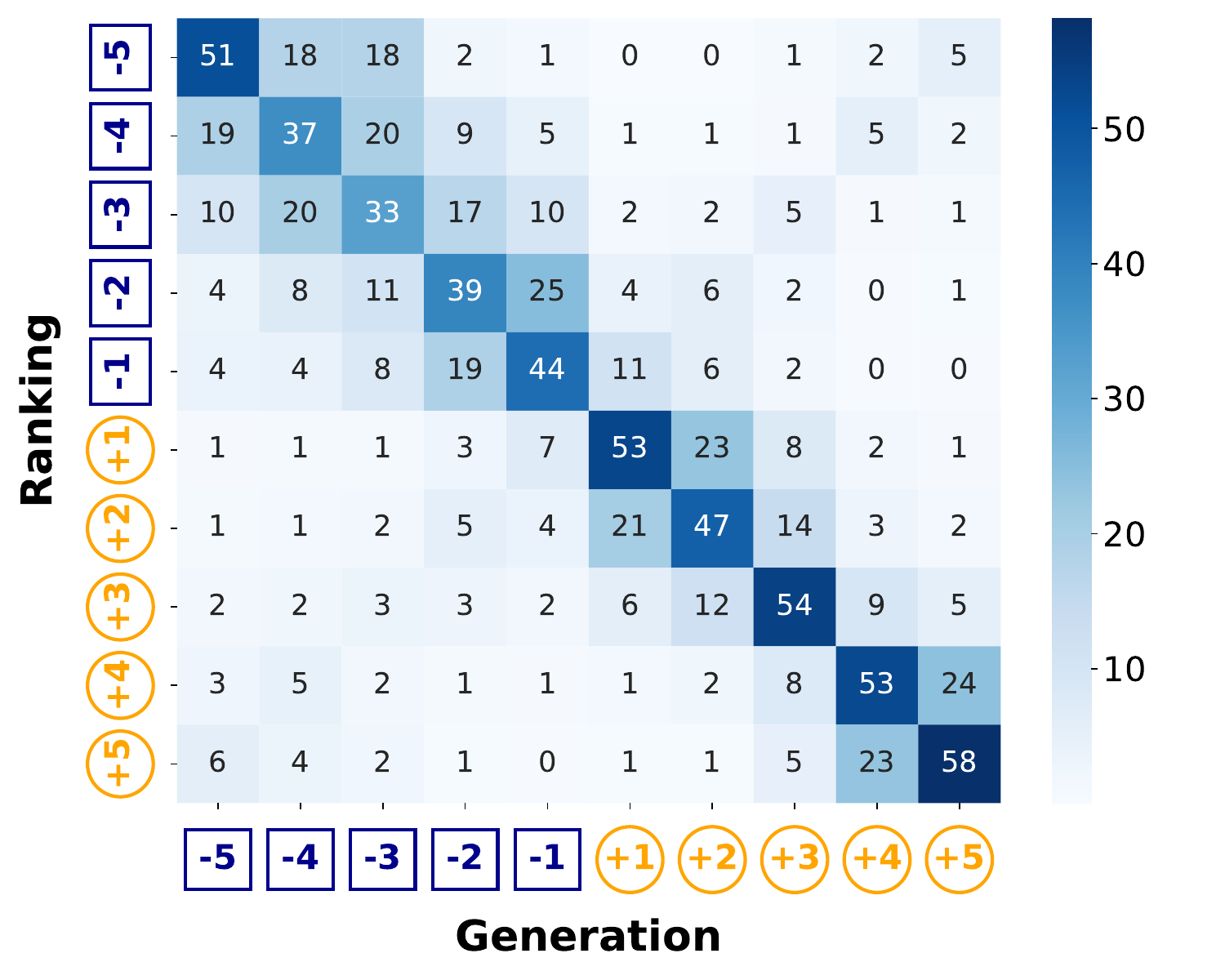}
            \caption{Claude 3.5 Sonnet. }
        \end{subfigure}
        \\
        \begin{subfigure}[b]{\heatmapwidth}
            \centering
            \includegraphics[width=\textwidth]{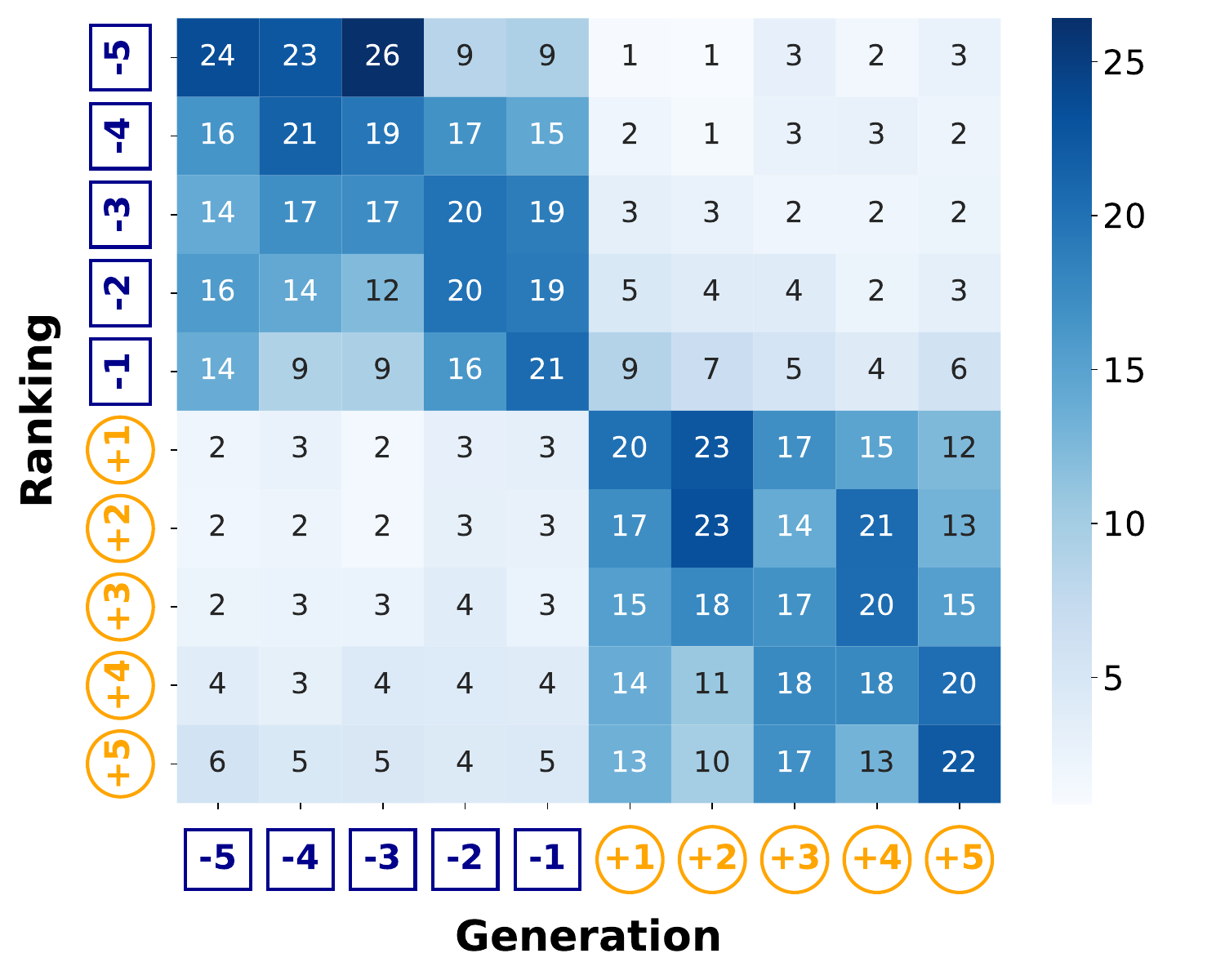}
            \caption{Gemini 1.5 Flash. }
        \end{subfigure} & 
        \begin{subfigure}[b]{\heatmapwidth}
            \centering
            \includegraphics[width=\textwidth]{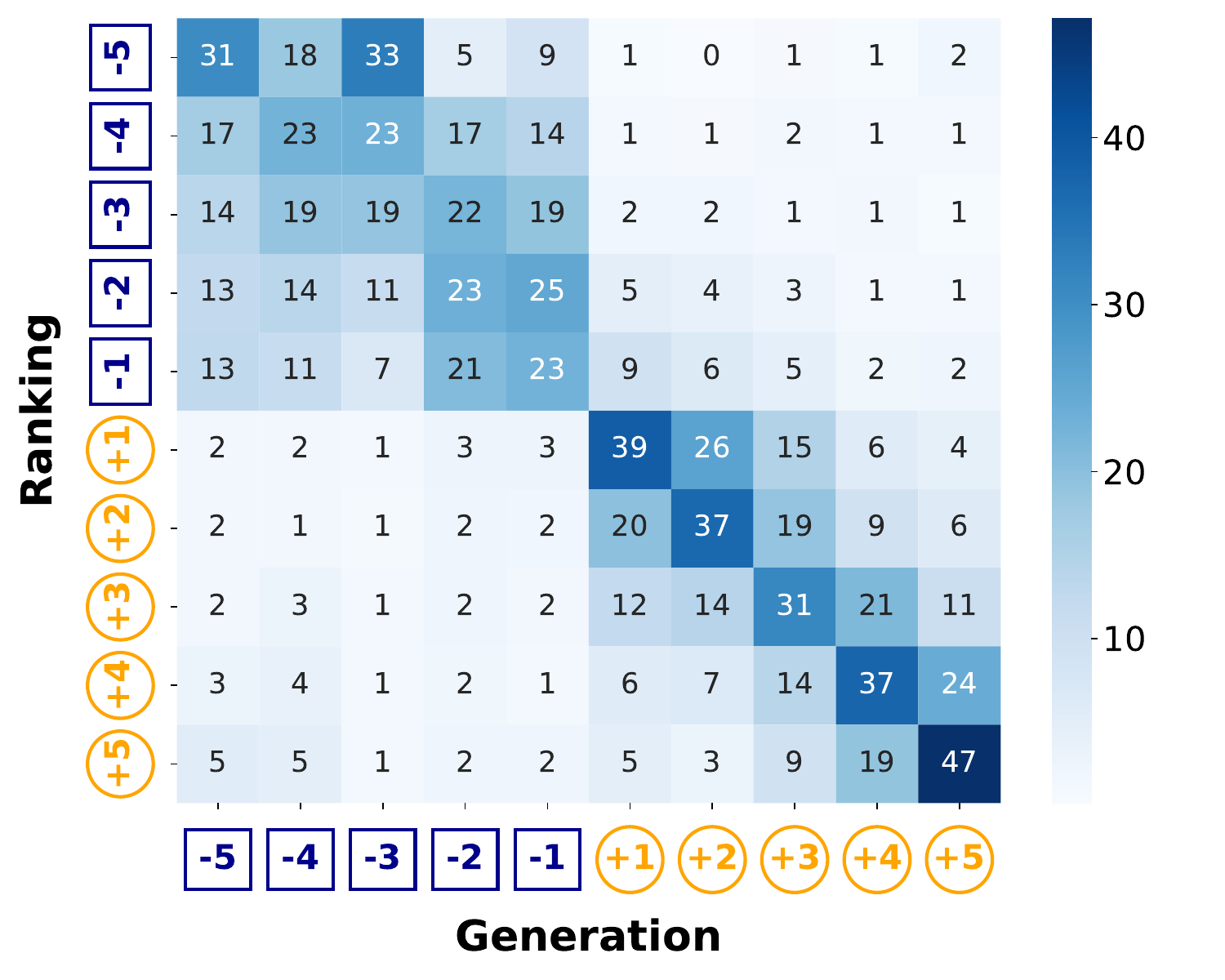}
            \caption{Gemini 1.5 Pro.}
        \end{subfigure}        
    \end{tabular}
    \caption{Visualization of causal epistemic consistency in different closed-source LLMs, highlighting their performance in maintaining consistency across generated intermediates. }
    \label{fig:closed_source_llm_heatmap}
\end{figure*}

\begin{figure*}[ht!]
    \ContinuedFloat
    \centering
    \begin{tabular}{ccc}
        \begin{subfigure}[b]{\heatmapwidth}
            \centering
            \includegraphics[width=\textwidth]{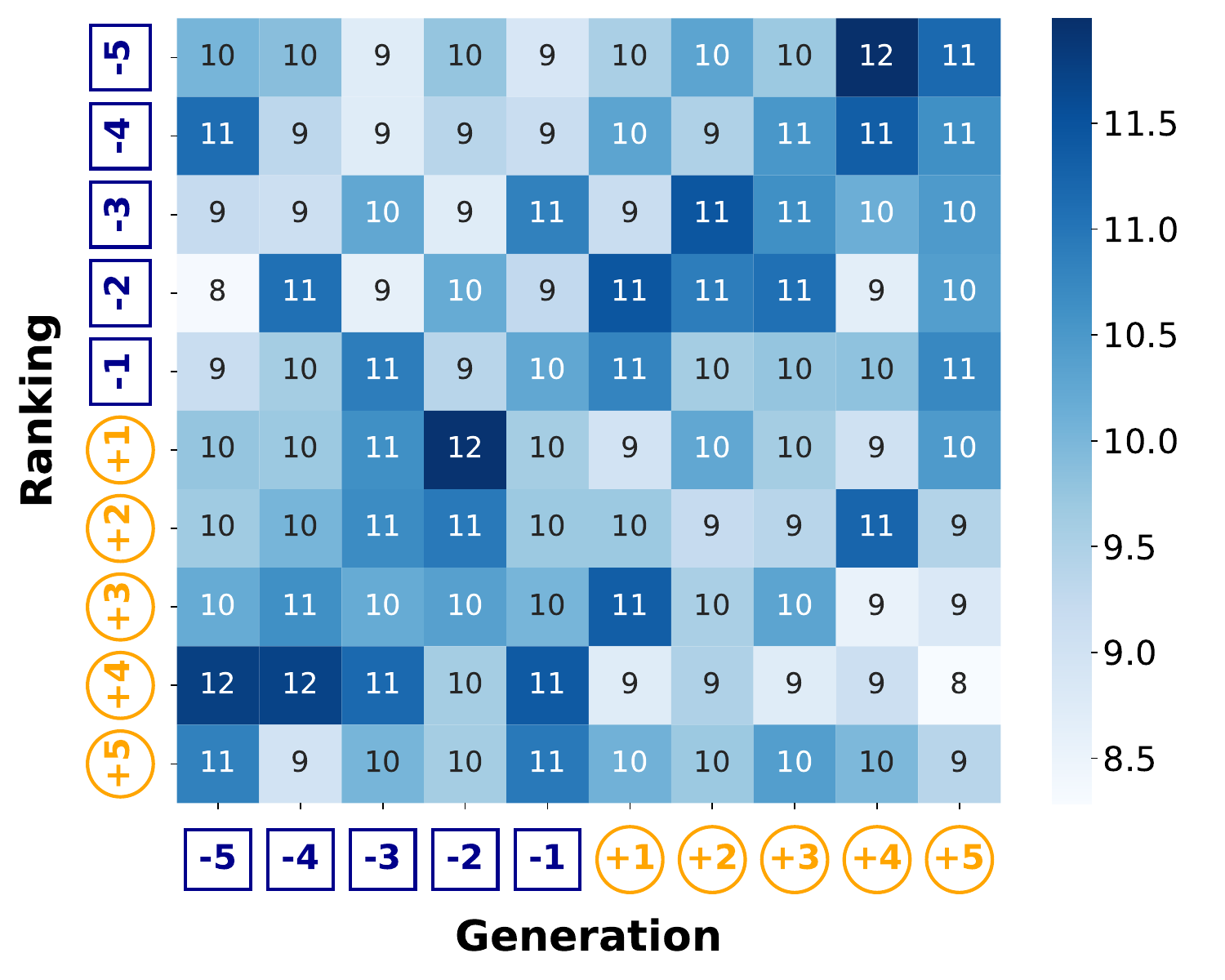}
            \caption{LLaMA2-7B.}
        \end{subfigure} &
        \begin{subfigure}[b]{\heatmapwidth}
            \centering
            \includegraphics[width=\textwidth]{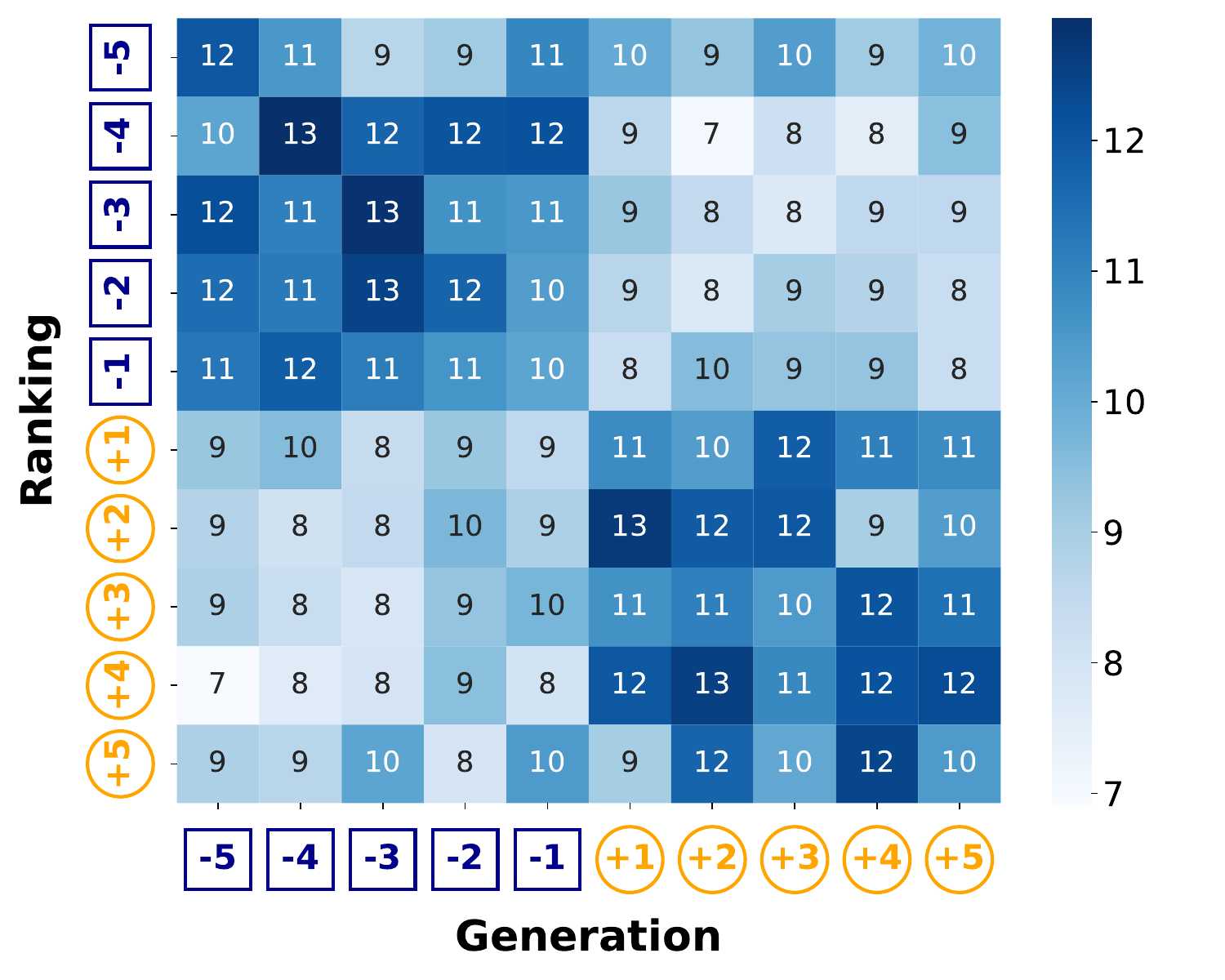}
            \caption{LLaMA2-13B.}
        \end{subfigure} &
        \begin{subfigure}[b]{\heatmapwidth}
            \centering
            \includegraphics[width=\textwidth]{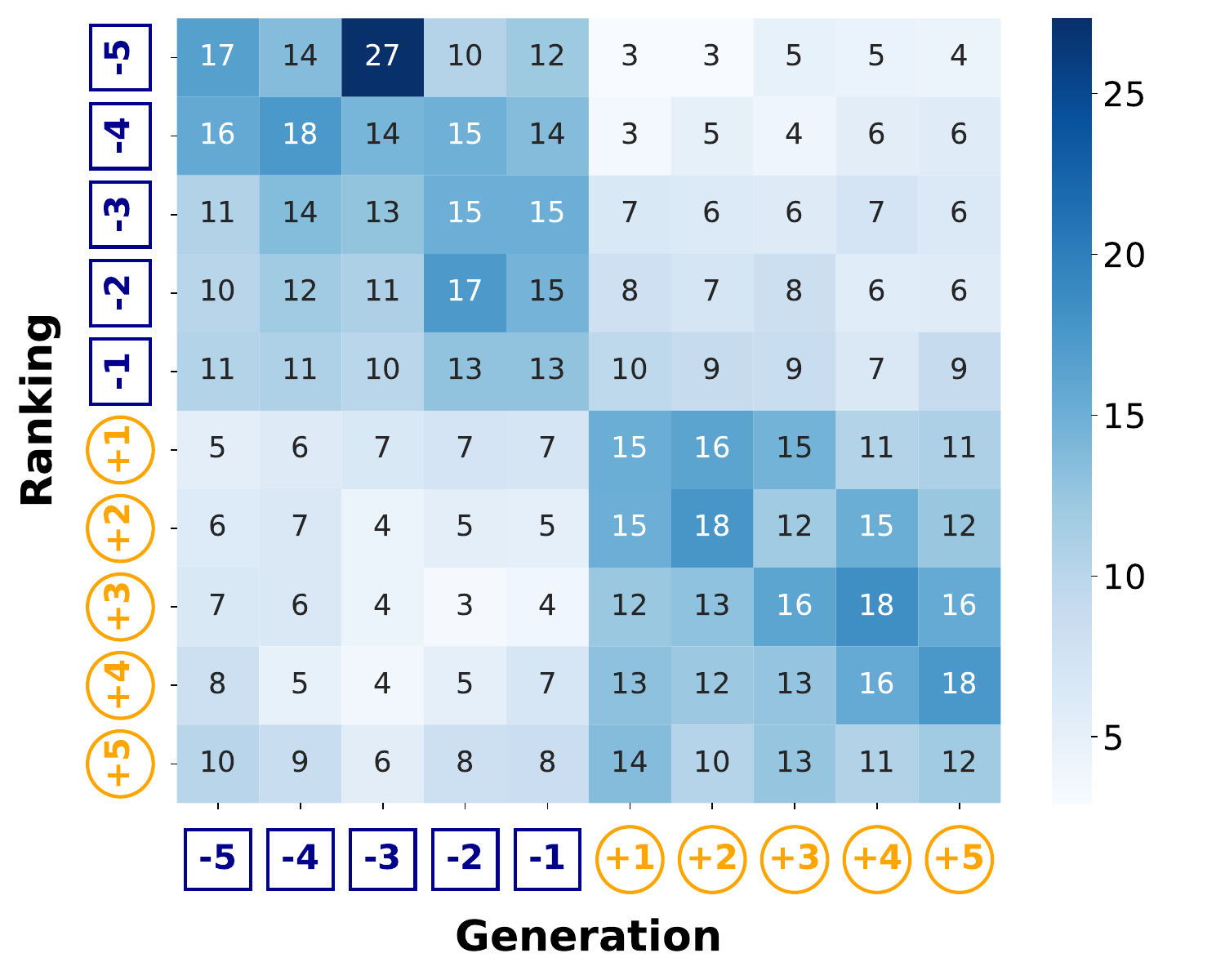}
            \caption{LLaMA-70B. }
        \end{subfigure} \\
        \begin{subfigure}[b]{\heatmapwidth}
            \centering
            \includegraphics[width=\textwidth]{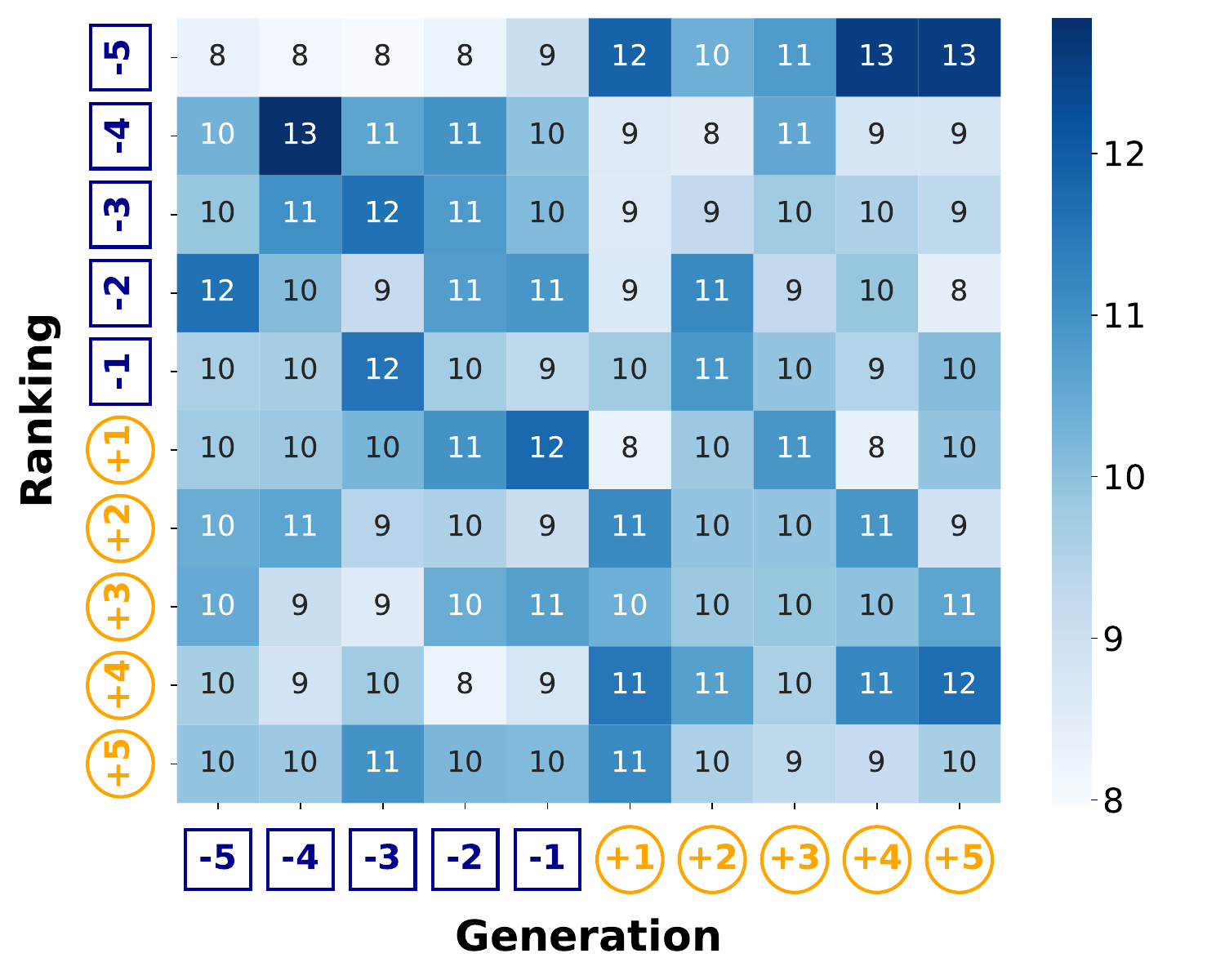}
            \caption{Gemma-2B. }
        \end{subfigure} &
        \begin{subfigure}[b]{\heatmapwidth}
            \centering
            \includegraphics[width=\textwidth]{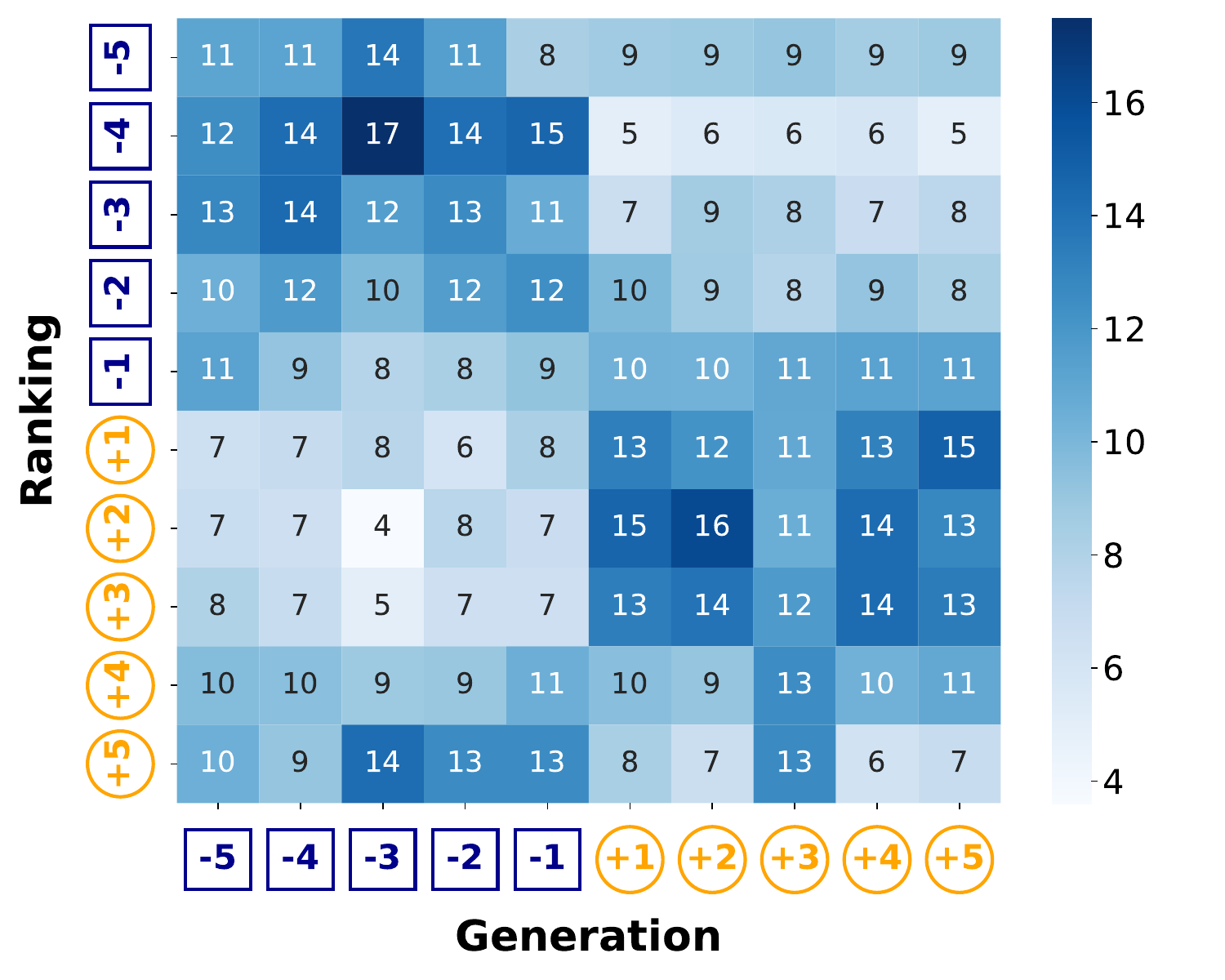}
            \caption{Gemma-7B. }
        \end{subfigure} &
        \begin{subfigure}[b]{\heatmapwidth}
            \centering
            \includegraphics[width=\textwidth]{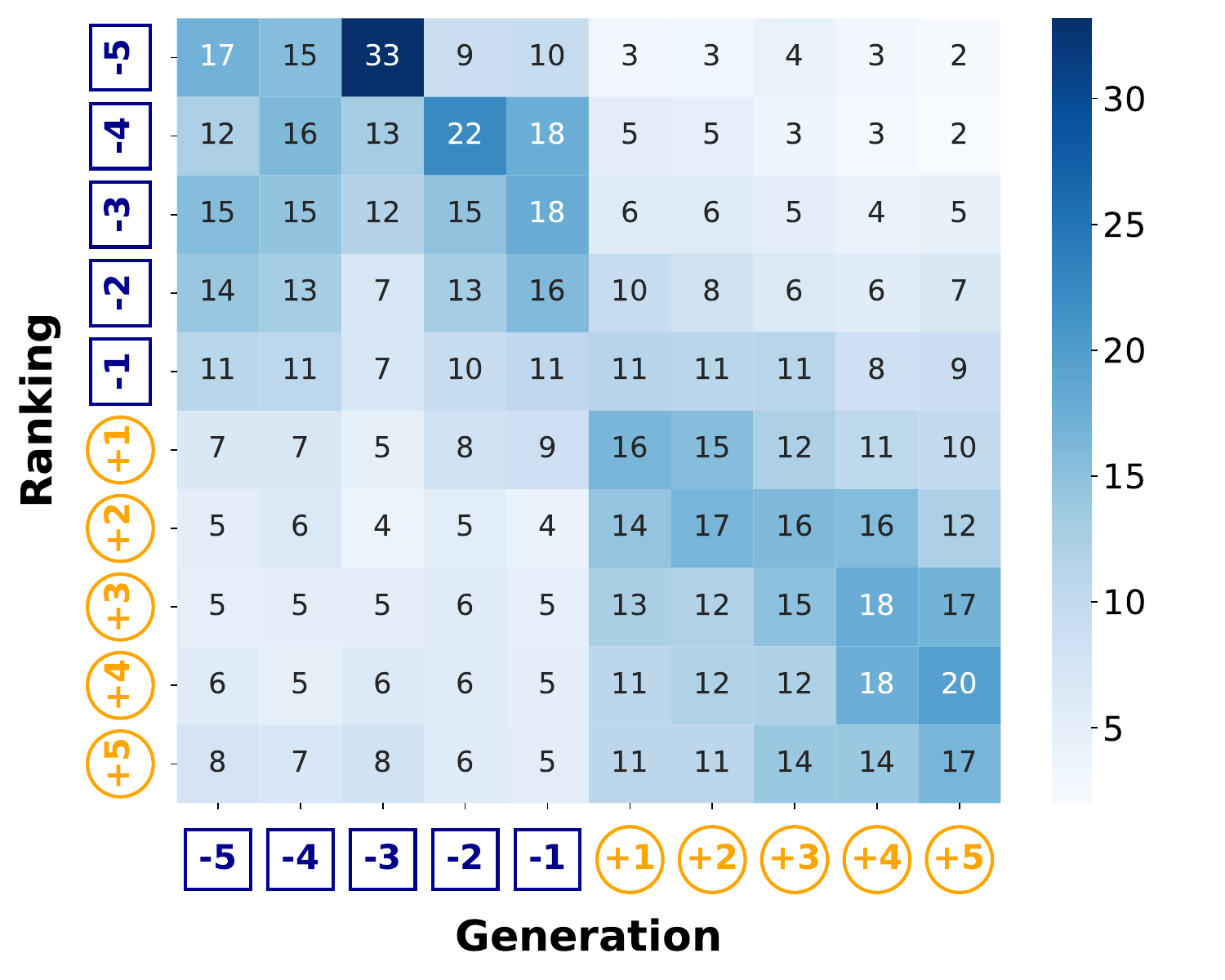}
            \caption{Phi-3 mini. }
        \end{subfigure} \\
        \begin{subfigure}[b]{\heatmapwidth}
            \centering
            \includegraphics[width=\textwidth]{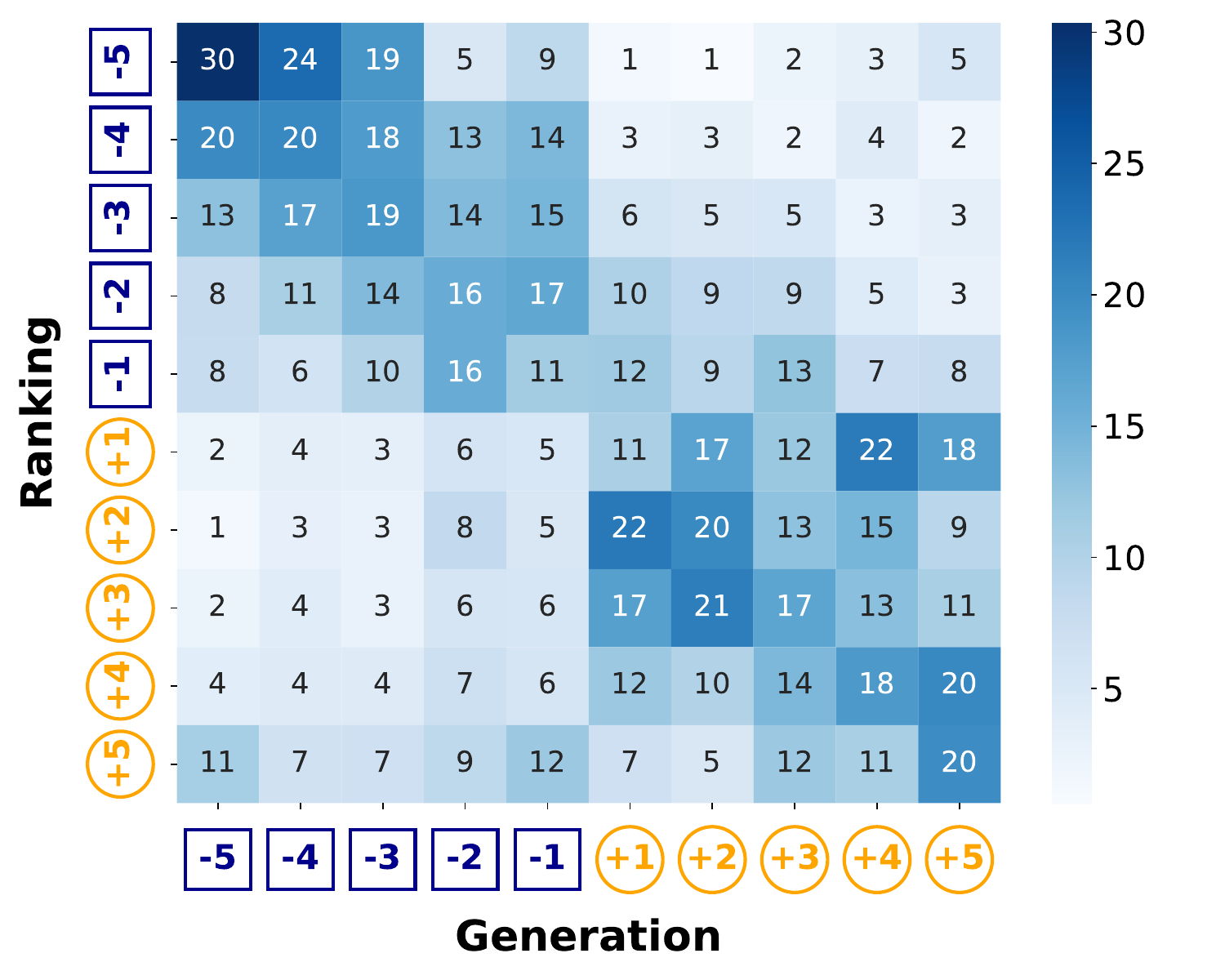}
            \caption{Phi-3 small. }
        \end{subfigure} &
        \begin{subfigure}[b]{\heatmapwidth}
            \centering
            \includegraphics[width=\textwidth]{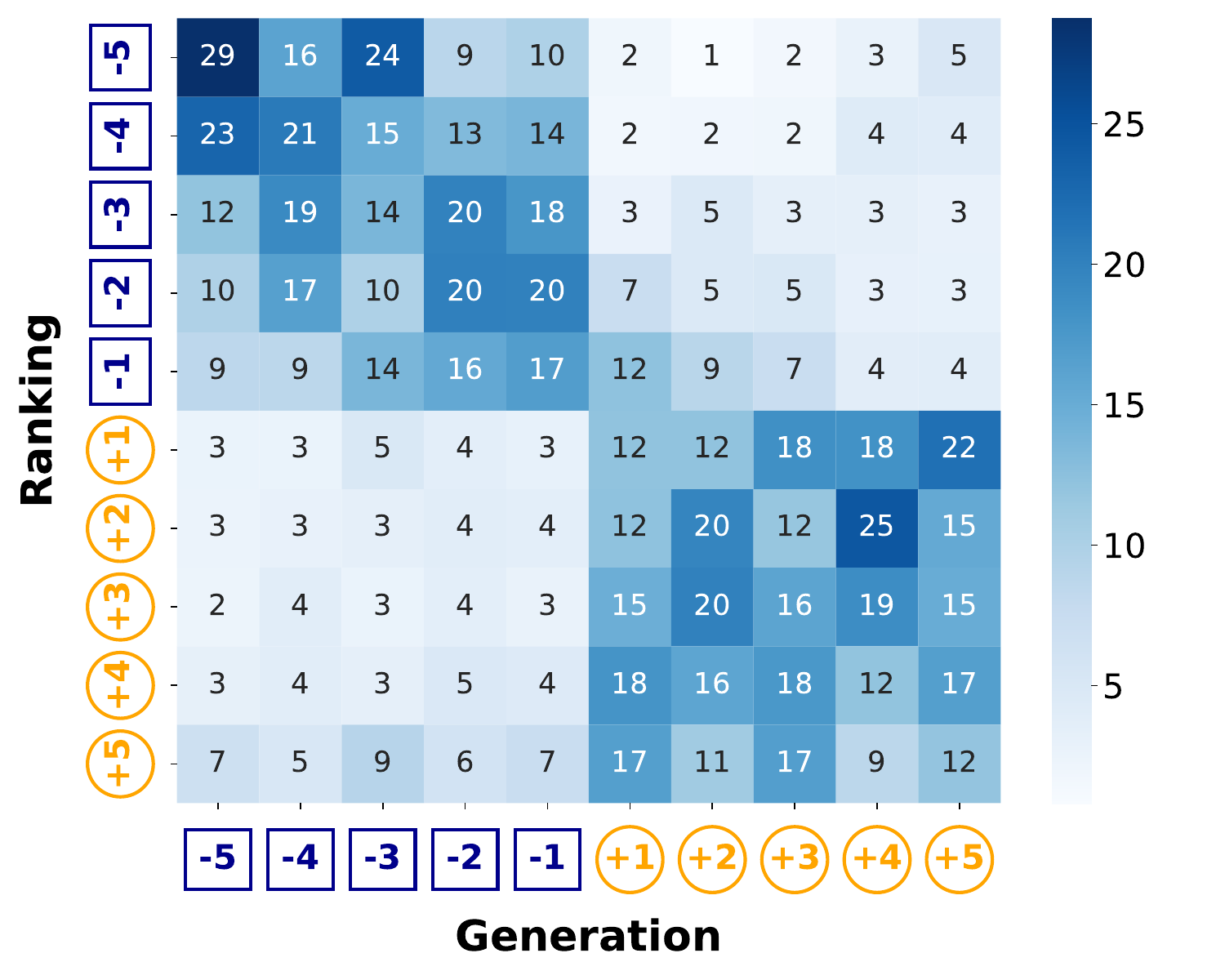}
            \caption{Phi-3 medium. }
        \end{subfigure} &
        \begin{subfigure}[b]{\heatmapwidth}
            \centering
            \includegraphics[width=\textwidth]{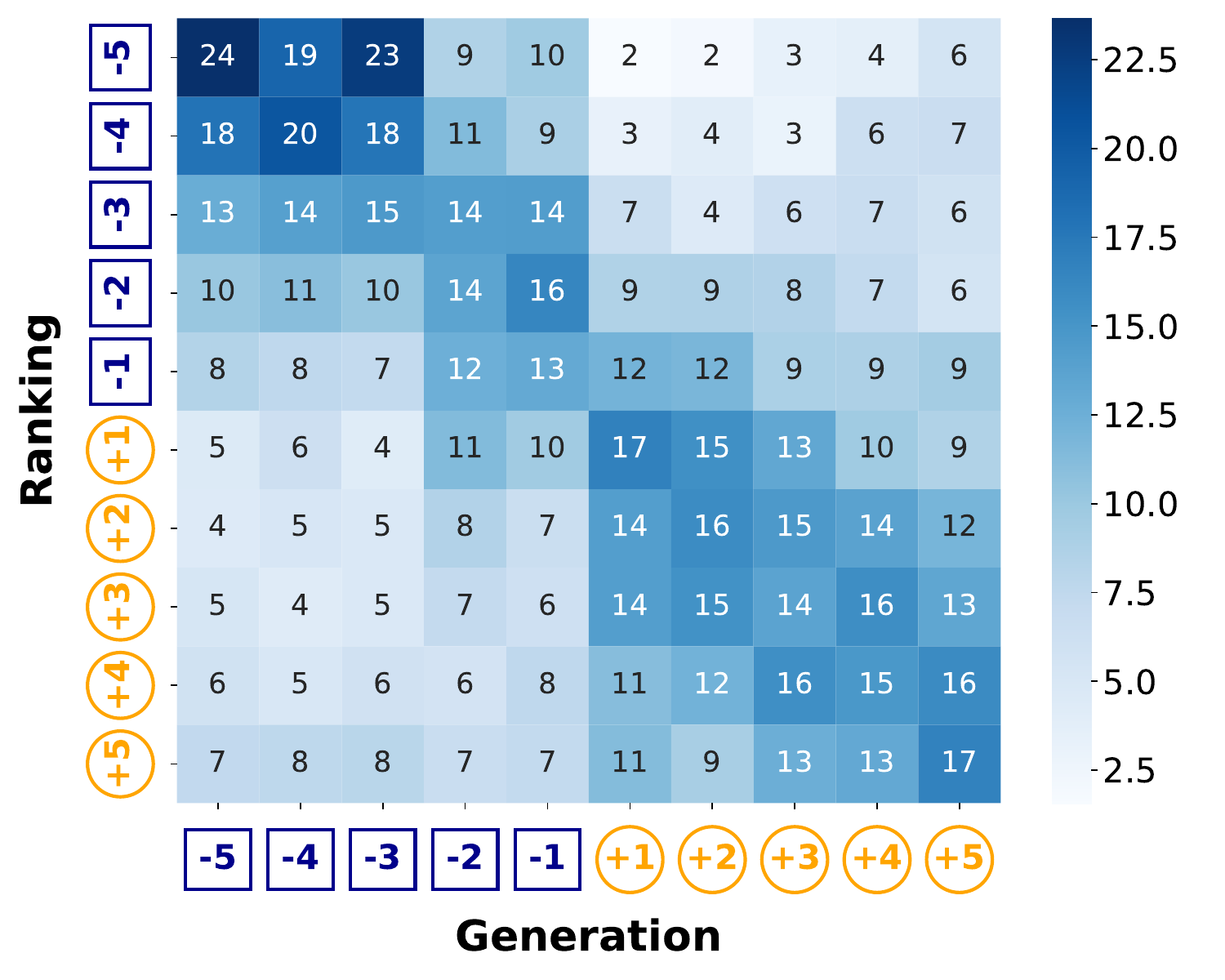}
            \caption{LLaMA3-8B. }
        \end{subfigure} \\
        \begin{subfigure}[b]{\heatmapwidth}
            \centering
            \includegraphics[width=\textwidth]{figures/results/confusion_matrices/confusion_matrix_ranking_Meta-Llama-3-70B-Instruct.pdf}
            \caption{LLaMA3-70B. }
        \end{subfigure}
    \end{tabular}
    \caption{Continued visualization of causal epistemic consistency in different open-source LLMs, providing a comparative analysis against closed-source counterparts.}
    \label{fig:open_source_llm_heatmap}
\end{figure*}

\subsection{More Results with Conjunction Words}
We present the full results of LLMs with different conjunction words in \Table~\ref{tab:results_models_with_different_conjunctions}. 
\onecolumn
\input{figures/table_minipages/conjunction_results}

%% file: figures/table_minipages/conjunction_results.tex
\newcommand{\stdvalue}[1]{$\pm$ \tiny #1}
\setlength{\columnsep}{2.0cm}

\begin{longtable}{l|p{\columnsep}p{\columnsep}p{\columnsep}|C{\columnsep}|C{\columnsep}}
\toprule 
Aspect & \multicolumn{3}{>{\columncolor{\rankingColor}}m{6.8cm}}{\centering \textit{Intensity ranking concordance}} 
& \multicolumn{1}{>{\columncolor{\crossColor}}m{2.0cm}}{\textit{Cross-group position}} 
& \multicolumn{1}{>{\columncolor{\clusteringColor}}m{2.0cm}}{\textit{Intra-group clustering}} 
\\
    \midrule
    & \tauA$\uparrow$ & \tauD$\uparrow$ & \tauOverall$\uparrow$ & \centering \CGP $\uparrow$ &  \multicolumn{1}{>{\columncolor{white}}c}{\IGC $\uparrow$} \\
    \midrule
    \multicolumn{6}{>{\columncolor{mygray}}c}{\textit{LLaMA2-7B}}\\ 
    \midgrayline
    prompting &-0.017 \stdvalue{0.403} & 0.007 \stdvalue{0.400} & -0.001 \stdvalue{0.249} & 0.475 \stdvalue{0.199} & 0.474 \stdvalue{0.090} \\
    \midgrayline
    ``so'' & 0.134 \stdvalue{0.432} & 0.122 \stdvalue{0.436} & 0.255 \stdvalue{0.305} & 0.678 \stdvalue{0.227} & 0.557 \stdvalue{0.199} \\
    ``because'' & 0.105 \stdvalue{0.429} & 0.073 \stdvalue{0.423} & 0.149 \stdvalue{0.309} & 0.599 \stdvalue{0.229} & 0.514 \stdvalue{0.155} \\
    ``since'' & 0.107 \stdvalue{0.436} & 0.067 \stdvalue{0.427} & 0.154 \stdvalue{0.308} & 0.604 \stdvalue{0.228} & 0.515 \stdvalue{0.160} \\
    ``as'' & 0.102 \stdvalue{0.436} & 0.062 \stdvalue{0.432} & 0.148 \stdvalue{0.315} & 0.600 \stdvalue{0.233} & 0.520 \stdvalue{0.163} \\
    ``therefore'' & 0.139 \stdvalue{0.445} & 0.140 \stdvalue{0.437} & 0.279 \stdvalue{0.301} & 0.695 \stdvalue{0.227} & 0.574 \stdvalue{0.211} \\
    ``thus'' & 0.126 \stdvalue{0.446} & 0.136 \stdvalue{0.426} & 0.278 \stdvalue{0.299} & 0.698 \stdvalue{0.229} & 0.575 \stdvalue{0.210} \\
    ``hence'' & 0.138 \stdvalue{0.440} & 0.135 \stdvalue{0.426} & 0.285 \stdvalue{0.299} & 0.702 \stdvalue{0.229} & 0.582 \stdvalue{0.214} \\
        \midgrayline

    \multicolumn{6}{>{\columncolor{mygray}}c}{\textit{LLaMA2-13B}}\\ 
    \midgrayline
    prompting &  -0.000 \stdvalue{0.411} & 0.026 \stdvalue{0.417} & 0.072 \stdvalue{0.256} & 0.560 \stdvalue{0.197} & 0.480 \stdvalue{0.109} \\
    \midgrayline
``so'' & 0.132 \stdvalue{0.423} & 0.128 \stdvalue{0.413} & 0.283 \stdvalue{0.281} & 0.703 \stdvalue{0.221} & 0.567 \stdvalue{0.206} \\
``because'' & 0.114 \stdvalue{0.430} & 0.078 \stdvalue{0.427} & 0.193 \stdvalue{0.298} & 0.635 \stdvalue{0.223} & 0.524 \stdvalue{0.168} \\
``since'' & 0.117 \stdvalue{0.429} & 0.087 \stdvalue{0.433} & 0.199 \stdvalue{0.299} & 0.639 \stdvalue{0.225} & 0.530 \stdvalue{0.174} \\
``as'' & 0.107 \stdvalue{0.432} & 0.091 \stdvalue{0.423} & 0.185 \stdvalue{0.299} & 0.627 \stdvalue{0.222} & 0.519 \stdvalue{0.164} \\
``therefore'' & 0.137 \stdvalue{0.427} & 0.133 \stdvalue{0.422} & 0.303 \stdvalue{0.282} & 0.719 \stdvalue{0.221} & 0.582 \stdvalue{0.217} \\
``thus'' & 0.123 \stdvalue{0.427} & 0.135 \stdvalue{0.420} & 0.305 \stdvalue{0.283} & 0.723 \stdvalue{0.219} & 0.589 \stdvalue{0.220} \\
``hence'' & 0.134 \stdvalue{0.423} & 0.143 \stdvalue{0.421} & 0.309 \stdvalue{0.280} & 0.722 \stdvalue{0.219} & 0.587 \stdvalue{0.219} \\
    \midgrayline
    \multicolumn{6}{>{\columncolor{mygray}}c}{\textit{LLaMA2-70B}}\\ 
    \midgrayline
    prompting & 0.012 \stdvalue{0.409} & 0.010 \stdvalue{0.434} & 0.234 \stdvalue{0.349} & 0.707 \stdvalue{0.271} & 0.629 \stdvalue{0.215} \\
    \midgrayline
    ``so'' & 0.097 \stdvalue{0.424} & 0.107 \stdvalue{0.415} & 0.231 \stdvalue{0.314} & 0.667 \stdvalue{0.241} & 0.561 \stdvalue{0.210} \\
    ``because'' & 0.072 \stdvalue{0.429} & 0.045 \stdvalue{0.429} & 0.115 \stdvalue{0.323} & 0.580 \stdvalue{0.237} & 0.512 \stdvalue{0.160} \\
    ``since'' & 0.080 \stdvalue{0.429} & 0.058 \stdvalue{0.425} & 0.132 \stdvalue{0.318} & 0.591 \stdvalue{0.240} & 0.518 \stdvalue{0.165} \\
    ``as'' & 0.074 \stdvalue{0.434} & 0.046 \stdvalue{0.428} & 0.128 \stdvalue{0.319} & 0.591 \stdvalue{0.238} & 0.519 \stdvalue{0.166} \\
    ``therefore'' & 0.090 \stdvalue{0.430} & 0.107 \stdvalue{0.409} & 0.250 \stdvalue{0.306} & 0.685 \stdvalue{0.236} & 0.571 \stdvalue{0.215} \\
    ``thus'' & 0.102 \stdvalue{0.430} & 0.122 \stdvalue{0.409} & 0.252 \stdvalue{0.311} & 0.682 \stdvalue{0.242} & 0.573 \stdvalue{0.216} \\
    ``hence'' & 0.102 \stdvalue{0.431} & 0.120 \stdvalue{0.410} & 0.255 \stdvalue{0.308} & 0.685 \stdvalue{0.237} & 0.569 \stdvalue{0.213} \\
    \multicolumn{6}{>{\columncolor{mygray}}c}{\textit{Gemma-2B}}\\ \midgrayline
    prompting &  -0.021 \stdvalue{0.412} & 0.001 \stdvalue{0.410} & -0.002 \stdvalue{0.245} & 0.502 \stdvalue{0.190} & 0.468 \stdvalue{0.083} \\ 
    \midgrayline
    ``so'' & 0.106 \stdvalue{0.424} & 0.103 \stdvalue{0.440} & 0.192 \stdvalue{0.334} & 0.631 \stdvalue{0.255} & 0.553 \stdvalue{0.204} \\
    ``because'' & 0.076 \stdvalue{0.420} & 0.050 \stdvalue{0.445} & 0.107 \stdvalue{0.334} & 0.571 \stdvalue{0.252} & 0.525 \stdvalue{0.179} \\
    ``since'' & 0.090 \stdvalue{0.425} & 0.063 \stdvalue{0.445} & 0.138 \stdvalue{0.332} & 0.594 \stdvalue{0.252} & 0.532 \stdvalue{0.187} \\
    ``as'' & 0.093 \stdvalue{0.422} & 0.070 \stdvalue{0.455} & 0.135 \stdvalue{0.330} & 0.589 \stdvalue{0.245} & 0.522 \stdvalue{0.175} \\
    ``therefore'' & 0.102 \stdvalue{0.431} & 0.089 \stdvalue{0.440} & 0.198 \stdvalue{0.340} & 0.640 \stdvalue{0.259} & 0.564 \stdvalue{0.217} \\
    ``thus'' & 0.093 \stdvalue{0.424} & 0.095 \stdvalue{0.441} & 0.194 \stdvalue{0.340} & 0.637 \stdvalue{0.260} & 0.563 \stdvalue{0.215} \\
    ``hence'' & 0.101 \stdvalue{0.426} & 0.096 \stdvalue{0.438} & 0.199 \stdvalue{0.339} & 0.639 \stdvalue{0.258} & 0.562 \stdvalue{0.213} \\
    \multicolumn{6}{>{\columncolor{mygray}}c}{\textit{Gemma-7B}}\\ 
    \midgrayline
    prompting & -0.006 \stdvalue{0.392} & 0.016 \stdvalue{0.389} & 0.085 \stdvalue{0.256} & 0.575 \stdvalue{0.203} & 0.484 \stdvalue{0.122} \\
    \midgrayline
    ``so'' & 0.095 \stdvalue{0.441} & 0.175 \stdvalue{0.431} & 0.287 \stdvalue{0.312} & 0.704 \stdvalue{0.240} & 0.592 \stdvalue{0.229} \\
    ``because'' & 0.066 \stdvalue{0.430} & 0.133 \stdvalue{0.425} & 0.220 \stdvalue{0.310} & 0.658 \stdvalue{0.239} & 0.553 \stdvalue{0.200} \\
    ``since'' & 0.066 \stdvalue{0.419} & 0.147 \stdvalue{0.428} & 0.218 \stdvalue{0.311} & 0.653 \stdvalue{0.236} & 0.547 \stdvalue{0.199} \\
    ``as'' & 0.062 \stdvalue{0.426} & 0.146 \stdvalue{0.426} & 0.213 \stdvalue{0.313} & 0.650 \stdvalue{0.235} & 0.544 \stdvalue{0.195} \\
    ``therefore'' & 0.097 \stdvalue{0.439} & 0.177 \stdvalue{0.432} & 0.290 \stdvalue{0.309} & 0.706 \stdvalue{0.237} & 0.586 \stdvalue{0.225} \\
    ``thus'' & 0.077 \stdvalue{0.438} & 0.189 \stdvalue{0.426} & 0.290 \stdvalue{0.307} & 0.708 \stdvalue{0.240} & 0.596 \stdvalue{0.233} \\
    ``hence'' & 0.072 \stdvalue{0.432} & 0.177 \stdvalue{0.429} & 0.292 \stdvalue{0.307} & 0.713 \stdvalue{0.239} & 0.598 \stdvalue{0.233} \\
    \multicolumn{6}{>{\columncolor{mygray}}c}{\textit{Phi3-3.8B}}\\ \midgrayline
    prompting & 0.135 \stdvalue{0.431} & 0.012 \stdvalue{0.393} & 0.300 \stdvalue{0.336} & 0.740 \stdvalue{0.275} & 0.659 \stdvalue{0.222} \\
    \midgrayline
    ``so'' & 0.122 \stdvalue{0.435} & 0.108 \stdvalue{0.408} & 0.221 \stdvalue{0.283} & 0.653 \stdvalue{0.216} & 0.535 \stdvalue{0.175} \\
    ``because'' & 0.074 \stdvalue{0.432} & 0.031 \stdvalue{0.428} & 0.113 \stdvalue{0.297} & 0.581 \stdvalue{0.222} & 0.503 \stdvalue{0.140} \\
    ``since'' & 0.074 \stdvalue{0.429} & 0.051 \stdvalue{0.416} & 0.130 \stdvalue{0.291} & 0.592 \stdvalue{0.217} & 0.501 \stdvalue{0.139} \\
    ``as'' & 0.066 \stdvalue{0.421} & 0.036 \stdvalue{0.424} & 0.114 \stdvalue{0.293} & 0.582 \stdvalue{0.217} & 0.496 \stdvalue{0.131} \\
    ``therefore'' & 0.108 \stdvalue{0.434} & 0.117 \stdvalue{0.411} & 0.223 \stdvalue{0.290} & 0.656 \stdvalue{0.222} & 0.537 \stdvalue{0.180} \\
    ``thus'' & 0.128 \stdvalue{0.436} & 0.121 \stdvalue{0.409} & 0.234 \stdvalue{0.285} & 0.661 \stdvalue{0.220} & 0.537 \stdvalue{0.182} \\
    ``hence'' & 0.124 \stdvalue{0.430} & 0.118 \stdvalue{0.413} & 0.242 \stdvalue{0.286} & 0.670 \stdvalue{0.223} & 0.547 \stdvalue{0.189} \\

    \multicolumn{6}{>{\columncolor{mygray}}c}{\textit{Phi3-7B}}\\ \midgrayline
    prompting & 0.092 \stdvalue{0.443} & 0.204 \stdvalue{0.422} & 0.347 \stdvalue{0.348} & 0.753 \stdvalue{0.254} & 0.672 \stdvalue{0.220} \\
    \midgrayline
    ``so'' & 0.101 \stdvalue{0.417} & 0.150 \stdvalue{0.420} & 0.243 \stdvalue{0.288} & 0.669 \stdvalue{0.217} & 0.538 \stdvalue{0.178} \\
    ``because'' & 0.030 \stdvalue{0.411} & 0.028 \stdvalue{0.426} & 0.040 \stdvalue{0.310} & 0.525 \stdvalue{0.226} & 0.490 \stdvalue{0.121} \\
    ``since'' & 0.060 \stdvalue{0.417} & 0.054 \stdvalue{0.433} & 0.068 \stdvalue{0.303} & 0.538 \stdvalue{0.220} & 0.487 \stdvalue{0.117} \\
    ``as'' & 0.024 \stdvalue{0.418} & 0.041 \stdvalue{0.421} & 0.053 \stdvalue{0.304} & 0.534 \stdvalue{0.224} & 0.490 \stdvalue{0.120} \\
    ``therefore'' & 0.111 \stdvalue{0.425} & 0.145 \stdvalue{0.421} & 0.230 \stdvalue{0.290} & 0.655 \stdvalue{0.220} & 0.534 \stdvalue{0.176} \\
    ``thus'' & 0.125 \stdvalue{0.409} & 0.155 \stdvalue{0.421} & 0.253 \stdvalue{0.285} & 0.672 \stdvalue{0.217} & 0.542 \stdvalue{0.182} \\
    ``hence'' & 0.123 \stdvalue{0.408} & 0.153 \stdvalue{0.416} & 0.260 \stdvalue{0.283} & 0.679 \stdvalue{0.215} & 0.544 \stdvalue{0.184} \\
    
    \multicolumn{6}{>{\columncolor{mygray}}c}{\textit{Phi3-14B}}\\ \midgrayline
    prompting & -0.056 \stdvalue{0.441} & 0.154 \stdvalue{0.406} & 0.356 \stdvalue{0.367} & 0.801 \stdvalue{0.286} & 0.801 \stdvalue{0.230}  \\ \midgrayline
    ``so'' & 0.150 \stdvalue{0.423} & 0.130 \stdvalue{0.403} & 0.286 \stdvalue{0.292} & 0.701 \stdvalue{0.226} & 0.576 \stdvalue{0.216} \\
    ``because'' & 0.096 \stdvalue{0.425} & 0.059 \stdvalue{0.412} & 0.120 \stdvalue{0.308} & 0.577 \stdvalue{0.234} & 0.509 \stdvalue{0.151} \\
    ``since'' & 0.106 \stdvalue{0.419} & 0.069 \stdvalue{0.429} & 0.142 \stdvalue{0.304} & 0.592 \stdvalue{0.229} & 0.509 \stdvalue{0.153} \\
    ``as'' & 0.105 \stdvalue{0.429} & 0.053 \stdvalue{0.424} & 0.124 \stdvalue{0.304} & 0.580 \stdvalue{0.229} & 0.506 \stdvalue{0.147} \\
    ``therefore'' & 0.135 \stdvalue{0.430} & 0.148 \stdvalue{0.409} & 0.293 \stdvalue{0.298} & 0.707 \stdvalue{0.226} & 0.577 \stdvalue{0.216} \\
    ``thus'' & 0.130 \stdvalue{0.426} & 0.149 \stdvalue{0.408} & 0.300 \stdvalue{0.290} & 0.714 \stdvalue{0.226} & 0.585 \stdvalue{0.218} \\
    ``hence'' & 0.134 \stdvalue{0.419} & 0.146 \stdvalue{0.417} & 0.303 \stdvalue{0.291} & 0.717 \stdvalue{0.226} & 0.586 \stdvalue{0.220} \\
    
    \multicolumn{6}{>{\columncolor{mygray}}c}{\textit{LLaMA3-8B}}\\ 
    \midgrayline
    prompting &  0.030 \stdvalue{0.444} & 0.139 \stdvalue{0.436} & 0.273 \stdvalue{0.387} & 0.712 \stdvalue{0.285} & 0.639 \stdvalue{0.217} \\
    \midgrayline
    ``so'' & 0.136 \stdvalue{0.432} & 0.139 \stdvalue{0.416} & 0.270 \stdvalue{0.310} & 0.688 \stdvalue{0.236} & 0.573 \stdvalue{0.212} \\
    ``because'' & 0.105 \stdvalue{0.428} & 0.103 \stdvalue{0.435} & 0.179 \stdvalue{0.325} & 0.619 \stdvalue{0.241} & 0.534 \stdvalue{0.175} \\
    ``since'' & 0.106 \stdvalue{0.441} & 0.111 \stdvalue{0.429} & 0.198 \stdvalue{0.330} & 0.635 \stdvalue{0.242} & 0.544 \stdvalue{0.179} \\
    ``as'' & 0.097 \stdvalue{0.436} & 0.101 \stdvalue{0.434} & 0.186 \stdvalue{0.325} & 0.628 \stdvalue{0.239} & 0.533 \stdvalue{0.172} \\
    ``therefore'' & 0.128 \stdvalue{0.441} & 0.137 \stdvalue{0.420} & 0.280 \stdvalue{0.319} & 0.699 \stdvalue{0.239} & 0.586 \stdvalue{0.217} \\
    ``thus'' & 0.114 \stdvalue{0.432} & 0.144 \stdvalue{0.415} & 0.286 \stdvalue{0.315} & 0.706 \stdvalue{0.237} & 0.591 \stdvalue{0.223} \\
    ``hence'' & 0.123 \stdvalue{0.438} & 0.139 \stdvalue{0.416} & 0.285 \stdvalue{0.315} & 0.704 \stdvalue{0.237} & 0.590 \stdvalue{0.224} \\
    \multicolumn{6}{>{\columncolor{mygray}}c}{\textit{LLaMA3-70B}}\\ 
    \midgrayline
    prompting & 0.357 \stdvalue{0.469} & 0.343 \stdvalue{0.419} & 0.586 \stdvalue{0.415} & 0.887 \stdvalue{0.274} & 0.923 \stdvalue{0.177} \\
    \midgrayline
    ``so'' & 0.150 \stdvalue{0.433} & 0.127 \stdvalue{0.416} & 0.302 \stdvalue{0.303} & 0.717 \stdvalue{0.227} & 0.588 \stdvalue{0.217} \\
    ``because'' & 0.089 \stdvalue{0.430} & 0.085 \stdvalue{0.427} & 0.176 \stdvalue{0.320} & 0.623 \stdvalue{0.232} & 0.523 \stdvalue{0.170} \\
    ``since'' & 0.099 \stdvalue{0.429} & 0.090 \stdvalue{0.421} & 0.193 \stdvalue{0.314} & 0.636 \stdvalue{0.236} & 0.535 \stdvalue{0.181} \\
    ``as'' & 0.102 \stdvalue{0.435} & 0.084 \stdvalue{0.414} & 0.187 \stdvalue{0.319} & 0.631 \stdvalue{0.236} & 0.533 \stdvalue{0.179} \\
    ``therefore'' & 0.155 \stdvalue{0.433} & 0.136 \stdvalue{0.427} & 0.301 \stdvalue{0.311} & 0.713 \stdvalue{0.235} & 0.593 \stdvalue{0.222} \\
    ``thus'' & 0.130 \stdvalue{0.434} & 0.130 \stdvalue{0.423} & 0.300 \stdvalue{0.313} & 0.718 \stdvalue{0.237} & 0.598 \stdvalue{0.225} \\
    ``hence'' & 0.133 \stdvalue{0.434} & 0.114 \stdvalue{0.423} & 0.296 \stdvalue{0.313} & 0.717 \stdvalue{0.238} & 0.600 \stdvalue{0.226} \\
    \bottomrule
\caption{Comparative results of different conjunction words on model performance, emphasizing their influence on maintaining causal epistemic consistency. }
\label{tab:results_models_with_different_conjunctions} 
\end{longtable}